\documentclass[10pt, letterpaper]{article}
\usepackage{hyperref}
\usepackage[utf8]{inputenc}
\usepackage[letterpaper, margin=1in]{geometry}
\usepackage{authblk}
\usepackage[english]{babel}
\usepackage{array,booktabs,multirow,threeparttable}
\usepackage{indentfirst}
\usepackage{url}
\usepackage{enumitem}

\usepackage{graphicx,subcaption}
\graphicspath{{figures/}, {figures/datasets/}, {figures/experiments/}}

\usepackage{amssymb,bm,mathtools}
\DeclareMathOperator*{\argmax}{arg\,max}

\usepackage[square,numbers,sort]{natbib}
\usepackage[ruled,vlined,linesnumbered]{algorithm2e}
\SetKwInOut{Parameter}{parameter}
\makeatletter
\newcommand{\algrule}[1][.01pt]{\par\vskip.5\baselineskip\hrule height #1\par\vskip.5\baselineskip}
\makeatother

\begin{document}
\title{Incremental cluster validity index-guided online learning for performance and robustness to presentation order}
\author[1]{Leonardo Enzo Brito da Silva}
\author[1]{Nagasharath Rayapati}
\author[2]{Donald C. Wunsch II}
\affil[1]{Guise AI Inc., USA}
\affil[2]{Applied Computational Intelligence Laboratory, Missouri University of Science and Technology, USA}
\date{\today}
\setcounter{Maxaffil}{0}
\renewcommand\Affilfont{\itshape\small}

\maketitle

\begin{abstract}
In streaming data applications incoming samples are processed and discarded, therefore, intelligent decision-making is crucial for the performance of lifelong learning systems. In addition, the order in which samples arrive may heavily affect the performance of online (and offline) incremental learners. The recently introduced incremental cluster validity indices (iCVIs) provide valuable aid in addressing such class of problems. Their primary use-case has been cluster quality monitoring; nonetheless, they have been very recently integrated in a streaming clustering method to assist the clustering task itself. In this context, the work presented here introduces the first adaptive resonance theory (ART)-based model that uses iCVIs for unsupervised and semi-supervised online learning. Moreover, it shows for the first time how to use iCVIs to regulate ART vigilance. The model achieves improved accuracy and robustness to ordering effects by combining topological adaptive resonance theory predictive mapping (TopoARTMAP) and an online iCVI framework --- thereby being named iCVI-TopoARTMAP ---, as well as employing iCVI-driven post-processing heuristics at the end of each learning step. The labeling component of the supervised learning system relies on iCVIs by replacing module B of TopoARTMAP with an online iCVI framework, which provides assignments of input samples to clusters at each iteration in accordance to any of several iCVIs. Moreover, when the novel iCVI-based match tracking is enabled, following successive worsening of the selected iCVI, the vigilance of iCVI-TopoARTMAP is changed until the iCVI value restarts to improve. At the end of each learning step, if enabled, any combination of the following operations can be performed: swap of categories, merge of clusters, split of clusters, prune-and-reassignment of categories, and compression of clusters. The iCVI-TopoARTMAP maintains useful properties shared by ARTMAP models, such as stability, immunity to catastrophic forgetting, and the many-to-one mapping capability via the map field module. On unsupervised learning experiments with a synthetic data set and deep embeddings of a real-world face image data set, iCVI-TopoARTMAP outperformed several state-of-the-art ART-based models and another iCVI-based online clustering algorithm. Moreover, while some methods showed drastic performance changes upon different orderings, the iCVI-TopoARTMAP yielded consistent performance in all experiments, which was either superior or comparable to performances of the other methods. Similarly, in our experiments, iCVI-TopoARTMAP in semi-supervised learning mode yielded results that were either superior or comparable to a supervised nearest neighbor classifier in prediction mode. 
\end{abstract}
\noindent \textbf{Index terms.} \textit{Data streams, clustering, incremental cluster validity index (iCVI), adaptive resonance theory (ART), adaptive resonance theory predictive mapping (ARTMAP), unsupervised online learning, semi-supervised online learning, face recognition.}

\section{Introduction} \label{sec:intro}

Cluster validation (or assessment)~\cite[Chapter~10]{xu2009} is an essential part of cluster analysis and consists of measuring the quality of data partitions yielded by clustering algorithms. In the past, this task was performed using \textit{batch cluster validity indices} (bCVIs)~\cite{milligan1985, Bezdek1997, Halkidi2002a, Halkidi2002b, vendramin2010, Arbelaitz2013, Hamalainen.2017a} to evaluate clustering algorithms after completion. In 2018, Moshtaghi et al.~\cite{Moshtaghi2018, Moshtaghi2018b} introduced the idea of \textit{incremental cluster validity indices} (iCVIs) to visually monitor and evaluate cluster footprints yielded by online clustering algorithms in data stream applications, thus being referred to as \textit{incremental stream monitoring functions} (iSMFs)~\cite{Bezdek.2021a, Wu.2020a}. A recursive formulation was derived for a quantity present in sum-of-squares (SS)-based CVIs, namely the fuzzy compactness, consequently allowing for the incremental computation of a related class of bCVIs. Since then, efforts have been employed to introduce incremental versions of popular SS- and non-SS-based bCVIs, such as: Xie-Beni (XB~\cite{Xie1991}, iXB~\cite{Moshtaghi2018, Moshtaghi2018b}), Davies-Bouldin (DB~\cite{db}, iDB~\cite{Moshtaghi2018, Moshtaghi2018b}), generalized Dunn 43 and 53 (GDs~\cite{Bezdek.1998a}, iGDs~\cite{Ibrahim2018b}), Calinski-Harabasz (CH~\cite{vrc}, iCH~\cite{leonardo.2019c, leonardo.2020a}), Pakhira-Bandyopadhyay-Maulik (PBM~\cite{pbm}, iPBM~\cite{leonardo.2019c, leonardo.2020a}), WB-index (WB~\cite{Zhao.2014a}, iWB~\cite{leonardo.2019c, leonardo.2020a}), centroid-based Silhouette (SIL~\cite{Rawashdeh2012}, iSIL~\cite{leonardo.2019c, leonardo.2020a}), partition coefficient and exponential separation (PCAES~\cite{Wu.2005a}, iPCAES~\cite{Ibrahim.2019a}), representative cross information potential (rCIP~\cite{araujo20132}, irCIP~\cite{leonardo.2019c, leonardo.2020a}), representative cross-entropy (rH~\cite{araujo20132}, irH~\cite{leonardo.2019c, leonardo.2020a}), negentropy increment (NI~\cite{fernandez2010}, iNI~\cite{leonardo.2019c, leonardo.2020a}), and conn\_index (conn\_index~\cite{tasdemir2011}, iconn\_index~\cite{leonardo.2019c, leonardo.2020a}) — where the above citations are in order (original bCVI [reference(s)], iCVI version [reference(s)]).\footnote{Before the increased use of iCVIs (particularly since 2018) there were mainly batch cluster validation indices, so they were known simply as cluster validity indices or CVIs. We, and others, have used the small letter ``i'' in the acronym for iCVIs to emphasize their incremental nature, suitable for iterative algorithms. To discuss the current and previous approaches together, we refer to the latter as bCVIs. We describe in~\cite{leonardo.2020c} that there is usually no reason to use a bCVI for which an iCVI has been developed.}

Batch cluster validity indices have also been traditionally used as fitness functions in optimization algorithms to perform  clustering (e.g., \cite{xu2012, leonardo.2014a}) or to aid in the clustering process --- for instance, Brito da Silva and Wunsch~\cite{leonardo2017} presented a fuzzy adaptive resonance theory (ART)~\cite{Carpenter1991} augmented with bCVIs as an additional vigilance test, while Smith and Wunsch~\cite{smith2015} used them for vigilance parameter setting within the neural network. Chenaghlou~\cite[Chapter~6]{Milad.2019a} introduced the usage of iCVIs in the online clustering task~\cite{Skrjanc.2019a, Carnein.2019a} by presenting an incremental clustering system that combined two online clustering algorithms (sequential k-means~\cite{skm} and online clustering and anomaly detection~\cite{Milad.2018a}) and an iXB-based controller to determine the creation and merging of cluster prototypes. In addition, Ibrahim et al.~\cite{Ibrahim2018} used the compactness to aid in making decisions regarding the emergence of clusters when performing online incremental clustering. Although iCVIs were not widely used prior to 2018, we did find earlier contributions: Lughofer~\cite{lughofer2008} presented an ART-like online incremental clustering algorithm that used a non-SS-based iCVI~\cite{Yang2001} to guide a splitting and merging heuristic. 

\textit{Adaptive resonance theory predictive mapping} (ARTMAP) neural networks~\cite{Carpenter1991a, carpenter1992, Carpenter1995c, kasuba1993} are built upon elementary \textit{adaptive resonance theory} (ART) models (e.g., \cite{Carpenter1987, Carpenter1991}), and their classical applications are for supervised learning~\cite{leonardo.2019b}. However, some variants exist that have been repurposed for unsupervised machine learning applications, such as biclustering (BARTMAP~\cite{xu2011}, hierarchical BARTMAP~\cite{Kim2016}, and TopoBARTMAP~\cite{Raghu.2020a}), hierarchical divisive clustering (SMART~\cite{Bartfai1994}), and offline iCVI-based clustering (iCVI-ARTMAP~\cite{leonardo.2020c}). Furthermore, the Unified ART model and its Extended version~\cite{seiffertt2010} can perform mixed-modality learning. In particular, hierarchical BARTMAP introduced a hierarchy of biclusters whose levels are evaluated using a bCVI; TopoBARTMAP introduced the usage of fuzzy TopoART~\cite{Tscherepanow2010} as fuzzy ARTMAP's building blocks; iCVI-ARTMAP introduced the usage of an iCVI-based framework to replace fuzzy ARTMAP's labeling module; Unified and Extended Unified ARTs introduced integrated learning dynamics which depends upon the type of inputs provided to the system, wherein supervised, unsupervised or reinforcement learning modes have different priorities.\footnote{Note that the ARTMAP-based model in~\cite{leonardo.2019c, leonardo.2020a} used to incrementalize the conn\_index relies on ground truth labels for the purposes of a clustering algorithm agnostic experimentation and hence does not perform unsupervised machine learning.}

Our previous work~\cite{leonardo.2020c} demonstrated how to integrate fuzzy ARTMAP~\cite{carpenter1992, Carpenter1995c} and iCVIs for offline incremental unsupervised learning. This work goes further to allow for online learning by incorporating: 
\begin{enumerate}[label=(\alph*)]
\item Online normalization, online complement-coding, and online long-term memory (weight vector) re-scaling based on~\cite{Swope2012, Meng.2015a, Meng.2019a}.
\item Split, merge, swap operations. Among methods designed to perform data stream related tasks, heuristics performing some combination of such strategies are typically used~\cite{Carnein.2019a}; however, to the best of our knowledge these do not make use of iCVIs.
\item Compression via fuzzy ARTMAP (``self-supervised learning'').
\item A prune-and-reassign heuristic.
\item Summary statistics associated with each category. Storing summary statistics as multiple ``micro-clusters'' are common approaches in the data stream literature~\cite{Carnein.2019a}.
\item A novel match tracking procedure (namely, the iCVI-based match tracking), to aid in dynamically restructuring clusters using feedback from iCVIs.
\end{enumerate}
\noindent The iCVI-TopoARTMAP model considers not only the local similarities of samples to categories, but also the global effect of assigning the presented sample to a given cluster (i.e., the outcome with respect to the entire cluster structure), which is considered by the iCVI-based label generation module. Experiments with synthetic data as well as online face clustering and classification applications were conducted to show the effectiveness of iCVI-TopoARTMAP. The contributions of this work are three-fold:
\begin{enumerate}
\item The iCVI-TopoARTMAP model. The latter is, to the best of our knowledge, the first combination of a model from the ARTMAP family with iCVIs for the purposes of online unsupervised and semi-supervised learning. No prior ART method has been able to combine iCVIs and multi-prototype representation-capable supervised methods such as ARTMAP to perform online learning.  
\item ``Weight re-scaled'' versions of dual vigilance fuzzy ART (DVFA)~\cite{leonardo.2018b} and topological fuzzy ART (TopoFA)~\cite{Tscherepanow2010}. These models cannot handle situations wherein the data ranges are unknown; therefore, to benchmark the main contribution of this work (iCVI-TopoARTMAP) against a larger pool of state-of-the-art ART-based models, it was necessary to augment them with the features of (a) online normalization, (b) online complement coding, and (c) online weight vector re-scaling, which were introduced in~\cite{Swope2012, Meng.2015a, Meng.2019a}, thereby yielding the WS-TopoFA and WS-DVFA variants.
\item A comparison of the performance of state-of-the-art ART-based models as well as incremental sequential k-means~\cite[Chapter~6]{Milad.2019a} (one of the two sole iCVI-based online clustering methods currently available in the literature) in streaming experiments with synthetic and real-world data sets.
\end{enumerate}

To the best of our knowledge, with the exception of~\cite{lughofer2008, Milad.2019a}, no other streaming clustering algorithm makes use of iCVIs. Moreover, these approaches do not make use of multi-prototype-based representation nor the variety of the iCVI-driven aforementioned operations (swap, merge, split). The remainder of this paper is organized as follows: Section~\ref{sec:design} describes the iCVI-TopoARTMAP model; Sections~\ref{Sec:exp_synthetic_data} and~\ref{Sec:exp_rw_data} discuss the experiments carried out, the results obtained, and our findings; finally, Section~\ref{sec:the_conclusion} concludes this paper.   

\section{iCVI-TopoARTMAP design} \label{sec:design}

\begin{figure}[!t]
\centering
\includegraphics[width=0.9\textwidth]{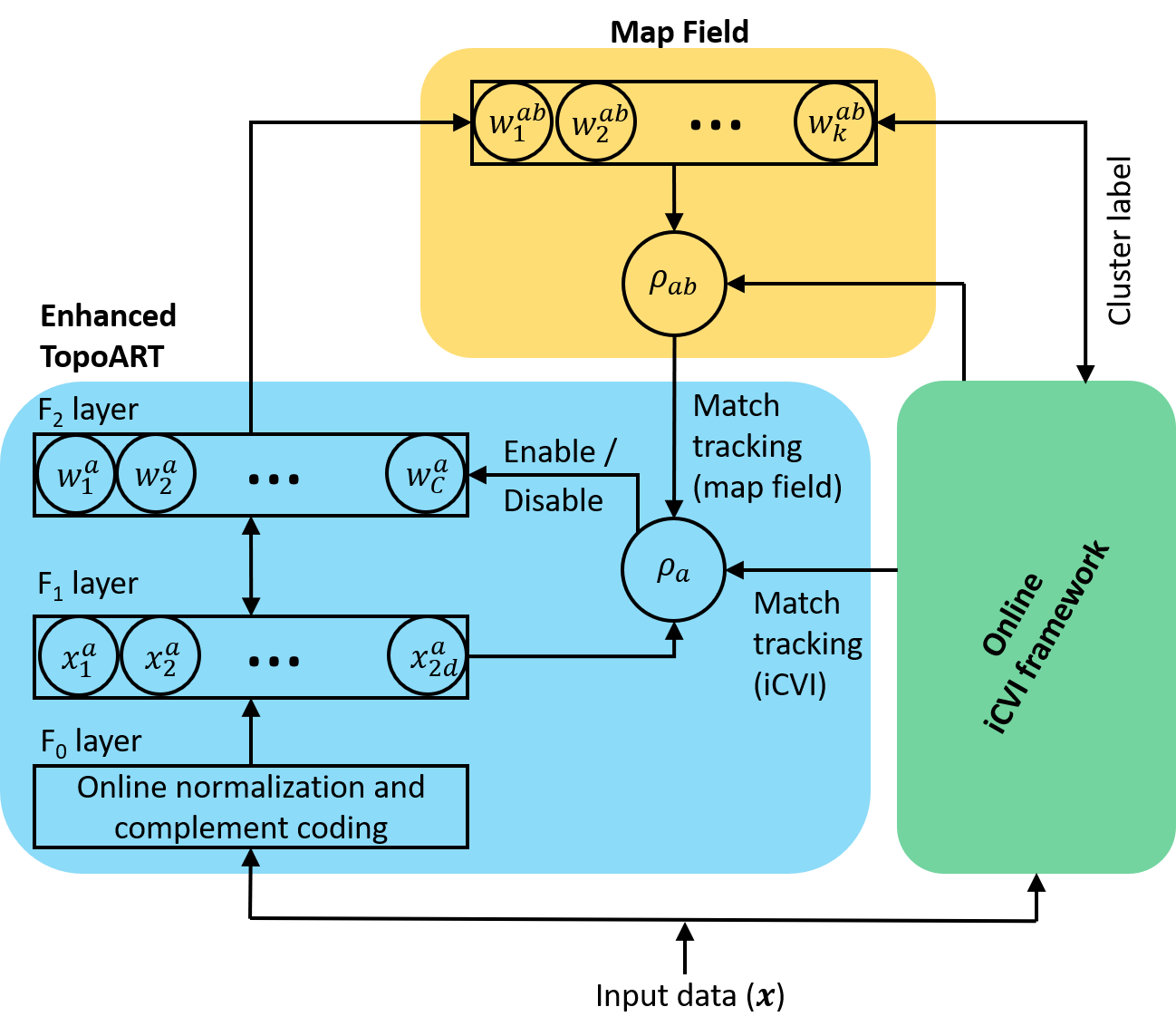}
\caption{The iCVI-TopoARTMAP neural network.}
\label{fig:icvitopoartmap}
\end{figure} 

Clustering is an unsupervised learning task; thus, no labels are provided to classify data into groups. The model presented here, namely the iCVI-TopoARTMAP, circumvents this problem by using iCVIs for the purpose of label generation. Specifically, the iCVI-TopoARTMAP (shown in Fig.~\ref{fig:icvitopoartmap}) allows a user to select an iCVI and perform online incremental multi-prototype-based unsupervised and semi-supervised learning. The iCVI-TopoARTMAP significantly improves the performance of state-of-the-art iCVI-based online clustering methods~\cite{Milad.2019a}; the latter work is, to the best of our knowledge, the sole other SS-iCVI-based alternative to online clustering. Specifically, the online learning method presented here uses an iCVI for online decision making regarding:
\begin{enumerate}[label=(\alph*)]
\item Sample allocation to clusters.
\item Prototype allocation to clusters.
\item Merging clusters.
\item Splitting of clusters (iCVI guidance depends on the split strategy).
\item iCVI-based match tracking.
\end{enumerate}
\noindent while incrementally building associative mappings between prototypes (categories) to clusters. Items (b)-(e) above can be individually enabled and disabled by the user (e.g., in accordance with the task and/or data at hand), while item (a) is automatically disregarded by the system if a supervised label is presented alongside the current sample.

The iCVI-TopoARTMAP is an ARTMAP-based neural network; therefore, it consists of three main parts:
\begin{enumerate}
\item \textbf{Module A}. The module A is a variant of fuzzy TopoART~\cite{Tscherepanow2010} that has been enhanced with online min-max normalization and complement-coding~\cite{carpenter1992} for input samples as well as re-scaling of weight vectors (based on~\cite{Swope2012, Meng.2015a, Meng.2019a}), summary statistics associated with each category (frequency, mean, and compactness), a $CONN$ matrix~\cite{tasdemir2006, tasdemir2009}, and an inactivity counter per category.
\item \textbf{Module B}. The module B is an online version of the iCVI framework~\cite{leonardo.2020c} that was further enhanced. The user can seamlessly select the iCVI to match the input data characteristics and/or the application at hand. The online iCVI framework is responsible for:
\begin{itemize}
\item Computing temporary iCVI values to assign the presented sample to each cluster of the current data partition and thus construct crisp labels for the presented input in the form of a matrix. The rows of such label matrix generated by the iCVI framework follow the one-hot encoding standard and correspond to the optimal assignment --- in terms of the selected iCVI --- of a sample at the time of its presentation. 
\item Updating the iCVI values and variables over time whenever necessary in the following cases: 
\begin{itemize}
\item After a sample assignment.
\item After a category swap.
\item After a cluster merge.
\item After a cluster split.
\item After a category prune-and-reassign.
\item After a compression.
\end{itemize}
\item Maintaining and updating additional quantities specific to the iCVI, which include, for instance:
\begin{itemize}
\item Cluster statistics and entire data statistics (i.e., frequencies, means, and compactnesses) when using SS-based iCVIs.
\item Mapping of prototypes to clusters, frequency statistics, number of samples exclusively within clusters (i.e., both first and second best matching units fall within the same cluster), and number of prototypes when using the iconn\_index. Note in the computation of iconn\_index in this work, the frequency statistics are considered independently of the $CONN$ matrix, and thus, while the sum of the frequency statistics is equal to the total number of samples seen, the sum of the upper (or lower) diagonal of the $CONN$ may not be because of the vigilance criteria imposed to the first and second resonant categories within module A.
\end{itemize} 
\end{itemize} 
\item \textbf{Map field}. The map field is a variant of fuzzy ARTMAP's map field~\cite{carpenter1992, Carpenter1995c}, and it has been enhanced with the following attributes: the iCVI-TopoARTMAP allows for a label matrix containing multiple cluster assignments in one-hot encoded form, when using the online iCVI-framework to generate labels (Section~\ref{sec:icvi-framework}). Therefore, the map field may perform multiple vigilance checks and then output the maximum across all cluster assignments within the label matrix generated by the iCVI-framework.  Note that the map field vigilance test is disregarded until a second cluster is detected. This will delay the computation of the selected iCVI value until a second category is created as per the user-selected vigilance parameter of module A; however, iCVI-specific quantities and variables are still continuously and incrementally updated. Notably, the map field associative network allows for multi-prototype representation of clusters.
\end{enumerate}

For clarity, the main parameters of the iCVI-TopoARTMAP (and their default values) are listed in Table~\ref{Tab:parameters}. The remainder of this section provides the design details of each module of this neural network.

\begin{table}[!t]
\centering
\caption{The iCVI-TopoARTMAP parameters.}
\resizebox{\columnwidth}{!}{
\begin{threeparttable}
\begin{tabular}{lll}
\toprule
\textbf{Parameter} 
& \textbf{Description} 
& \textbf{Default value}
\\
\midrule
\midrule
$\rho_a$
& 
Module A's vigilance parameter.
& 
$0$\tnote{a} \\
$\beta_1$
& 
Module A's learning rate for the first resonant category. 
& 
$1$ \\
$\beta_2$
& 
Module A's learning rate for the second resonant category. 
& 
$0$ \\
$\alpha$
& 
Module A's choice parameter. 
& 
$0.001$ \\
$M_{type}$
& 
Module A's match function type
& 
Eq.~(\ref{eq:res_2}) \\
$EN\_{T^u}$
& 
Module A's enable flag for the uncommitted category 
& 
True \\
$\varepsilon$
&
Map field's match tracking parameter. 
& 
$0.01$ \\
$\rho_{ab}$
&
Map field's vigilance parameter. 
& 
$1$ \\
$\beta_{ab}$
& 
Map field's learning rate. 
& 
$1$ \\
$L_{type}$
&
Map field learning type (i.e., fixed or variable number of clusters). 
& 
$variable$ \\
iCVI
&
Selected iCVI and its corresponding parameters (if any). 
& 
- \\
$EN\_MT_{icvi}$
&
Enable flag for iCVI-based match tracking. 
& 
True \\
$\varepsilon_{icvi}$
&
iCVI-based match tracking parameter. 
& 
$\rho_{MT_{icvi}} - \rho_a$ \\
$\rho_{MT_{icvi}}$
&
iCVI-based match tracking vigilance parameter. 
& 
- \\
$EN\_swap$
&
Enable flag for swap strategy.
& 
True \\
$EN\_merge$
&
Enable flag for merge strategy.
& 
True \\
$EN\_split$
&
Enable flag for split strategy.
& 
True \\
$S\_type$
&
Split strategy type.
& 
activity-based \\
$EN\_compress$
&
Enable flag for compress strategy.
& 
True \\
$\rho_c$
&
Vigilance parameter for compression strategy.
& 
- \\
$EN\_prune\_reassign$
&
Enable flag for prune-and-reassign strategy.
& 
True \\
$\tau$
&
iCVI checks threshold.
& 
- \\
$\phi$
&
Threshold  for the number of samples of a cluster.
& 
- \\
$\xi$
&
Category inactivity threshold.
& 
- \\
\bottomrule
\end{tabular}
\begin{tablenotes}[normal,flushleft]
\item[a] Except for iconn\_index, wherein $\rho_a$ must reflect the required multi-prototype representation. The default $M_{type}$ is being considered.
\end{tablenotes}
\end{threeparttable}
}
\label{Tab:parameters}
\end{table}

\subsection{Module A: Enhanced Topological Fuzzy ART}  \label{Sec:module_a}

Module A of iCVI-TopoARTMAP is a variant of Topological Fuzzy ART~\cite{Tscherepanow2010}, namely the Enhanced Topological Fuzzy ART (E-TopoFA), which was augmented with several features that will be discussed in the following sections.

\subsubsection{Online normalization of inputs, complement coding, and online re-scaling of weight vectors} \label{sec:scaling}

In general, knowledge of the data range is absent in online learning cases. This is a challenge for Fuzzy ART-based neural networks~\cite{leonardo.2019b} because they need normalized (i.e., in the range $[0, 1]$) and complement coded inputs~\cite{carpenter1992}. Moreover, when input samples push the lower and upper bounds of the data range interval, its weight vectors (long-term memory) must be re-scaled. This is one of the general challenges in ART-based online learning~\cite{leonardo.2019b}, and recent works have employed  different approaches to address it~\cite{Swope2012, Meng.2015a, Meng.2019a, Park.2019a}. Here, the recursive re-scaling formulation presented in~\cite{Swope2012} is used to adapt the weight vectors and normalize the input samples. However, unlike~\cite{Swope2012}, the minimum and maximum statistics stored are immediately overwritten if the incoming sample pushes the data interval like in~\cite{Meng.2015a, Meng.2019a}. Note that the input is complement-coded; thus the first half and the complement of the second half of the weight vector undergo such re-scaling. 

If the maximum and/or minimum values of the data features are different from the current stored statistics (i.e., $\bm{x}_{max} \in \mathbb{R}^d$ and $\bm{x}_{min} \in \mathbb{R}^d$, respectively), module A's weight vectors (i.e., $\bm{w}_j = \left[\bm{u}_j, \bar{\bm{v}}_j \right]\in \mathbb{R}^{2d}$) undergo online re-scaling. First, the components $\bm{u}_j = \left[\bm{w}_{(j,1)}, \cdots, \bm{w}_{(j,d)}\right]$ and $\bm{v}_j = \Vec{\bm{1}} - \bar{\bm{v}}_j=\Vec{\bm{1}} - \left[w_{(j,d+1)}, \cdots, w_{(j,2d)}\right]$ of each category $j$ of module A undergo an inverse transformation given by

\begin{equation}
\bm{u}_j \leftarrow \mathcal{T}^{-1} (\bm{u}_j),
\end{equation}
\begin{equation}
\bm{v}_j \leftarrow \mathcal{T}^{-1} (\bm{v}_j),
\end{equation}
\noindent where the transformation $\mathcal{T}(\bm{z})$ represents a component-wise standard min-max normalization of a vector $\bm{z}$ whose component-wise maxima and minima are the vectors $\bm{z}_{max}$ and $\bm{z}_{min}$:
\begin{equation}
\mathcal{T}(z_i) = \dfrac{z_i - z_{min,i}}{z_{max,i} - z_{min,i}}.
\end{equation}

Then, $\bm{x}_{max}$ and $\bm{x}_{min}$ are updated to reflect the new feature-wise maxima and minima. Finally, components $\bm{u}_j$ and $\bm{v}_j$ are re-scaled using
\begin{equation}
\bm{u}_j^{new}  = \mathcal{T}\left( \bm{u}_j^{old} \right),
\label{Eq:T_u}
\end{equation}
\begin{equation}
\bm{v}_j^{new}  = \mathcal{T}\left( \bm{v}_j^{old} \right),
\label{Eq:T_v}
\end{equation}
\begin{equation}
\bm{u}_j^{new} \leftarrow \max \left( \bm{u}_j^{new} , \Vec{\bm{0}} \right),
\label{Eq:u_rescaled}
\end{equation}
\begin{equation}
\bm{v}_j^{new}  \leftarrow \min \left( \bm{v}_j^{new} , \Vec{\bm{1}} \right),
\label{Eq:v_rescaled}
\end{equation}
\noindent where the $\max(\cdot, \cdot)$ and $\min(\cdot, \cdot)$ operations are taken component-wise. Note that if $v_{j,i}^{new} \leq 1, \forall i$, then:
\begin{equation}
\bar{\bm{v}}_j^{new} = \Vec{\bm{1}} - \bm{v}_j^{new} = \mathcal{T}(\bar{\bm{v}}_j^{old}) = 
\mathcal{T}(\Vec{\bm{1}} - \bm{v}_j^{old}),
\label{Eq:vbar_rescaled}
\end{equation}
\noindent and that the component-wise recursive formulations of Eqs.~(\ref{Eq:T_u}), (\ref{Eq:T_v}), and~(\ref{Eq:vbar_rescaled}) are given by:
\begin{equation}
z_{j,i}^{new} = 
\left( \dfrac{ x_{max,i}^{old} - x_{min,i}^{old} }{ x_{max,i}^{new} - x_{min,i}^{new} } \right) z_{j,i}^{old} + \dfrac{ x_{min,i}^{old} - x_{min,i}^{new}}{ x_{max,i}^{new} - x_{min,i}^{new} },
\label{Eq:rec_1}
\end{equation}
\begin{equation}
z_{j,i}^{new}  = \left( \dfrac{x_{max,i}^{old} - x_{min,i}^{old}}{x_{max,i}^{new} - x_{min,i}^{new}} \right) z_{j,i}^{old}  + \dfrac{ x_{max,i}^{new} - x_{max,i}^{old}}{x_{max,i}^{new} - x_{min,i}^{new}},
\label{Eq:rec_2}
\end{equation}
\noindent where $z_{j,i}$ represents $u_{j,i}$ or $v_{j,i}$ in Eq.~(\ref{Eq:rec_1}) and $\bar{v}_{j,i}$ in Eq.~(\ref{Eq:rec_2}).

Finally, the weight vector of each category $j$ is updated by concatenating the re-scaled components obtained from Eqs.~(\ref{Eq:u_rescaled}) and~(\ref{Eq:v_rescaled}):
\begin{equation}
\bm{w}_j  \leftarrow \left[ \bm{u}_j^{new} , \Vec{\bm{1}} - \bm{v}_j^{new} \right].
\end{equation}

\subsubsection{Summary statistics, inactivity vector, and connectivity matrix} \label{sec:footprints}

Online clustering algorithms usually store summary statistics (or ``footprints'') which are incrementally updated once an assignment has been made for the current input sample~\cite{Carnein.2019a}. Such footprints can be in the form of ``micro-clusters'' (as per the data stream literature terminology) which can be used to alter the perceived cluster structure underlying the streaming data (e.g., via heuristics for splitting, merging, etc.). Note that we follow the neural-network-based clustering terminology and refer to these ``micro-clusters'' as multi-prototype representation. 

Here, E-TopoFA also stores footprints: in addition to the standard weight vectors $\bm{w}_j$, each category $j$ has associated with it an inactivity counter ($\bar{a}_j$) as well as the summary statistics of frequency ($n_j \in \mathbb{N}^1$), mean ($\bm{\mu}_j \in \mathbb{R}^d$), and hard compactness ($CP_j \in \mathbb{R}^1$):
\begin{equation}
\bm{\mu}_j = \dfrac{1}{n_j}\sum \limits_{i=1}^{n_j} \bm{x}_i,
\label{Eq:CP}
\end{equation}
\begin{equation}
CP_j = \sum \limits_{i=1}^{n_j} \| \bm{x}_i - \bm{\mu}_j \|^2_2.
\label{Eq:CP}
\end{equation}
\noindent Furthermore, a $P \times P$ connectivity matrix ($CONN_{P \times P}$)~\cite{tasdemir2006, tasdemir2009} containing local density information is also stored, where $P$ represents the number of categories. All of these are continuously and incrementally updated during learning. Note that TopoFA~\cite{Tscherepanow2010} only stores the accumulated frequency (i.e., sample count) for each category and a binary adjacency matrix in addition to the standard weight vectors, while the probabilistic fuzzy ARTMAP~\cite{Lim1997, Lim2000a} only stores additional centroids.

\subsubsection{Learning}  \label{sec:module_a_learning}

When a sample $\bm{x}$ is presented, first the online re-scaling procedure discussed in Section~\ref{sec:scaling} takes place. Next, a copy of $\bm{x}$ undergoes min-max normalization using the updated $\bm{x}_{max}$ and $\bm{x}_{min}$ statistics and subsequent complement coding~\cite{carpenter1992}:
\begin{equation}
\bm{x}^a  \leftarrow \left[ \mathcal{T}(\bm{x}) , \Vec{\bm{1}} - \mathcal{T}(\bm{x}) \right].
\end{equation}

Since module A is an enhanced TopoART, it retains the main dynamics of fuzzy TopoART~\cite{Tscherepanow2010}, which in turn are akin to the dynamics of fuzzy ART~\cite{Carpenter1991}. After presenting the input $\bm{x}^a$, the activation function $T_j^a$ of each category is computed as
\begin{equation}
T_j^a = \dfrac{\|\bm{x}^a \wedge \bm{w}_j \|_1}{\alpha + \| \bm{w}_j \|_1},~ \alpha > 0,
\label{eq:activation}
\end{equation}
\noindent where $T_j^a$ and $\bm{w}_j$ are the activation and weight vector of category $j$, respectively; the operator $\wedge$ represents a component-wise minimum between two vectors, and $\|\cdot\|_1$ is the $\ell_1$ norm. To be considered resonant, an existing category $J_1$ must simultaneously satisfy all the following criteria (Eqs.~(\ref{eq:res_1})-(\ref{eq:res_2})):
\begin{equation}
T_{J_1}^a > T^u,
\label{eq:res_1}
\end{equation}
\begin{equation}
M_{J_1}^a = \dfrac{\|\bm{x}^a \wedge \bm{w}_{J_1} \|_1}{ \| \bm{x}^a \|_1} \geq \rho_{a}, 
\label{eq:res_2}
\end{equation}
\noindent where $T^u=\frac{d}{\alpha + d}$ (the practitioner has the option to parameterize the system to disregard the constraint represented by Eq.~(\ref{eq:res_1}), wherein $T^u=-1$), $M_{J_1}^a$ is the match value of category $J_1$ with weight vector $\bm{w}_{J_1}$, and $0 \leq \rho_{a} \leq 1$ is the vigilance parameter of module A. According to the problem at hand, a cosine distance-based match function $M_{J_1}^a$ can be used instead of Eq.~(\ref{eq:res_2}):
\begin{equation}
M_{J_1}^a = 1 - \dfrac{\langle \bm{x}^b , \bm{\mu}_{J_1} \rangle}{ \| \bm{x}^b \|_2 \| \bm{\mu}_{J_1} \|_2} \leq \rho_{a},~ 0 \leq \rho_{a} \leq 2, 
\label{eq:cosine_vigilance}
\end{equation}
\noindent where $\bm{x}^b=\bm{x}$ is the original input to the online iCVI framework, $\bm{\mu}_{J_1}$ is the local mean statistic associated with category $J_1$, $\| \cdot \|_2$ is the $\ell_2$ norm, and $\langle\cdot,\cdot\rangle$ is the inner product. The cosine similarity has been extensively used to compare embeddings generated by deep feature extractors (e.g., person re-identification~\cite{Wojke.2017a, Wojke.2018a}, face recognition~\cite{Klemen.2018a, Wang.2018a, Cao.2018a, Elmahmudi.2019a}).

When a resonant category $J_1$ is found for sample $\bm{x}$, then its associated $\bm{w}_{J_1}$, $n_{J_1}$, $\bm{\mu}_{J_1}$, and $CP_{J_1}$ are updated incrementally using the following formulae~\cite{Moshtaghi2018,Moshtaghi2018b,carpenter1992,leonardo.2020c}:
\begin{equation}
\bm{w}_{J_1} \leftarrow (1 - \beta_1)\bm{w}_{J_1} + \beta_1(\bm{x}^a \wedge \bm{w}_{J_1}), ~ 0 < \beta_1 \leq 1,
\label{Eq:weight_update}
\end{equation}
\begin{equation}
\bm{\mu}_{J_1} \leftarrow \dfrac{n_{J_1}}{n_{J_1} + 1}\bm{\mu}_{J_1} + \dfrac{1}{n_{J_1} + 1}\bm{x}^b,
\label{Eq:mean_update}
\end{equation}
\begin{equation}
CP_{J_1} \leftarrow CP_{J_1} + \dfrac{n_{J_1}}{n_{J_1} + 1} || \bm{x}^b - \bm{\mu}_{J_1} ||_2^2,
\label{Eq:comp_update}
\end{equation}
\begin{equation}
n_{J_1} \leftarrow n_{J_1} + 1.
\label{Eq:freq_update}
\end{equation}

If category $J_1$ does not satisfy Eqs.~(\ref{eq:res_1}) and~(\ref{eq:res_2}) or~(\ref{eq:res_1}) and~(\ref{eq:cosine_vigilance}), then it is inhibited, the search continues with the next highest ranked category, and the process is repeated. This loop continues until either (i) a category meets all resonance criteria, or (ii) the set of categories with an activation greater than $T^u$ is exhausted and thus a new category is created. If the latter case takes place, i.e., if none of the categories simultaneously satisfy the constraints, then a new category is created and initialized as
\begin{equation}
\bm{w}_{J_1} \leftarrow \bm{x}^a,
\label{Eq:weight_new}
\end{equation}
\begin{equation}
n_{J_1} \leftarrow 1,
\label{Eq:freq_new}
\end{equation}
\begin{equation}
\bm{\mu}_{J_1} \leftarrow \bm{x}^b,
\label{Eq:mean_new}
\end{equation}
\begin{equation}
CP_{J_1} \leftarrow 0.
\label{Eq:comp_new}
\end{equation}

The inactivity counters of all categories are then incremented by one, i.e., $\bar{a}_j \leftarrow \bar{a}_j + 1, \forall j$. However, the resonant category $J_1$ has its inactivity counter reset to zero (or initialized in case a new category is created), i.e., $\bar{a}_{J_1} \leftarrow 0$. 

If a first resonant category $J_1$ is found within the set of existing categories, then the search for the second resonant category $J_2$ starts from the subsequent highest ranked (if any) as per the activation function (Eq.~(\ref{eq:activation})). However, if a new category is created, then the search for a second resonant category is performed on the full set of previously existing categories in descending order of activation (Eq.~(\ref{eq:activation})). Note that the second resonant category only needs to satisfy the constraints expressed by inequalities~(\ref{eq:res_1}) and~(\ref{eq:res_2}) or~(\ref{eq:res_1} and~(\ref{eq:cosine_vigilance}) while using the baseline vigilance parameter (i.e., the original user-defined module A vigilance parameter) --- within the iCVI-TopoARTMAP architecture, the first resonant category also needs to satisfy the map field vigilance test (Section~\ref{Sec:mapfield}). If a second resonant category $J_2$ is found, then only its weight vector is updated:
\begin{equation}
\bm{w}_{J_2} \leftarrow (1 - \beta_2)\bm{w}_{J_2} + \beta_2(\bm{x}^a \wedge \bm{w}_{J_2}), ~ 0 \leq \beta_2 \leq \beta_1 \leq 1,
\label{Eq:weight_update_J2}
\end{equation}

Finally, when both the first and a second resonant categories are found ($J_1$ and $J_2$, respectively) then two entries of the $CONN$ matrix are incrementally updated as follows:
\begin{equation}
CONN(J_1, J_2) \leftarrow CONN(J_1, J_2) + 1,
\label{Eq:CONN_ij}
\end{equation}
\begin{equation}
CONN(J_2, J_1) \leftarrow CONN(J_2, J_1) + 1.
\label{Eq:CONN_ji}
\end{equation}

\subsection{Module B: Online iCVI-framework}  \label{sec:icvi-framework}

The online iCVI-framework extends the component in~\cite{leonardo.2020c} to online learning applications. It makes use of iCVIs and thus, for brevity, we refer the reader to~\cite{Moshtaghi2018, Moshtaghi2018b, Ibrahim2018b, leonardo.2019c, leonardo.2020a} for a comprehensive treatment of their formulations.

\subsubsection{iCVI-based label generation}

The online iCVI-framework is responsible for generating, using iCVIs, a cluster label matrix $\bm{Y}$ whose rows follow a one-hot encoding and represent the best current sample assignment regarding the cluster structures according to the selected iCVI. Let $T_i^b$ be the temporary iCVI value (or its negative if the iCVI is min-optimal) corresponding to assigning the current sample $\bm{x}$ to cluster $i$ thereby yielding the vector $\bm{T}^b=[T_1^b, \cdots, T_i^b, \cdots, T_k^b]$, where $k$ is the current number of clusters. Moreover, let set $\mathcal{C}=\{c_i | c_i=T^b_{max}\}$, where $T_{max}^b = \max\limits_i(T_i^b)$ --- i.e., $\mathcal{C}$ is the set of clusters that equally optimize the iCVI for the current sample $\bm{x}$. Then, a matrix of labels $\bm{Y}=[y_{i,j}]_{m \times k}$ is generated as:
\begin{equation}
y_{i,j} = 
\begin{cases}
1 & \text{, if} \quad j = c_i\\
0 & \text{, otherwise}
\end{cases}, \quad i \in \left\{1, \cdots, m\right\}, \quad j \in \left\{1, \cdots, k\right\}
\label{eq:y_matrix}
\end{equation}
\noindent where $1 \leq m \leq k$ is the number of clusters that equally optimize the selected iCVI, and $c_i \in \mathcal{C}$ is the $i^{th}$ cluster that optimize the iCVI (note that $c_i$ is not necessarily equal to $i$, for instance, $\mathcal{C}$ may be $\left\{2, 7, 5\right\}$). There may be situations in which assignments to different clusters may lead to the same maximum value $T^b_{max}$; matrix $\bm{Y}$ addresses such a case of existing more than one optimal assignment (i.e., $|\mathcal{C}|>1$). Each row $\bm{y}_r$ ($r \in \{1, \cdots, m\}$) of $\bm{Y}$ is thus a one-hot encoded label vector indicating a cluster assignment that optimizes the selected iCVI. If there is a unique optimal cluster assignment that optimizes the selected iCVI for the presented sample $\bm{x}$ (i.e., $|\mathcal{C}|=1$), then $\bm{Y}$ reduces to a single row vector as defined in~\cite{leonardo.2020c}; the latter work disregards the case of multiple optimal assignments. 

\paragraph{SS-based iCVIs.} If the selected iCVI is SS-based, then the online iCVI framework stores cluster-wise statistics, the whole data statistic, the variables and current value of the selected iCVI. The computations required for the temporary iCVI values ($T^b_j$) to consider the presented sample $\bm{x}^b$ are performed incrementally: the statistics are recomputed using their corresponding recursive equations (e.g., Eqs.~(\ref{Eq:mean_update})-(\ref{Eq:freq_update}) with $i$ in place of $J_1$ to represent cluster $i$); only the iCVI variables that should change because of the assignment are recomputed (like~\cite{leonardo.2020a,leonardo.2020c}, iCVI variables are cached), and then the iCVI values are recomputed. After all the temporary values are computed and $\bm{T}^b$ is generated, then the set $\mathcal{C}$ is identified and Eq.~(\ref{eq:y_matrix}) is employed to construct the label matrix $\bm{Y}$. 

\paragraph{iconn\_index.} The iconn\_index is a graph-based iCVI that relies on multi-prototype representation of clusters. This representation allows for the inference of information about local densities between prototypes, which are encoded within the $CONN$ matrix. In order to compute $T^b_j$ for the assignment of sample $\bm{x}$ to cluster $j$, the first and second (if any) prototype winners for that sample must be identified. Therefore, if the selected iCVI is the iconn\_index, then the strategy used to generate set $\mathcal{C}$ is different. First note that the temporary values $T^b_j$ for the SS-based iCVIs are computed ``as if'' $\bm{x}$ was assigned to cluster $j$. Analogously, to temporarily recompute the iconn\_index, first, a row vector $\bm{y}^{(c_i)}$ representing the one-hot encoded label for cluster $c_i$ is generated as
\begin{equation}
y_j^{(c_i)} = 
\begin{cases}
1 & \text{, if} \quad j = c_i\\
0 & \text{, otherwise}
\end{cases}.
\end{equation}
\noindent Next, virtual copies of current module A and the map field are trained for a single epoch using the pair $(\bm{x},\bm{y}^{(c_i)})$. Then, the first and second (if any) resonant categories are identified and used to incrementally update the $CONN$ matrix of module A's virtual copy (Eqs.~(\ref{Eq:CONN_ij})-(\ref{Eq:CONN_ji})). This $CONN$ matrix is then used to solely recompute the iconn\_index variables that should change (like the SS-based iCVIs, iconn\_index variables are also cached), and then the iconn\_index value is incrementally recomputed ($T^b_j$). This process is repeated using other virtual copies (a copy of current module A and a copy of the map field per cluster) for all clusters identified by the system to generate $\bm{T}^b$. Finally, the set $\mathcal{C}$ is identified and Eq.~(\ref{eq:y_matrix}) is employed to construct the label matrix $\bm{Y}$. 

\subsubsection{iCVI-based match tracking} \label{sec:icvi_mt}

If the iCVI-based match tracking is enabled, then it is engaged as follows: when the iCVI tracker variable ($v$) reaches a user-defined predetermined threshold value ($\tau$), then the vigilance parameter of module A ($\rho_a$) is altered as
\begin{equation}
\rho_a \leftarrow \max\left[ \min \left(\rho_a + \varepsilon_{icvi}, \rho_{MT_{icvi}} \right), 0\right]
\label{Eq:icvi_MT_sd}
\end{equation}
in cases where the user has selected the cosine distance as the match function, $\rho_a$ is instead changed using
\begin{equation}
\rho_a \leftarrow \min\left[ \max \left(\rho_a - \varepsilon_{icvi}, \rho_{MT_{icvi}} \right), 2\right]
\label{Eq:icvi_MT_cos}
\end{equation}
\noindent where $\varepsilon_{icvi}$ is the iCVI-based match tracking parameter (user-defined) and $\rho_{MT_{icvi}}$ is the iCVI-based match tracking associated vigilance parameter. For an instantaneous change, it can be automatically set, for instance, as $\varepsilon_{icvi}=(\rho_{MT_{icvi}} - \rho_a)$ for Eq.~(\ref{Eq:icvi_MT_sd}). When the iCVI tracker variable ($v$) falls below $\tau$ then the vigilance value of module A is reset to its original user-defined value.

\subsubsection{Learning}  \label{sec:module_b_learning}

After the dynamics of module A and the map field have taken place during training, the online iCVI-framework incrementally updates specific quantities, variables, and value of the user-specified iCVI. This is accomplished using the label associated module A's resonant category $J_1$ (i.e., $\argmax\limits_i(\bm{w}^{ab}_{J_1, i})$) and either the presented sample $\bm{x}$ for SS-based iCVIs or $J_1$, $J_2$ (if any), and the $CONN$ matrix for the iconn\_index.

\subsection{Map field}  \label{Sec:mapfield}

\subsubsection{Vigilance test and match tracking}

The iCVI-TopoARTMAP can process a label matrix that can possibly encode multiple cluster assignments (i.e., $\bm{Y}=[y_{i,j}]_{m \times k}$, $m \geq 1$ in Eq.~(\ref{eq:y_matrix}), generated by the online iCVI-framework --- see Section~\ref{sec:icvi-framework}), and this must be considered by the dynamics of the map field. This is accomplished by performing multiple vigilance checks when necessary, and in such case, outputting the maximum across all possible cluster assignments. 

Specifically, the effective map field match value with respect to category $J_1$ is given by
\begin{equation}
M_{J_1}^{ab} = \max\limits_{i} \left[ M_{J_1, i}^{ab} \right],
\label{eq:MF_function}
\end{equation}
\noindent where
\begin{equation}
M_{J_1, i}^{ab} = \dfrac{\|\bm{y}_i \wedge \bm{w}_{J_1}^{ab} \|_1}{\| \bm{y}_i \|_1},
\label{eq:res_4}
\end{equation}
\noindent and $\bm{w}_{J_1}^{ab}$ is the $J_1^{th}$ row vector of the map field mapping matrix $\bm{W}^{ab} \in \mathbb{R}^{P \times k}$ associated with category $J_1$, and $\bm{y}_i$ corresponds to the $i^{th}$ row of matrix $\bm{Y}$. The effective map field match value (Eq.~(\ref{eq:MF_function})) is defined as the maximum among the match values (Eq.~(\ref{eq:res_4})). Note that if a supervised label $\bm{y}^{(SL)}$ was provided, then the following map field vigilance test would be performed instead of (Eqs.~(\ref{eq:MF_function})-(\ref{eq:res_4})):
\begin{equation}
M_{J_1}^{ab} = \dfrac{\|\bm{y}^{(SL)} \wedge \bm{w}_{J_1}^{ab} \|_1}{\| \bm{y}^{(SL)} \|_1}.
\label{eq:res_5}
\end{equation}

Naturally, if the map field vigilance test is disabled, then it would not be performed, and a resonant category $J_1$ would only need to satisfy Eqs.~(\ref{eq:res_1}) and~(\ref{eq:res_2}) (or only Eq.~(\ref{eq:res_2}) if the constraint represented by Eq.~(\ref{eq:res_1}) was disabled). Otherwise, in order to learn, the current module A resonant category $J_1$ must satisfy Eqs.~(\ref{eq:res_1}) and~(\ref{eq:res_2}) as well as the following inequality: 
\begin{equation}
M_{J_1}^{ab} \geq \rho_{ab}, 
\label{eq:res_3}
\end{equation}
\noindent where $0 \leq \rho_{ab} \leq 1$ is the vigilance parameter of the map field. If module A's category $J_1$ satisfies Eqs.~(\ref{eq:res_1}) and~(\ref{eq:res_2}) or~(\ref{eq:res_1}) and~(\ref{eq:cosine_vigilance}) but does not satisfy the map field vigilance test expressed by Eq.~(\ref{eq:res_3}), then a mismatch occurs, and standard match tracking engages (either MT+~\cite{Carpenter1991a,carpenter1992} or MT-~\cite{Carpenter1998b}) thus modifying the vigilance parameter by a small value $\varepsilon$. In case of a standard match function (i.e., Eq.~(\ref{eq:res_2})), the vigilance is changed to
\begin{equation}
\rho_a \leftarrow \max\left[ \min \left(M_{J_1}^a + \varepsilon, 1 \right), 0\right],
\end{equation}
\noindent in case of the cosine distance match function (i.e., Eq.~(\ref{eq:cosine_vigilance})), then $\rho_a$ is instead changed using
\begin{equation}
\rho_a \leftarrow \min\left[ \max \left(M_{J_1}^a - \varepsilon, 0 \right), 2\right],
\end{equation}
\noindent where $\epsilon$ is the map field match tracking parameter. Then, category $J_1$ is inhibited, the search continues with the next highest ranked category, and the process repeats until a resonant category is found or the set of categories is exhausted and a new one is created (see Section~\ref{sec:module_a_learning}). 
  
\subsubsection{Learning} \label{sec:map_field_learning}
  
The map field adaptation is given by
\begin{equation}
\bm{w}_{J_1}^{ab} \leftarrow (1 - \beta_{ab})\bm{w}_{J_1}^{ab} + \beta_{ab}(\bm{x}^{F_{ab}} \wedge \bm{w}_{J_1}^{ab}), \quad  0 < \beta_{ab} \leq 1,
\label{Eq:MF_update}
\end{equation}
\noindent where $\bm{x}^{F_{ab}} = \bm{y}^{(SL)}$ in case a supervised label is provided, $\bm{x}^{F_{ab}} = \bm{y}_{\argmax\limits_i\left( M_{J_1,i}^{ab}\right)}$ if the iCVI framework generated a label matrix $\bm{Y}$, or $\bm{x}^{F_{ab}} = \Vec{\bm{1}}$ if the map field vigilance test was disabled. In cases involving the creation of a new category in module A, the system can update the map field matrix in two different manners, which should be defined a priori by the user ($L_{type}$ parameter in Table~\ref{Tab:parameters}):
\begin{itemize}
\item Variable: the map field mapping matrix $\bm{W}^{ab}$ is expanded (1 row and 1 column) such that the new category encodes a new cluster:  $\bm{x}^{F_{ab}} = [\Vec{\bm{0}} ~ 1] \in \mathbb{R}^{k+1}$  (i.e., $\bm{x}^{F_{ab}}$ is the concatenation of the k-dimensional vector $\Vec{\bm{0}}$ and the 1-dimensional vector $1$) and $\bm{w}_{J_1}^{ab}=\Vec{\bm{1}}$ in Eq.~(\ref{Eq:MF_update}).
\item Fixed: the map field matrix is expanded (1 row) and $\bm{w}_{new}^{ab} \leftarrow \bm{y}$, where $\bm{y} = \bm{y}^{(SL)}$ in case a supervised label is provided, or $\bm{y} = \bm{Y}$ when the online iCVI-framework generates a label matrix with a single row (i.e., SS-based iCVIs). This case corresponds to a fixed number of clusters.
\end{itemize}

\subsection{Post-processing strategies} \label{sec:operations}

During online learning, incorrect decisions made in the past may need to be corrected. These may occur, for instance, due to ordering effects, no knowledge of the data range, and system parameterization. Therefore, online learning systems usually make use of additional operations to improve performance such as split and merge strategies~\cite{lughofer2008, Skrjanc.2019a, Carnein.2019a, leonardo.2020b}. Here the iCVI-TopoARTMAP has the capability of using five strategies that can be individually enabled: (1) swap, (2) merge, (3) split, (4) prune-and-reassign, and (5) compress. These operations aim to re-organize the cluster structures, incrementally, for each sample presentation. Note that after these operations are carried out, the iCVI-TopoARTMAP variables must be changed accordingly; if these strategies take place, then some or all of the following items occur:
\begin{enumerate}[label=(\alph*)]
\item The iCVI-framework updates its internal iCVI variables and value to reflect changes.
\item The map field mapping matrix $\bm{W}^{ab}$ entries (i.e., the mapping of categories to clusters) are modified accordingly.
\item The module A variables (e.g., weight vectors, summary statistics, etc.) are modified as per the compression operation.
\end{enumerate}

\subsubsection{Swap} \label{sec:swap}

The system attempts to swap categories among clusters if such operation incurs a better iCVI value, i.e., the swap strategy consists of the following: the system emulates a constrained swap operation for all categories from their original to another cluster. The swap operation is constrained because the set of clusters to which a category can be swapped is limited to the clusters it is connected to as per the $CONN$ matrix: a category $c$ from cluster $i$ can be (temporarily or not) swapped to cluster $j$ if and only if it is connected to at least one category belonging to cluster $j$, i.e., if $CONN(c,g)>0$ for some category $g$ of cluster $j$. If the best iCVI value among these swap operations incurs an improvement of the original clustering structure (as measured by the iCVI), then this operation effectively takes place. The swap strategy is repeated until no swap operation leads to an iCVI improvement. In addition, a swap operation is only carried out if there are currently more than two categories. Note that a cluster represented by a single category will disappear if it is selected to undergo a swap operation; however, a swap whose outcome would be a single unique cluster is not allowed. Concretely, a swap operation entails removing a category c (i.e., a split) from a cluster $i$ and immediately assigning it to another cluster $j$ (i.e., a merge).

\paragraph{SS-based iCVIs.} 

In a case where an SS-based iCVI was selected, since each prototype has local statistics associated with it in the raw data domain, swapping corresponds to computing the following:
\begin{enumerate}
\item Split
\end{enumerate}
\begin{equation}
\bm{\mu}_i \leftarrow \dfrac{n_i}{n_i - n_c} \bm{\mu}_i - \dfrac{n_c}{n_i - n_c} \bm{\mu}_c,
\label{Eq:mean_s1}
\end{equation}
\begin{equation}
CP_i \leftarrow CP_i - CP_c - \dfrac{n_i n_c}{n_i - n_c}||\bm{\mu}_c - \bm{\mu}_i||_2^2,
\label{Eq:comp_s1}
\end{equation}
\begin{equation}
n_i \leftarrow n_i - n_c.
\label{Eq:freq_s1}
\end{equation}
\begin{enumerate}
\item Merge
\end{enumerate}
\begin{equation}
\bm{\mu}_j \leftarrow \dfrac{n_j}{n_j + n_c} \bm{\mu}_j + \dfrac{n_c}{n_j + n_c} \bm{\mu}_c,
\label{Eq:mean_m1}
\end{equation}
\begin{equation}
CP_j \leftarrow CP_j + CP_c + \dfrac{n_jn_c}{n_j + n_c} ||\bm{\mu}_c - \bm{\mu}_i||_2^2,
\label{Eq:comp_m1}
\end{equation}
\begin{equation}
n_j \leftarrow n_j + n_c.
\label{Eq:freq_m1}
\end{equation}

\paragraph{iconn\_index.} 

If the selected iCVI is the iconn\_index (graph-based iCVI), then swapping a category $c$ from cluster $i$ to cluster $j$ (mapped via map field) entails changing the mapping of categories to clusters for the iconn\_index while the CONN matrix, which is stored within module A, remains the same. When emulating or performing a swap operation, the cached iCVI specific quantities and variables that are related to such operation are recomputed within the online iCVI-framework, thus allowing for the iCVI value incremental recomputation. 

When a swap is performed, the iCVI framework updates the specific quantities, variables, and value of the selected iCVI. The map field matrix $\bm{W}^{ab}$ is also updated to reflect such change. Module A remains the same.

The swap strategy relates to the heuristic in the iCVI-ARTMAP model~\cite{leonardo.2020c}. However, the key difference is that, instead of swapping samples, categories (or prototypes) are swapped between clusters. The swap strategy is similar to the approach presented in~\cite{Araujo.2010a} in the context of offline clustering, wherein prototype labels are swapped to optimize an information-theoretic criterion.

\subsubsection{Merge} \label{sec:merge}

The system emulates consecutive cluster footprint merging until only two clusters are left, and the best cluster structure, according to the selected iCVI, is retained. Two clusters $i$ and $j$ are merged if they yield the best iCVI value compared to all other cluster merges at that stage. If the best iCVI value across all merge stages is an improvement over the current clustering state, then this operation effectively takes place, and the associated cluster footprint structure is retained. The merge strategy is only carried out if there are currently more than two clusters and the iCVI tracker variable is currently equal to zero (i.e, $v=0$ --- see Section~\ref{sec:train}).

\paragraph{SS-based iCVIs.}

In cases where an SS-based iCVI was selected, the merge operation is carried out like~\cite{leonardo.2020c}. In particular, merging of entire clusters $i$ and $j$ is accomplished via Eqs.~(\ref{Eq:mean_m1})-(\ref{Eq:freq_m1}), where $j$ and $c$ would represent clusters (either single or multiple-prototype), i.e., clusters $i$ and $j$ statistics are merged rather than merging cluster $j$ statistics with the local statistics of module A's category $c$.

\paragraph{iconn\_index.}

Like the swap operation (Section~\ref{sec:swap}), if the selected iCVI is the iconn\_index (graph-based iCVI), merging clusters $i$ and $j$ (mapped via map field) requires changing the mapping of categories to clusters in the iconn\_index while the $CONN$ matrix remains the same (it is stored within module A). When emulating or performing a merge operation, the cached iCVI specific quantities and variables that are related to such operation are recomputed. Next, the iCVI value is recomputed.

When merge is effectively performed, the iCVI framework updates the specific quantities, variables, and value of the selected iCVI. The map field matrix $\bm{W}^{ab}$ is also updated to reflect such change. Module A remains the same.

\subsubsection{Split} \label{sec:split}

The split strategy ($S_{type}$ = activity-based) consists of the following: the most recently active category is enforced to constitute a cluster on its own (category $i$ inactivity is measured by its associated inactivity counter value $\bar{a}_i$). Note, however, that if the category selected for splitting is the only category assigned to a cluster (i.e., a single-prototype cluster), then the search proceeds with the subsequent most recent category~$j$ ($\bar{a}_j \leq \bar{a}_k, \forall k \neq i$). This process continues until (i) a new cluster consisting of a single category is created or (ii) no category could become a new cluster and thus the split operation could not take place. The split strategy is only carried out if the iCVI tracker variable is currently greater than the same threshold defined for iCVI-based match tracking (i.e., $v > \tau$ --- see Section~\ref{sec:icvi_mt}).

\paragraph{SS-based iCVIs.}

In the case of an SS-based iCVI, splitting a cluster $i$ footprint consists of removing category $c$ that was mapped to it and creating a new cluster $l$. This is accomplished via Eqs.~(\ref{Eq:mean_s1})-(\ref{Eq:freq_s1}), and the new cluster $l$ statistics are the local statistics associated with module A's category $c$:
\begin{equation}
n_l \leftarrow n_c,
\end{equation}
\begin{equation}
\bm{\mu}_l \leftarrow \bm{\mu}_c,
\end{equation}
\begin{equation}
CP_l \leftarrow CP_c.
\end{equation}

\paragraph{iconn\_index.}

If the selected iCVI is the iconn\_index (graph-based iCVI), then splitting cluster $i$ entails changing the mapping of categories to clusters in the iconn\_index while the CONN matrix remains the same ($CONN$ matrix is stored within module A). When emulating or performing a split operation, the cached iCVI specific quantities and variables that are related to such operation are recomputed, thus allowing for the iCVI value incremental recomputation.

The user may also choose from two other types of split strategy to be carried out when the splitting is enabled (the split type to be used by the system must be decided by the user a priori when setting the parameters of the iCVI-TopoARTMAP). The other two split variations are:
\begin{itemize}
\item $S_{type}$ = full-cluster-decomposition: Each cluster is completely decomposed, i.e., all its categories are considered as their own cluster and the temporary iCVI value is incrementally recomputed (note that in SS-based iCVIs, the statistics of these categories now represent the statistics of new clusters; for the iconn\_index, splitting involves a change of the mapping of categories to clusters). Next, the merge operation is performed until there is no longer an improvement of the iCVI value. This operation is performed for each cluster, and the best split cluster is compared to the original cluster iCVI value: if the split strategy resulted in an improvement, then that structure is retained. This split strategy is repeated until it yields no iCVI improvement. 
\item $S_{type}$ = partial-cluster-decomposition: A cluster $i$ is split following 2 steps: (1) the category that, when detached from cluster $i$, yields the best temporary iCVI value compared to the other categories from the same cluster is selected to initialize a new cluster, and (2) categories from cluster $i$ are swapped between this new cluster and cluster $i$ (the category from item (1) cannot be swapped) until there is no temporary iCVI improvement. Across all clusters that underwent this 2-step split, the one with the best temporary iCVI value is selected to be effectively split. iCVI values are recomputed incrementally (only necessary specific quantities and variables are changed). The splitting operation may have some additional constraints: clusters already represented by a single category cannot be split. 
\end{itemize}

Our preliminary results indicated that $S_{type}$ set to activity-based yielded the most consistent results across the iCVI-TopoARTMAP variants, and thus this split heuristic is the one used in the experimental Sections~\ref{Sec:exp_synthetic_data} and~\ref{Sec:exp_rw_data}. When a split strategy is performed (regardless of the split type selected by the user), then the iCVI framework updates the specific quantities, variables, and value of the selected iCVI. The map field matrix $\bm{W}^{ab}$ is also updated to reflect such change. Module A remains the same.

\subsubsection{Compress} \label{sec:compress}

In many applications, memory/storage requirements must be observed, and thus iCVI-TopoARTMAP is equipped with a compression strategy to reduce its memory footprint (considering the number of categories as a proxy for model size). The system attempts to compress the categories of module A using a modified fuzzy ARTMAP. Let $\mathcal{H}$ be the set of iCVI-TopoARTMAP categories whose inactivity are greater or equal to a user defined threshold $\xi$ and let $\bar{\mathcal{H}}$ be its complement, i.e. $\mathcal{H} = \{ category~i | \bar{a}_i \geq \xi\}$ and $\bar{\mathcal{H}} = \{ category~i | \bar{a}_i < \xi\}$. The compression strategy is only carried out if $\mathcal{H} \neq \{ \emptyset \}$. First, iCVI-TopoARTMAP module A's categories from $\bar{\mathcal{H}}$ are copied to initialize the modified fuzzy ARTMAP module A's weight vectors $\bm{h}_i^a$ and map field's mapping matrix $\bm{H}^{ab}$. These will not be allowed to learn. Next, the input pairs $(\bm{w}_i^a,\bm{w}_i^{ab})$, $i \in \mathcal{H}$, are fed to the modified fuzzy ARTMAP module with a vigilance parameter $\rho_c$ (a user-defined parameter for its module A) and remaining parameters identical to the ones of iCVI-TopoARTMAP (i.e., $\varepsilon$, $\alpha$, $\beta_1$, and $\rho_{ab}$). 

The dynamics of the modified fuzzy ARTMAP are like a standard fuzzy ARTMAP. When $(\bm{w}_i^a,\bm{w}_i^{ab})$ is presented, the activations are computed as
\begin{equation}
T_j = \dfrac{\|\bm{w}_i^a \wedge \bm{h}_j^a \|_1}{\alpha + \| \bm{h}_j^a \|_1}.
\end{equation}
A resonant category $J$ with weight vector $\bm{h}_J$ from the modified fuzzy ARTMAP must simultaneously (i) not belong to $\bar{\mathcal{H}}$  (i.e., $J \notin \bar{\mathcal{H}}$) and (ii) satisfy the following two equations: 
\begin{equation}
\dfrac{\|\bm{w}_i^a \wedge \bm{h}_j^a \|_1}{d} \geq \rho_{c}, \quad 0 \leq \rho_{c} \leq 1, \quad i \in \mathcal{H}, 
\label{Eq:comp_res_1}
\end{equation}
\begin{equation}
\dfrac{\|\bm{w}_i^{ab} \wedge \bm{h}_j^{ab} \|_1}{d} \geq \rho_{ab}, \quad 0 \leq \rho_{ab} \leq 1, \quad i \in \mathcal{H}. 
\label{Eq:comp_res_2}
\end{equation}

The resonant category $J$ updates its weight vector $\bm{h}_J^a$ as per Eq.~(\ref{Eq:weight_update}) (with $\bm{w}_i^a$ in place of $\bm{x}^a$). Note that the entry of the map field matrix associated with a category remains fixed after it is first assigned (e.g., by initialization or by Eq.~(\ref{Eq:comp_hab_new})). If a category satisfies Eq.~(\ref{Eq:comp_res_1}) but not~(\ref{Eq:comp_res_2}), then match tracking engages with parameter $\varepsilon$. If a new category is created, then
\begin{equation}
\bm{h}_{new}^a \leftarrow \bm{w}_{i}^a,
\label{Eq:comp_ha_new}
\end{equation}
\begin{equation}
\bm{h}_{new}^{ab} \leftarrow \bm{w}_{i}^{ab}.
\label{Eq:comp_hab_new}
\end{equation}

Note that categories created by the modified fuzzy ARTMAP are appended to set $\mathcal{H}$. The training is performed until convergence (i.e., until there is no change in weight vectors $\bm{h}_i^a$ from one epoch to another). This procedure can be thought of as a form of ``self-supervised'' learning for iCVI-TopoARTMAP, since the label for each category $i \in \mathcal{H}$ with weight vector $\bm{w}_i^a$ is given by
\begin{equation}
label_i =\argmax\limits_i \left( \bm{w}_{i}^{ab} \right), \quad i \in \mathcal{H}.
\end{equation}

After compression, if the number of categories of the modified fuzzy ARTMAP ($P_{new}$) is smaller than the number of categories of the iCVI-TopoARTMAP ($P_{old}$), then the category and map field weights of the former are used to replace (i.e., overwrite) the latter:
\begin{equation}
\bm{W}^a \leftarrow \bm{H}^a,
\label{Eq:comp_Wa}
\end{equation}
\begin{equation}
\bm{W}^{ab} \leftarrow \bm{H}^{ab},
\label{Eq:comp_Wab}
\end{equation}
\noindent where $\bm{W}^a$ and $\bm{H}^a$ represent the category weight matrices (rows correspond to  category weight vectors) of iCVI-TopoARTMAP and the modified fuzzy ARTMAP, respectively; $\bm{W}^{ab}$ and $\bm{H}^{ab}$ represent the map field matrices of iCVI-TopoARTMAP and the modified fuzzy ARTMAP, respectively. Furthermore, iCVI-TopoARTMAP's variables must be updated to reflect such change. For instance, let $\mathcal{B}_i$ be the set of iCVI-TopoARTMAP module A's categories mapped to the trained modified fuzzy ARTMAP category $i$, thus the following is performed:
\begin{enumerate}[label=(\alph*)]
\item local statistics: the local statistics of the categories belonging $\mathcal{B}_i$ are used to incrementally compute the local statistic associated with category $i$ of the trained modified fuzzy ARTMAP via merges (Eqs.~(\ref{Eq:mean_m1})-(\ref{Eq:freq_m1}), where both $j$ and $c$ would represent categories).
\item inactivity: the inactivity of category $i$ of the trained modified fuzzy ARTMAP is defined as the smallest among the categories belonging $\mathcal{B}_i$.
\end{enumerate}

Items (a) and (b) above are performed for all categories of the trained modified fuzzy ARTMAP and used to overwrite the respective quantities of iCVI-TopoARTMAP's module A (as per Eqs.~(\ref{Eq:comp_Wa})-(\ref{Eq:comp_Wab}), the categories of the trained modified fuzzy ARTMAP categories were used to replace the ones of iCVI-TopoARTMAP). 

Finally, iCVI-TopoARTMAP's $CONN_{P_{old} \times P_{old}}$ matrix is updated to reflect such compression. Specifically, a new $CONN_{P_{new} \times P_{new}}$ is constructed from $CONN_{P_{old} \times P_{old}}$: 
\begin{equation}
CONN_{P_{new} \times P_{new}} (i,j) \leftarrow \sum\limits_{k \in \mathcal{B}_i}^{| \mathcal{B}_i |}\sum\limits_{l \in \mathcal{B}_j}^{| \mathcal{B}_j |} CONN_{P_{old} \times P_{old}}(k,l),
\end{equation} 
\noindent $CONN_{P_{new} \times P_{new}}$ is then used to overwrite $CONN_{P_{old} \times P_{old}}$ of iCVI-TopoARTMAP. Additional changes include, but are not limited to, removing categories and associated variables for which $\mathcal{B}_i=\{ \emptyset \}$.

Note that the online iCVI framework's specific quantities, variables and iCVI value remain the same for SS-based iCVIs, whereas these must be recomputed for the iconn\_index because of the $CONN$ matrix change.

The compress operation is inspired by the compression step used in~\cite{leonardo.2020b}, the key difference is that here a fuzzy ARTMAP is used for ``self-supervised'' learning, i.e., the inputs to its module A are the current weight vectors of the iCVI-TopoARTMAP's module A, whereas the inputs to module B are the weight vectors of iCVI-TopoARTMAP's map field. In this manner, the mapping of categories to clusters obtained so far is assumed to encode the true labels of iCVI-TopoARTMAP's module A categories. This operation has the goal of mitigating category proliferation, thus making the iCVI-TopoARTMAP model more compact. The outcome of this operation is a modification to the weight vectors ($\bm{W}^a$), local statistics and inactivity counter of module A's categories and the $CONN$ matrix, as well as the map field mapping matrix $\bm{W}^{ab}$ entries --- a change in the iCVI variables and value may also occur if the selected iCVI is iconn\_index. 

\subsubsection{Prune-and-reassign} \label{sec:prune_reassign}

A category is pruned if (i) its inactive counter is above the threshold parameter $\xi$, and (ii) the cluster to which it is associated encoded less than $\phi$ samples. Let $\mathcal{P}$ be the set of pruned categories; these categories are reassigned to the cluster associated with the closest category that does not belong to $\mathcal{P}$, i.e., the assignment of a category $i \in \mathcal{P}$ is given by
\begin{equation}
label_i = \argmax\limits_i \left( T_j^a \right), \quad j \notin \mathcal{P},
\end{equation}
\noindent where
\begin{equation}
T_j^a = \dfrac{\|\bm{w}_i^a \wedge \bm{w}_j^a  \|_1}{\alpha + \| \bm{w}_j^a \|_1}, \quad  i \in \mathcal{P}, \quad  j \notin \mathcal{P}.
\end{equation}
\noindent This approach is reminiscent of fuzzy topoART pruning strategy~\cite{Tscherepanow2010}; however, the latter considers the number of samples encoded by a category (not the cluster to which it belongs) and deletes such category. Moreover, it does not consider a category inactivity counter. The prune-and-reassign strategy is only carried out if $0<|\mathcal{P}|<P$ ($P$ is the current number of categories). Furthermore, this strategy must be accompanied by adequate changes to the map field and the iCVI specific quantities, variables, and value (iCVI related changes are performed by the iCVI framework).

\subsection{Training} \label{sec:train}

\begin{algorithm}[!p]
\DontPrintSemicolon
\SetNoFillComment
\SetKwInOut{KwIn}{Input}
\SetKwInOut{KwOut}{Output}
\SetKwComment{Comment}{}{}
\LinesNumbered
\KwIn{Streaming data $\bm{X}$, class label $y^{(SL)}$ (optional), and iCVI-TopoARTMAP parameters (Table~\ref{Tab:parameters}).}
\KwOut{Trained iCVI-TopoARTMAP.}
\algrule
\tcc{Notation}
\Comment*[l]{$t$: time (proxy for the number of samples presented).}
\Comment*[l]{$k_t$: number of clusters at time $t$.} 
\tcc{Training}
$t \leftarrow 0$.\\
\For{$\bm{x} \in \bm{X}$}{\label{alg:start}
$t \leftarrow t + 1$\\
\uIf{$t=1$}{
Initialize iCVI-TopoARTMAP.\\
}
\Else{
Temporarily store current iCVI value ($iCVI_{start}$).\\
\uIf{$\bm{y}^{(SL)} \neq \{\emptyset\}$}{
$\bm{Y} \leftarrow  \bm{y}^{(SL)}$. \\
}
\Else{	
\uIf{$k_t>1$}{
Present $\bm{x}$ to the online iCVI-framework to generate the label matrix $\bm{Y}$. \\
\If{$EN\_MT_{icvi}$}{
Call the $MT_{icvi}$ routine. \\
}
}
\Else{	
Disable the map field's vigilance test.\\
}
}
Carry out the learning dynamics of module A.\\  
Carry out the learning dynamics of map field.\\  
Update, within the iCVI-framework, the specific quantities, variables, and value of the iCVI. \\
\If{$EN\_merge$}{
Call the merge strategy routine. \\
}
\If{$EN\_split$}{
Call the split strategy routine. \\
}
\If{$EN\_swap$}{
Call the swap strategy routine. \\
}
\If{$EN\_compress$}{
Call the compress strategy routine. \\
}
\If{$EN\_prune\_reassign$}{
Call the prune-and-reassign strategy routine. \\
}
Get current iCVI value ($iCVI_{end}$).\\
\uIf{$iCVI_{start}$ is better than $iCVI_{end}$}{
$v \leftarrow v + 1$. \\
}
\Else{	
$v \leftarrow \max(0, v - 1)$.\\
}
}
}
\caption{iCVI-TopoARTMAP}\label{alg:icvi_topoartmap}
\end{algorithm}

Algorithm~\ref{alg:icvi_topoartmap} shows an online training procedure for the iCVI-TopoARTMAP after setting its parameters. The dynamics of the system are carried out as follows: it starts by presenting the input(s) to the network. If the current input(s) is/are the very first fed to the system, then the input(s) is/are used to initialize the network, i.e., module A, map field and iCVI framework and all their associated specific quantities, variables and values. For instance, the weight vector and local statistics of module A are initialized using Eqs.~(\ref{Eq:weight_new})-(\ref{Eq:comp_new}), the $CONN$ matrix is initialized as $[0]_{1 \times 1}$, the inactivity of the first category is initialized as $\bar{a}_1=0$ and the map field mapping matrix $\bm{W}^{ab}$ is initialized as $[1]_{1 \times 1}$. Next, the current value of the iCVI is stored for future comparison. The existence of a supervised class label $\bm{y}^{(SL)}$ is verified by the system, and, in the affirmative case, then the $(\bm{x},\bm{y}^{(SL)})$ pair is used to train module A and proceed with the remaining training steps; in the negative case, then the number of clusters ($k$) detected by the system is verified and if it is equal to one, then the map field vigilance test is disabled for the module A training before proceeding with the remaining training steps. Conversely, if $k>1$ (which is a necessary condition to compute iCVI values), then, at each iteration, the iCVI framework is engaged to generate a cluster label matrix $\bm{Y}$ using the selected iCVI representing the best placement(s) of the input sample (Section~\ref{sec:icvi-framework}), where an iCVI value is incrementally recomputed for the temporary assignment of the presented input sample to each existing cluster. Note that not all the iCVI quantities and variables are recomputed, solely the ones associated with the respective temporary cluster assignment. In this manner, supervised labels have priority in the system, like the mixed-modality learning system presented in~\cite{seiffertt2010}. The capability of also handling supervised labels allows for inclusion of external knowledge into the system at any time (e.g., initialization of the system using a priori knowledge and semi-supervised online learning mode --- i.e., map field with parameter $L_{type}$ set to fixed). In cases where supervised labels are provided, the iCVI variables and values are updated in accordance to such sample and label pairs. Note that the input(s) to the iCVI framework can be the raw input sample as well as data from module A and the map field. 

The map field and module A components retain many of their original dynamics~\cite{carpenter1992, Carpenter1995c, Carpenter1991, Tscherepanow2010}. Within module A, a copy of the presented sample $\bm{x}$ undergoes min-max normalization and complement-coding, thus becoming the input to module A ($\bm{x}^a$), whereas its raw copy (original sample $\bm{x}$) is used to update the summary statistics (Section~\ref{sec:footprints}) of the resonant category (or initialize the local statistics of a new category). The raw input (i.e., non-normalized) is also used for the SS-based iCVI computations performed via the iCVI framework. A sample $\bm{x}$ and its processed copy are discarded after being presented to the system. When a category resonates (i.e., satisfies both the module and Map field resonance checks), then learning ensues as described in Sections~\ref{sec:module_a_learning} and~\ref{sec:map_field_learning}. After the presentation of each sample, the system attempts to swap prototypes across clusters (if the swap operation is enabled), merge clusters (if the merge operation is enabled) and/or split clusters (if the split operation is enabled) to improve the partition quality measured by the selected iCVI. In addition, if the compress operation is enabled then the system attempts to reduce the number of categories in module A via a process that can be viewed as form of ``self-supervised learning''. Finally, the system attempts to prune-and-reassign prototypes across clusters (if the prune-and-reassign operation is enabled). Note that when performing these operations the iCVI-TopoARTMAP internal variables must be updated accordingly. Moreover, the iCVI variables associated with the current partition are cached~\cite{leonardo.2019c, leonardo.2020a, leonardo.2020c} to speed up the learning process in operations related to iCVIs.

At the end of the training (i.e., after the dynamics of module A, map field and enabled post-processing strategies have taken place), the previously stored iCVI value  is compared to its current value: if it is worse, then the iCVI check tracker variable $v$ is incremented by one, otherwise it is decreased by one. Specifically, in the latter case,
\begin{equation}
v \leftarrow \max(0, v-1).
\end{equation} 

\section{Experiments on synthetic data} \label{Sec:exp_synthetic_data}

\subsection{Data set and experimental setup} \label{sec:data_1_exps}

The synthetic data set\footnote{\label{ft1}available at \url{https://github.com/iskmeans/iskm}} used in this paper is akin to the one used in~\cite{Milad.2019a}. Specifically, it comprises $7$ $2$-dimensional Gaussian clusters and a total of $1600$ samples. This data set was used solely for benchmark purposes within the scope of this research paper.

\begin{figure}[!t]
\centering
\subcaptionbox{Class-incremental presentation}{\includegraphics[width=0.32\textwidth]{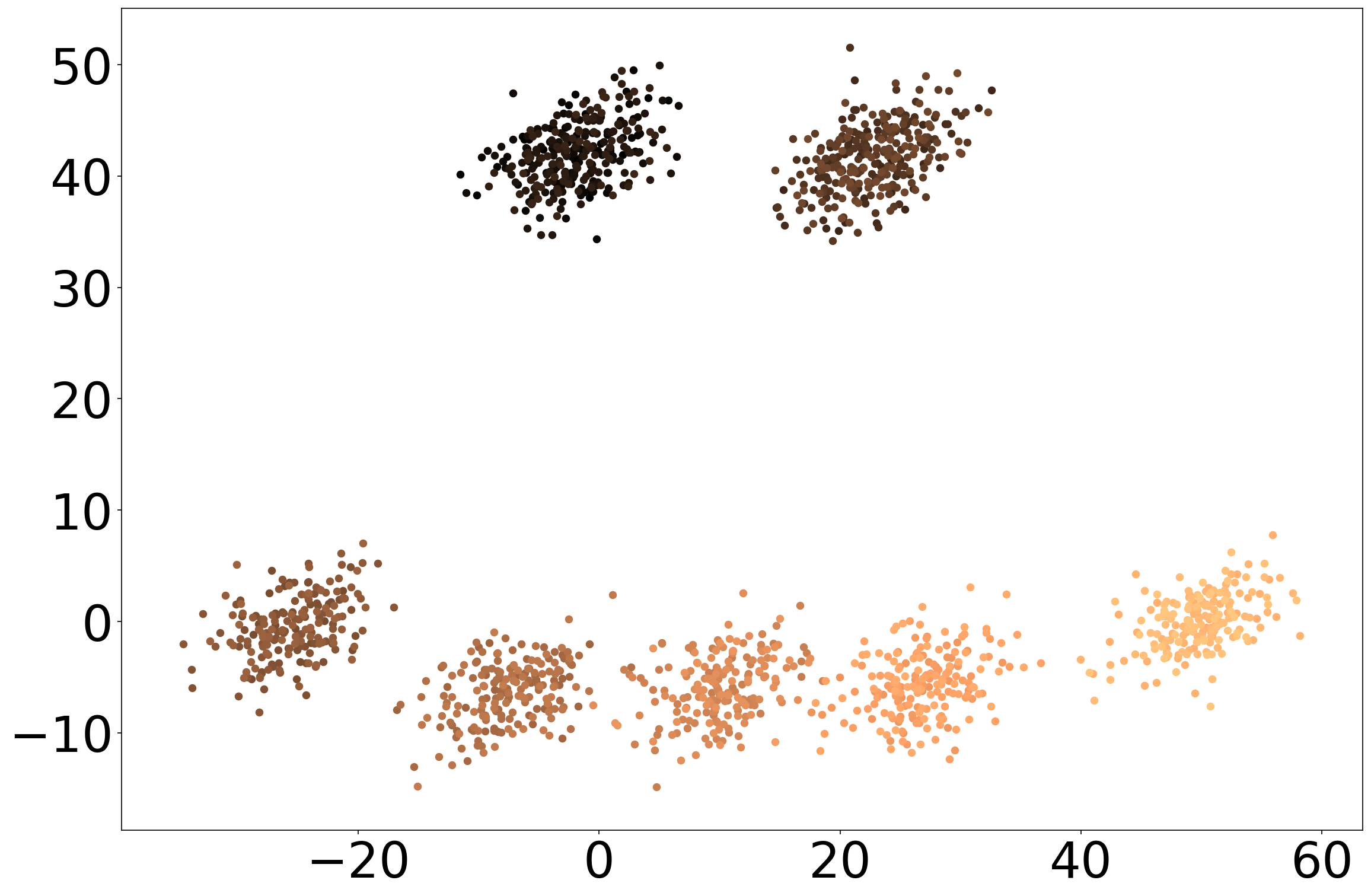}}
\hfill
\subcaptionbox{Mixed presentation}{\includegraphics[width=0.32\textwidth]{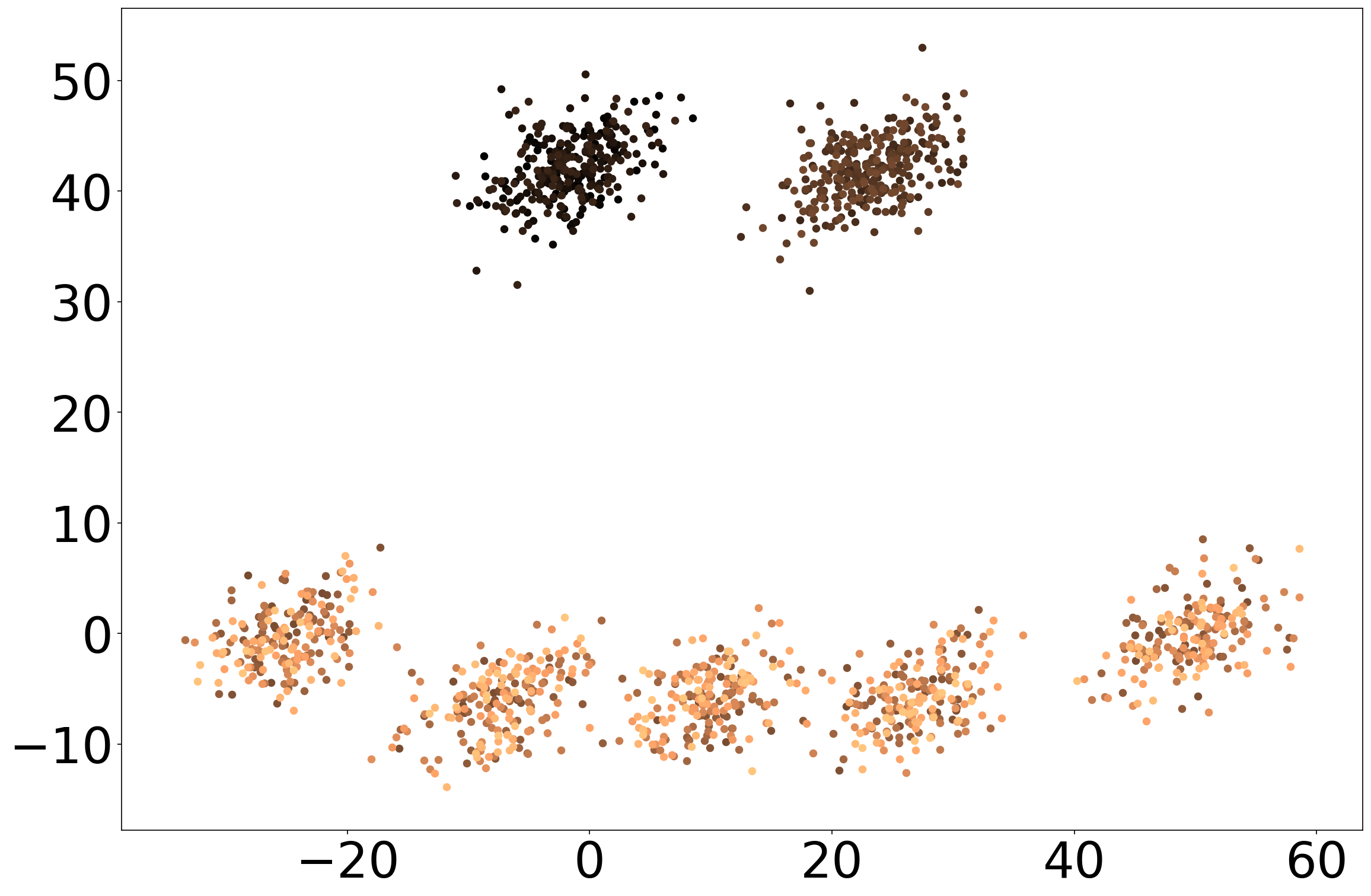}}
\hfill
\subcaptionbox{Random presentation}{\includegraphics[width=0.32\textwidth]{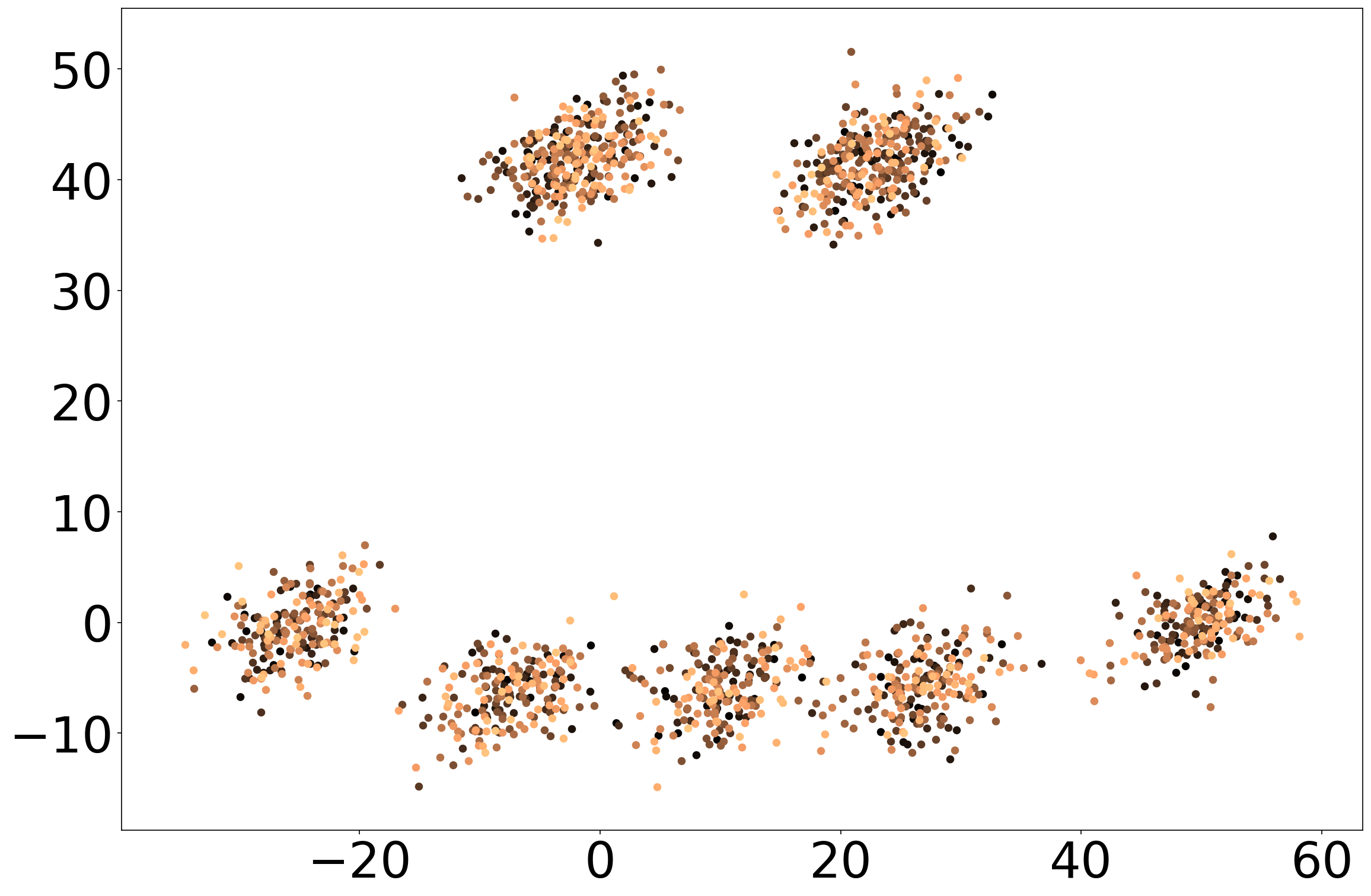}}
\caption{Synthetic data set. Unsupervised experiments differ on ordering: (a) corresponds to a cluster-by-cluster presentation, (b) corresponds to a mixed presentation (cluster-by-cluster for the top two clusters and random for the bottom five clusters), and (c) corresponds to a completely shuffled presentation. The darker the sample the earlier it was presented in the respective experiment.}
\label{Fig:exps_data_synthetic}
\end{figure}

Input ordering tends to have a dramatic effect on the performance of online agglomerative clustering algorithms such as ART~\cite{leonardo2018}. Therefore, the unsupervised experiments were conducted in three manners that differ regarding the order in which samples were presented: (i) cluster-by-cluster presentation (an unsupervised version of class-incremental~\cite{Rebuffi.2017a, Castro.2018a} presentation), (ii) mixed presentation and (iii) random presentation. The data set and color-coded sample ordering for each experiment are depicted in Fig.~\ref{Fig:exps_data_synthetic}. The cluster-by-cluster (i)\footnotemark[\value{footnote}] and mixed (ii)\footnotemark[\value{footnote}] orderings are akin to~\cite{Milad.2019a}, whereas in the random presentation (iii) experiment conducted here, as opposed to (ii), the entire data set was shuffled rather than only the samples belonging to the five bottom clusters shown in Fig.~\ref{Fig:exps_data_synthetic}. Note that the ordering (iii) corresponds to a stationary data distribution, whereas orderings (i) and (ii) correspond to non-stationary. 

The supervised prediction / semi-supervised learning experiment was conducted with shuffled presentation: a single sample from each cluster was used for training, and the remaining samples were used for testing (and also semi-supervised learning in the case of iCVI-TopoARTMAP).

The evaluation protocol consisted of presenting --- at the end of the learning --- the entire data for each method, recording their predictions, and then measuring their performance. The algorithms (Section~\ref{sec:algs_params}) were evaluated using either the adjusted rand index ($ARI$)~\cite{hubert1985} or classification accuracy ($ACC$), which corresponds to the tasks of online clustering and classification, respectively.

\subsection{Algorithms} \label{sec:algs_params}

In this study, the iCVI-TopoARTMAP in pure unsupervised learning mode was compared to sequential k-means (skm)~\cite{skm}, the iXB\textsubscript{$\lambda$}-based (i.e., iXB with forgetting factor $\lambda$) method of incremental sequential k-means (iskm)~\cite{Milad.2019a} as well as the following ART-based clustering algorithms: dual vigilance fuzzy ART (DVFA)~\cite{leonardo.2018b}, topological fuzzy ART (TopoFA)~\cite{Tscherepanow2010} --- only a single module of TopoFA was used, and developmental resonance network (DRN)~\cite{Park.2019a}.

Note that, except for DRN, none of the previously mentioned ART-based models can handle use-cases in which the data ranges are unknown, i.e., they require a priori knowledge of the data maximum and minimum statistics to perform min-max normalization and subsequent complement-coding~\cite{Swope2012, Tscherepanow2012a, Meng.2015a, Meng.2019a, Park.2019a, leonardo.2019b}. Therefore, we equip these ART-based models with online normalization of inputs and weight vector re-scaling~\cite{Swope2012, Meng.2015a, Meng.2019a} (Section~\ref{sec:scaling}). This would permit these algorithms to be applicable in such use-cases and thus allow for benchmark comparisons. To indicate such difference between these variants and their original versions, the prefix ``\underline{w}eight re-\underline{s}caled'' (WS) was added to these models, which are henceforth referred to as weight re-scaled dual vigilance fuzzy ART (WS-DVFA) and weight re-scaled topological fuzzy ART (WS-TopoFA). 

The iCVI-TopoARTMAP in semi-supervised mode was compared to the nearest neighbor (NN) rule~\cite{duda2000} using Euclidean distance. The latter is commonly used in face recognition tasks~\cite{Wang.2018a}, which is the real-world data used in this work, and thus NN was chosen for consistency across experiments.

The hyper-parameter tuning efforts are detailed in Appendix~\ref{Sec:parameterization_1}. 

\subsection{Implementation and reproducibility} \label{sec:the_code}

The experiments were almost exclusively carried out using python. The iCVI-TopoARTMAP, WS-DVFA, WS-TopoFA, NN, and skm python code are provided at Guise AI's GitHub repository\footnote{available for academic purposes only at \url{https://github.com/GoGetter-Inc/iCVI-TopoARTMAP}}. Some of the iCVI-TopoARTMAP components are based on the \textit{iCVI-toolbox for Matlab}~\cite{leonardo.2020a} available at the Applied Computational Intelligence Laboratory's GitHub repository\footnote{available at \url{https://github.com/ACIL-Group/iCVI-toolbox}}. The code for the iskm was run on Octave and is from the \textit{iskmeans} GitHub repository\footnote{available at \url{https://github.com/iskmeans/iskm}}~\cite{Milad.2019a}. The DRN method is from \textit{Incremental-Learning} GitHub repository\footnote{available at \url{https://github.com/Uehwan/Incremental-Learning}}~\cite{Yoon.2019a, yoon2019stabilized}. The ARI computations and min-max normalizations were made using \textit{scikit-learn}\footnote{available at \url{https://scikit-learn.org}}~\cite{scikit-learn}. 

\subsection{Results and discussion} \label{sec:results}

Table~\ref{Tab:results_synthetic} reports the results obtained from the unsupervised learning experiments (online clustering) across the different orderings discussed in Section~\ref{sec:data_1_exps} while following the parameter tuning described in Appendix~\ref{Sec:parameterization_1}. Specifically, Table~\ref{Tab:results_synthetic} shows that at least 4 out of the 6 iCVI-TopoARTMAP variants outperformed all other ART-based methods in each individual experiment: 5 out 6 in class-incremental presentation, 6 out 6 in mixed presentations, and 4 out of 6 in random presentation. With the exception of the iCONN-TopoARTMAP performance in random presentation, the other variants (iDB- and iXB-based) achieved comparable performance. 

\begin{table}[!t]
\centering
\caption{Results of the unsupervised learning experiments. The best performance is reported in bold and the second-best performance is in italics. $P$ and $\hat{k}$ represent the number of prototypes (or categories) and the estimated number of clusters.}
\begin{tabular}{lrrrrrrrrr}
\toprule
\multirow{2}[4]{*}{\textbf{model}} & \multicolumn{3}{c}{\textbf{class-incremental order}} & \multicolumn{3}{c}{\textbf{mixed order}} & \multicolumn{3}{c}{\textbf{random order}} \\
\cmidrule(lr){2-4} 
\cmidrule(lr){5-7}
\cmidrule(lr){8-10}
& $ARI$ & \textbf{$\hat{k}$} & \textbf{$P$} & $ARI$ & \textbf{$\hat{k}$} & \textbf{$P$} & $ARI$ & \textbf{$\hat{k}$} & \textbf{$P$} \\
\midrule
\midrule
skm~\cite{skm} 
& 0.2474 & 7     & 7     
& 0.2985 & 7     & 7     
& 0.9890 & 7     & 7 \\
iskm (uses iXB\textsubscript{$\lambda$})~\cite{Milad.2019a}
& 0.9878 & 7     & 7     
& \textbf{0.9975} & 7     & 7     
& 0.0000 & 1     & 1 \\
\midrule
DRN~\cite{Park.2019a}	
& 0.8463 & 13	 & 13
& 0.7705 & 9	 & 9	
& 0.7458 & 17	 & 17 \\
\midrule
WS-DVFA 
& 0.5989 & 21    & 21    
& 0.7423 & 15    & 235   
& 0.8770 & 8     & 137 \\
WS-TopoFA 
& 0.8531 & 9     & 73    
& 0.6183 & 10    & 62    
& 0.7846 & 5     & 26 \\
\midrule
iCH-TopoFAM 
& \textbf{0.9926} & 7     & 11    
& 0.9963 & 7     & 29    
& 0.9890 & 7     & 27 \\
iWB-TopoFAM 
& 0.9902 & 7     & 19    
& \textbf{0.9975} & 7     & 32    
& \textbf{0.9902} & 7     & 33 \\
iPBM-TopoFAM 
& \textit{0.9914} & 7     & 13    
& \textit{0.9969} & 8     & 28    
& \textit{0.9902} & 9     & 49 \\
iXB-TopoFAM 
& 0.9766 & 9     & 37    
& 0.9842 & 7     & 59    
& 0.8561 & 6     & 59 \\
iDB-TopoFAM 
& 0.8375 & 158   & 232   
& 0.8519 & 40    & 141   
& 0.8797 & 6     & 42 \\
iCONN-TopoFAM 
& 0.9576 & 10    & 50    
& 0.9611 & 11    & 39    
& 0.4054 & 6     & 31 \\
\bottomrule
\end{tabular}
\label{Tab:results_synthetic}
\end{table}
\begin{table}[!t]
\centering
\caption{Results of the supervised/semi-supervised learning experiment. The best and second-best performances are reported in bold and italics, respectively. $P$, $c$, and $n_{mis}$ represents the number of prototypes (or categories), the number of classes, and the number of misclassified samples.}
\begin{tabular}{lrrrrrr}
\toprule
\multirow{2}[2]{*}{\textbf{model}} & \multicolumn{2}{c}{\textbf{train}} & \multicolumn{2}{c}{\textbf{test}} & \multirow{2}[2]{*}{$c$} & \multirow{2}[2]{*}{$P$} \\
\cmidrule(lr){2-3} 
\cmidrule(lr){4-5}
& $ACC$ & $n_{mis}$ & $ACC$ & $n_{mis}$ &       &  \\
\midrule
\midrule
NN~\cite{duda2000}   
& 1.0000 & 0     & 0.9768 & 37    & 7     & 7 \\
iCH-TopoFAM 
& 1.0000 & 0     & \textbf{0.9937} & 10    & 7     & 8 \\
iWB-TopoFAM 
& 1.0000 & 0     & \textbf{0.9937} & 10    & 7     & 8 \\
iPBM-TopoFAM 
& 1.0000 & 0     & \textbf{0.9937} & 10    & 7     & 8 \\
iXB-TopoFAM 
& 1.0000 & 0     & 0.9617 & 61    & 7     & 20 \\
iDB-TopoFAM 
& 1.0000 & 0     & \textit{0.9931} & 11    & 7     & 11 \\
\bottomrule
\end{tabular}
\label{Tab:results_synthetic_ssl}
\end{table}

Results also show that iskm performs poorly on random presentation whereas skm, as expected~\cite{Moshtaghi2018, Moshtaghi2018b, Milad.2019a}, performs poorly on class-incremental presentation (i.e., change-point detection~\cite{Ibrahim2018b}). Conversely, most iCVI-TopoARTMAP variants show consistent performance across all orderings, thus either outperforming skm and iskm or yielding comparable performance across the experiments. The partitions associated with these results are shown in Appendix~\ref{sec:results_figs} for the different orderings.

Table~\ref{Tab:results_synthetic_ssl} reports the results comparing the prediction of the NN classifier and iCVI-TopoARTMAP in semi-supervised mode (online classification), as per the random ordering and parameter tuning described in Section~\ref{sec:data_1_exps} and Appendix~\ref{Sec:parameterization_1}, respectively. Note that while the NN only predicts, the iCVI-TopoARTMAP keeps learning with each incoming sample. Table~\ref{Tab:results_synthetic_ssl} shows that almost all iCVI-TopoARTMAP variants outperform NN; furthermore, 4 out of the 5 iCVI-TopoARTMAP variants mis-classify almost 4 times less samples than NN. The exception was iXB-TopoARTMAP, which still yields a comparable performance.

The common trade-off for the superior performance of iCVI-TopoARTMAP is a less compact but still competitive model (i.e., the larger number of categories listed in Tables~\ref{Tab:results_synthetic} and~\ref{Tab:results_synthetic_ssl}) and the additional computational cost associated with the computation of the iCVIs as well as the post-processing strategies for each sample presentation. This is particularly noticeable on the iDB-TopoARTMAP variant which clearly shows a tendency for over-partition and category proliferation; the latter finding is aligned with the one in~\cite{leonardo2017}.

\subsubsection{Robustness to vigilance parameter setting} \label{sec:robustness_rho}

The vigilance parameter is typically critical to the performance of ART-based models~\cite{leonardo.2019b}. Therefore, this section investigates the online learning outcome of the ART-based models in terms of accuracy, number of clusters, and number of categories, while varying their main vigilance parameter. Specifically, as per Appendix~\ref{Sec:parameterization_1}, the vigilance parameters were swept while holding constant the remaining parameters associated with the results reported in Table~\ref{Tab:results_synthetic}.

Figure~\ref{Fig:vigilance_robustness} shows that, compared to the other ART-based models, the iCVI-TopoARTMAP variants maintain a consistent outcome for a wide range of vigilance parameters. Moreover, the performance levels of most iCVI-TopoARTMAP variants are superior to the other ART-based models across several vigilance values, particularly the iCH-, iWB-, and iPBM-based models. As indicated in Section~\ref{sec:results}, the number of categories created by the iCVI-TopoARTMAP variants is usually larger than the other ART-based models. 

\newcommand{\rb}{.32}
\begin{figure}[!t]
\centering
\includegraphics[width=0.95\textwidth]{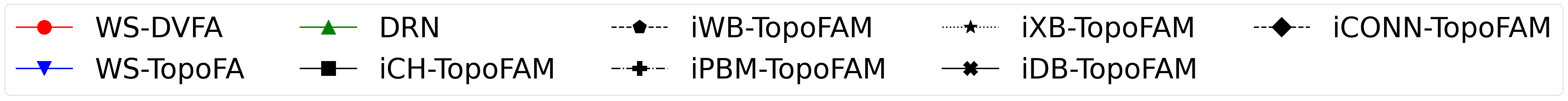}
\hfill

\subcaptionbox{ARI}{\includegraphics[width=\rb\textwidth]{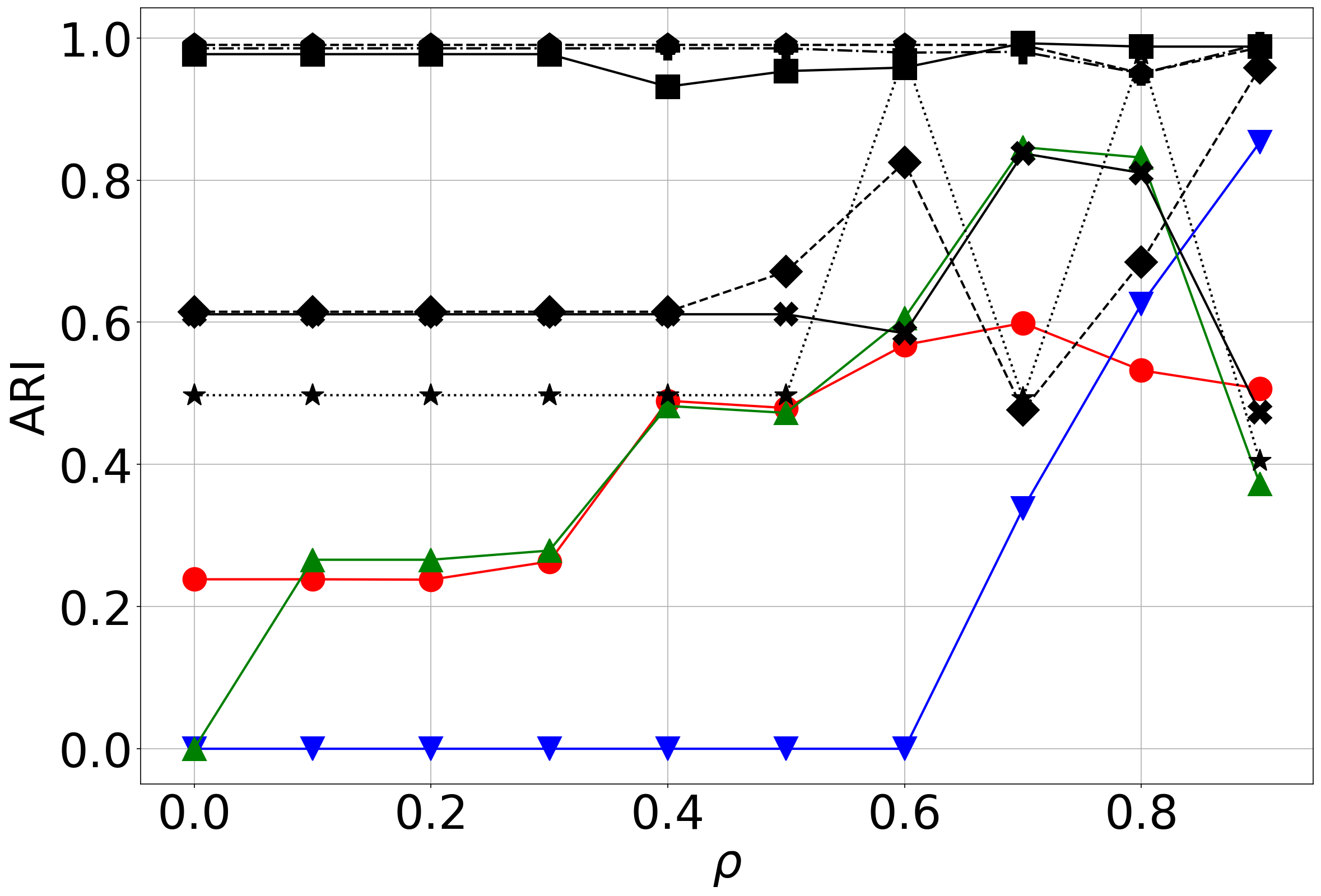}} 
\hfill
\subcaptionbox{Categories}{\includegraphics[width=\rb\textwidth]{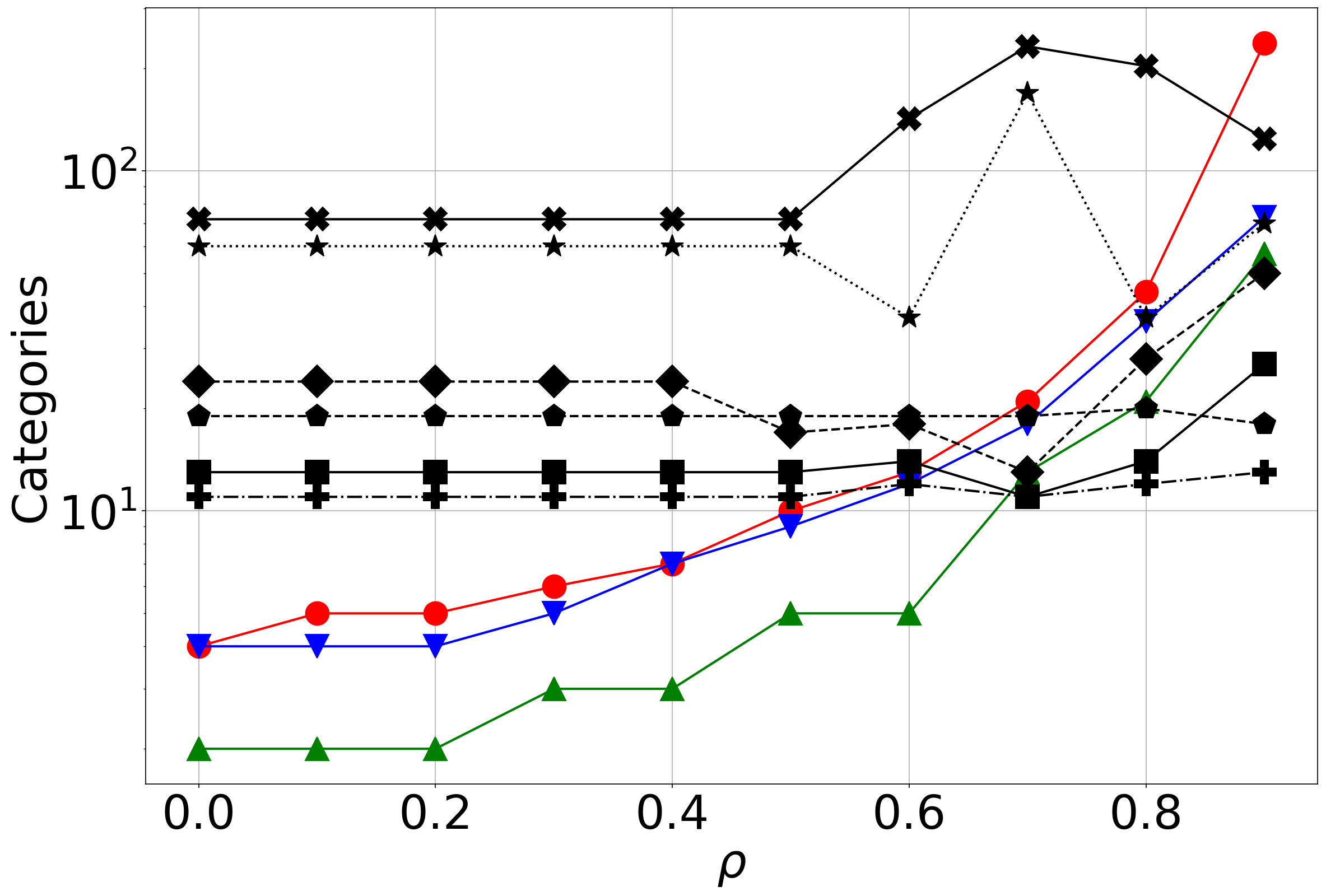}}
\hfill
\subcaptionbox{Clusters}{\includegraphics[width=\rb\textwidth]{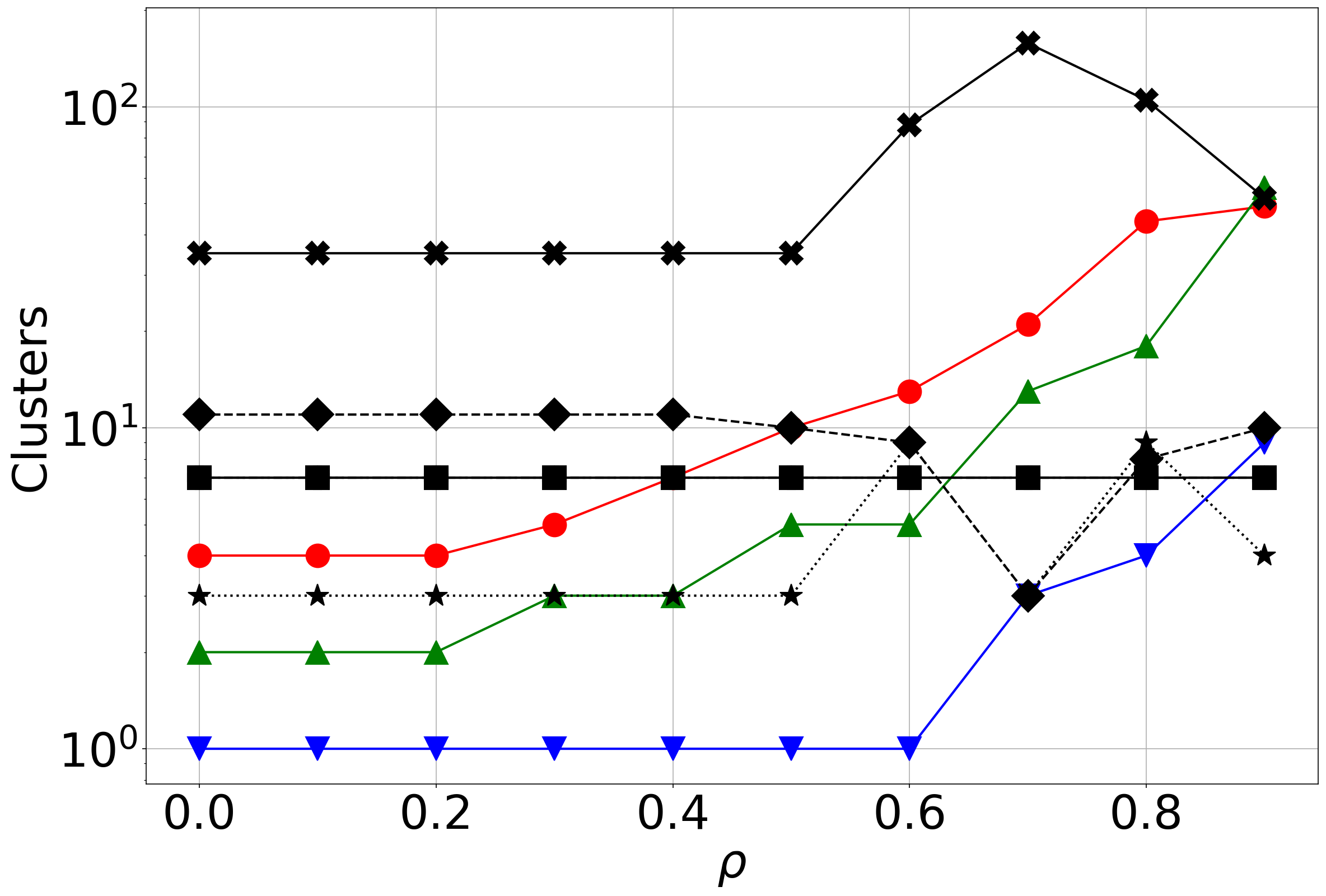}}

\subcaptionbox{ARI}{\includegraphics[width=\rb\textwidth]{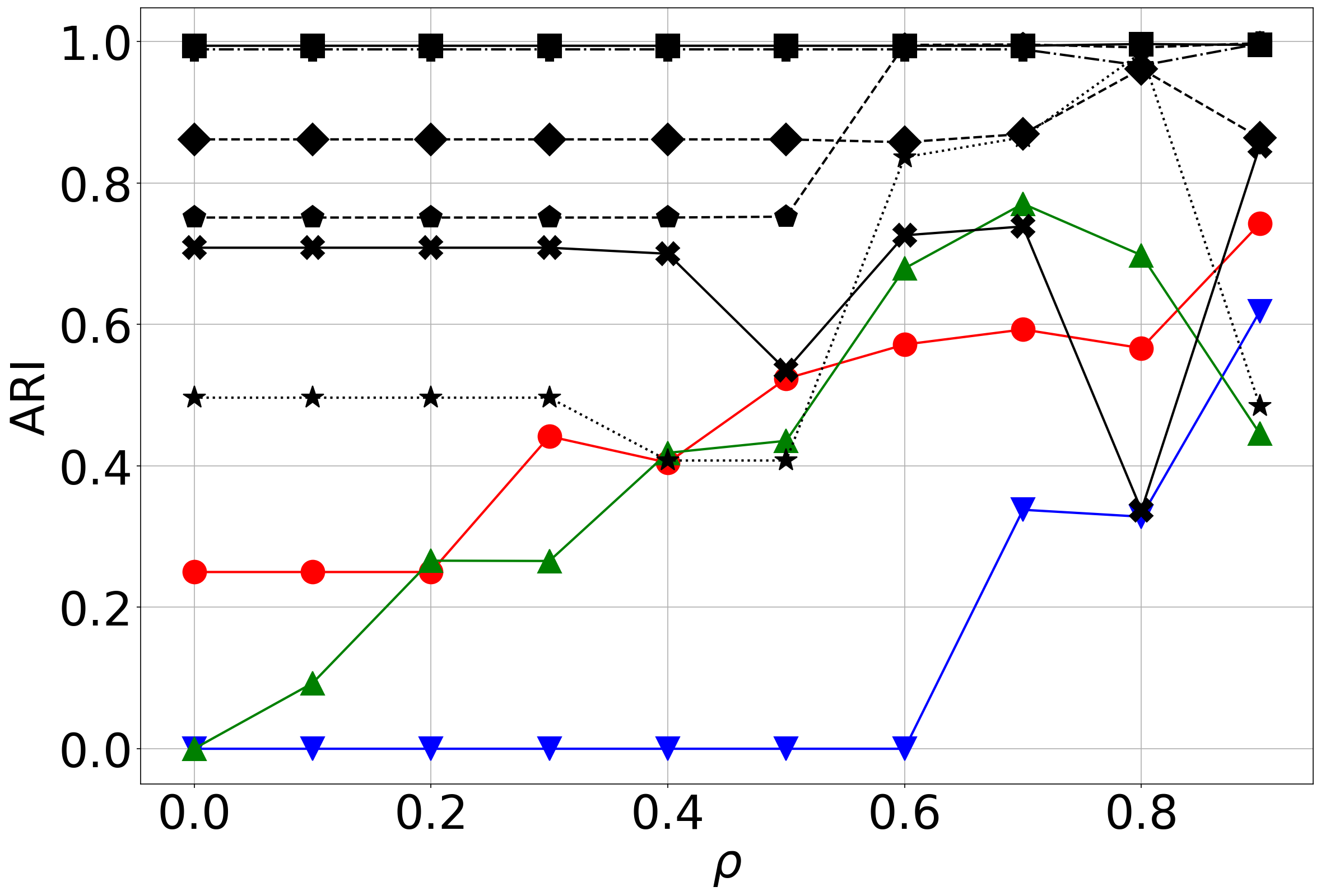}} 
\hfill
\subcaptionbox{Categories}{\includegraphics[width=\rb\textwidth]{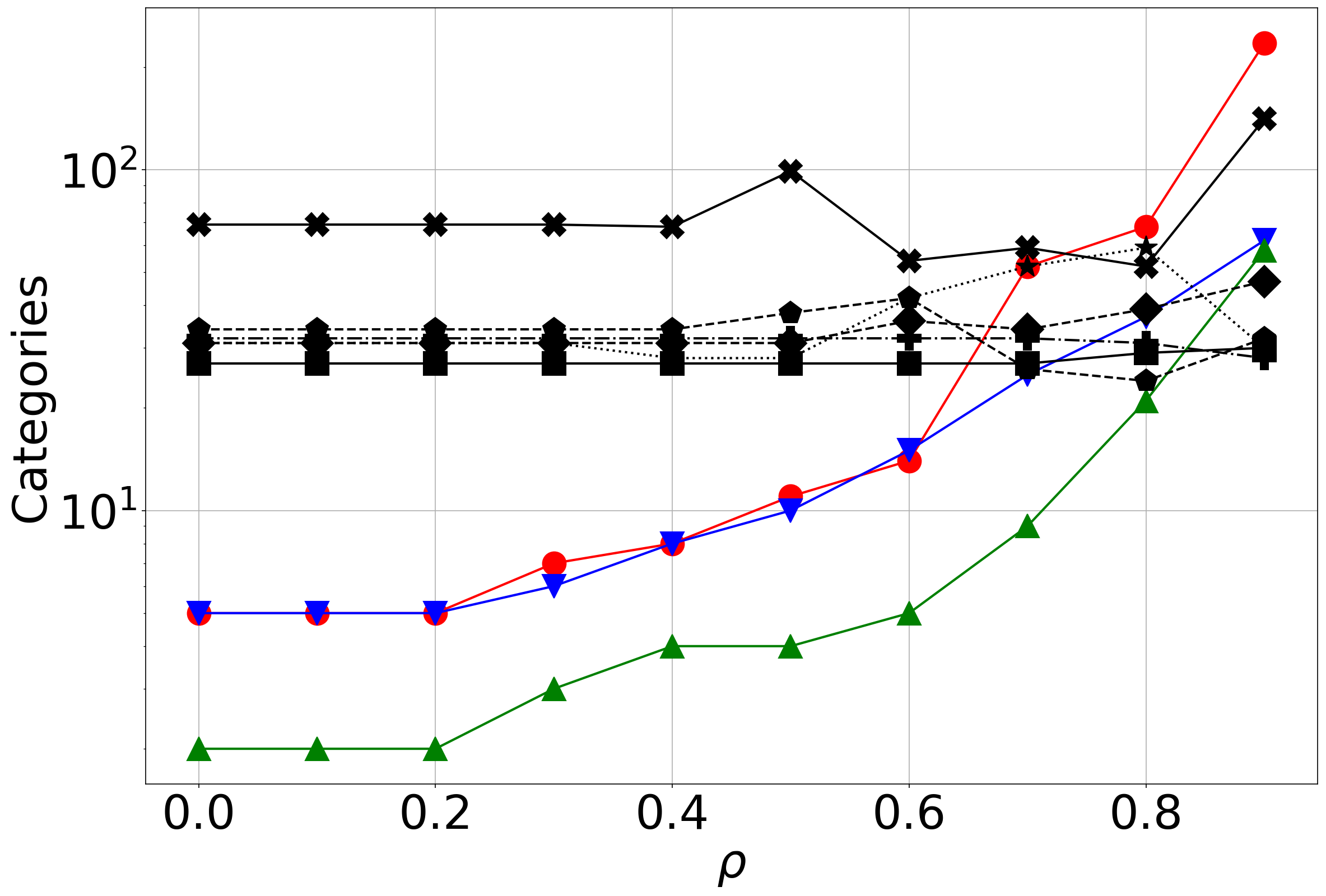}}
\hfill
\subcaptionbox{Clusters}{\includegraphics[width=\rb\textwidth]{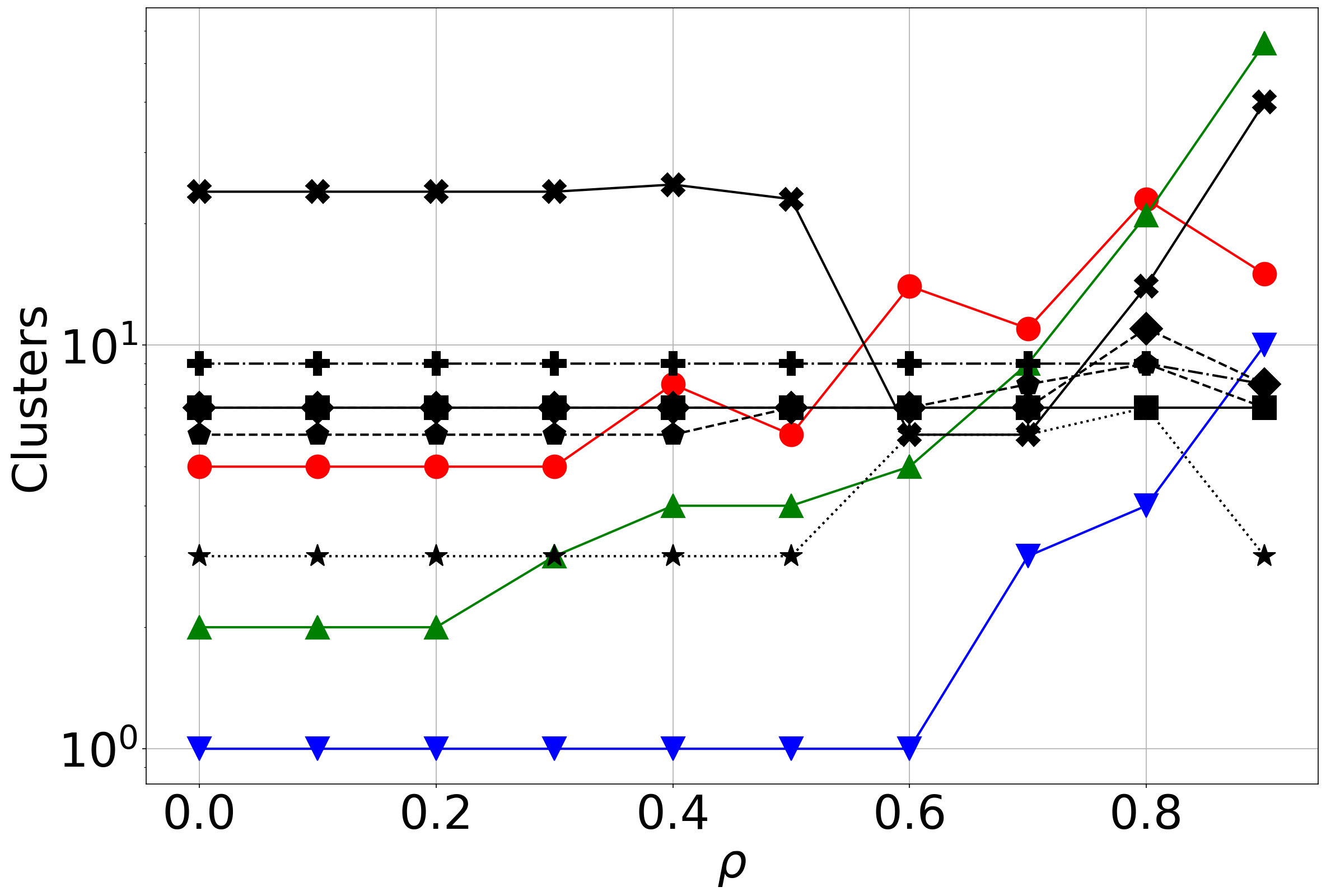}}

\subcaptionbox{ARI}{\includegraphics[width=\rb\textwidth]{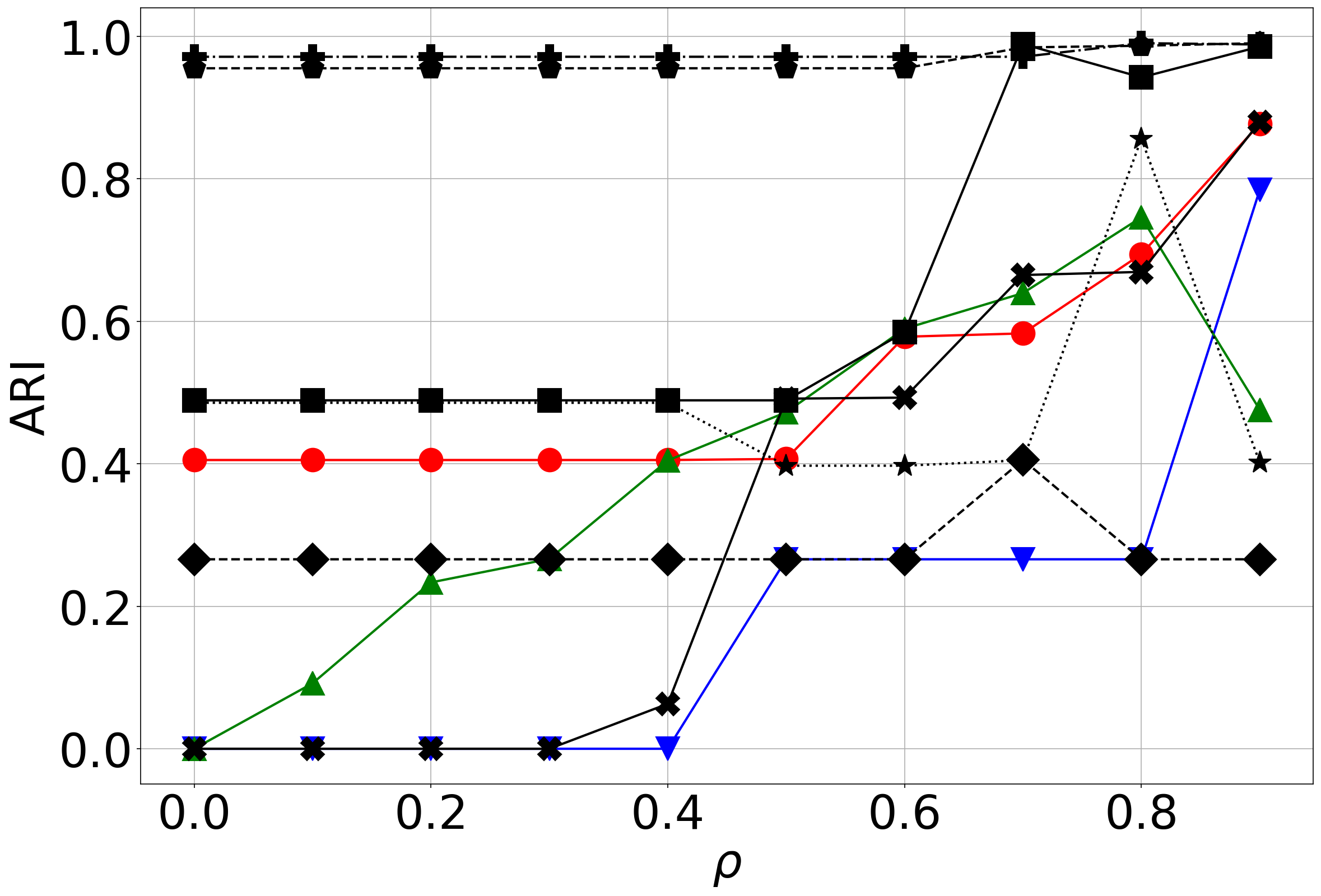}} 
\hfill
\subcaptionbox{Categories}{\includegraphics[width=\rb\textwidth]{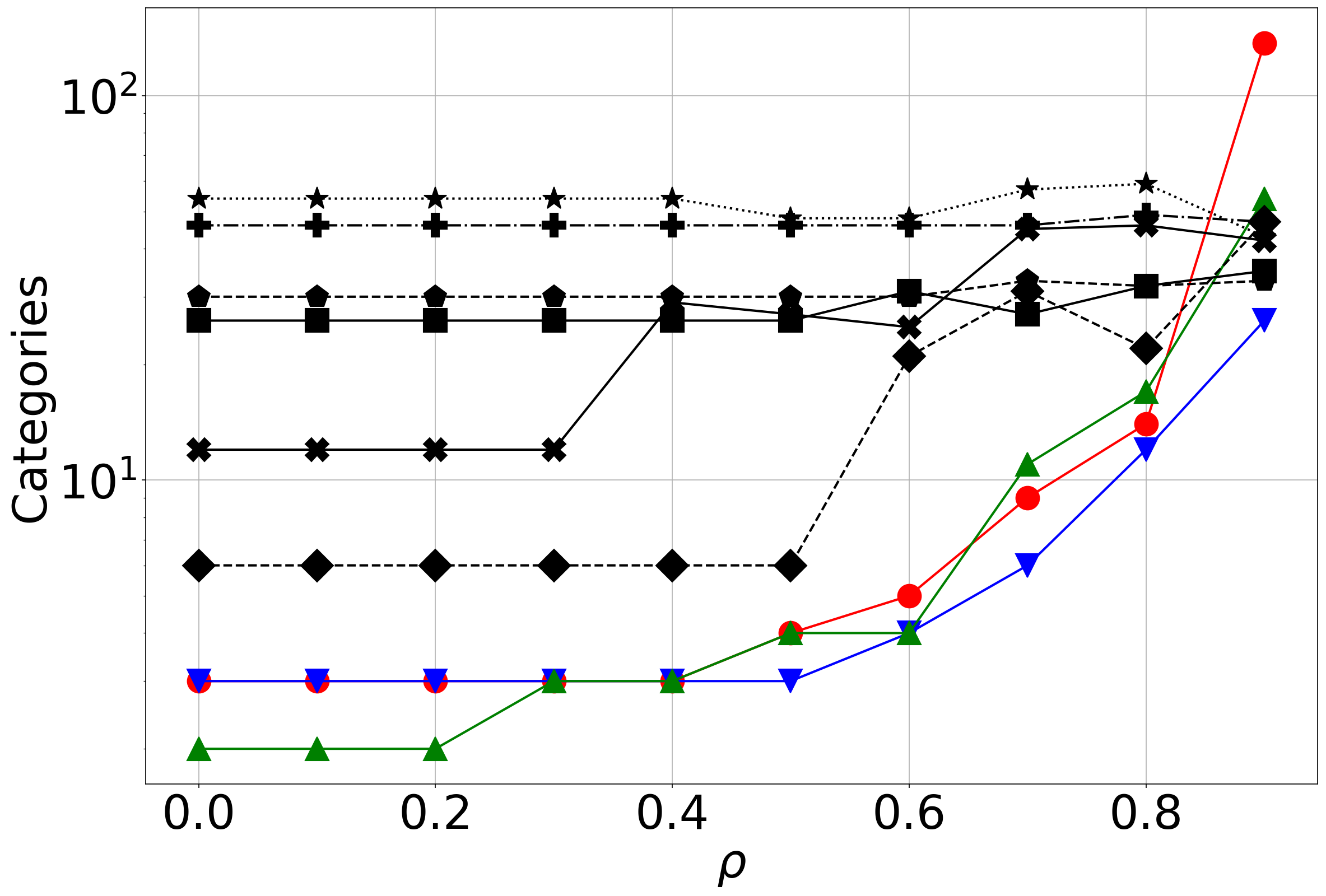}}
\hfill
\subcaptionbox{Clusters}{\includegraphics[width=\rb\textwidth]{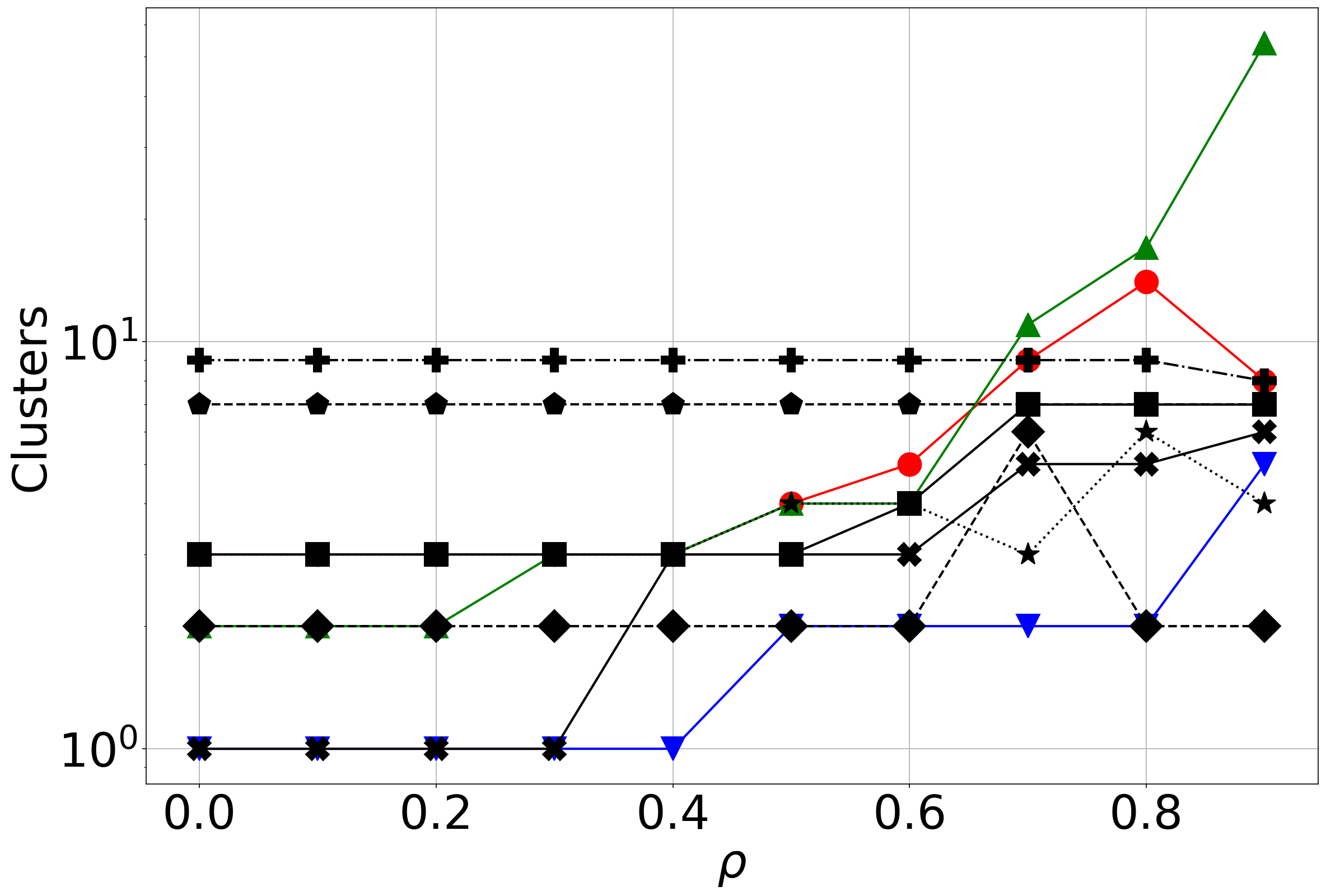}}
\caption{Accuracy (in terms of ARI), the number of categories and number of clusters yielded by the ART-based models as a function of their main vigilance parameter for the class-incremental ((a)-(c)), mixed ((d)-(f)) and random ((g)-(i)) presentation experiments.}
\label{Fig:vigilance_robustness}
\end{figure}

\section{Experiments on real-world data} \label{Sec:exp_rw_data}

\subsection{Data set and experimental protocol} \label{Sec:data_rw}

\begin{figure}[!t]
\centering
\includegraphics[width=\textwidth]{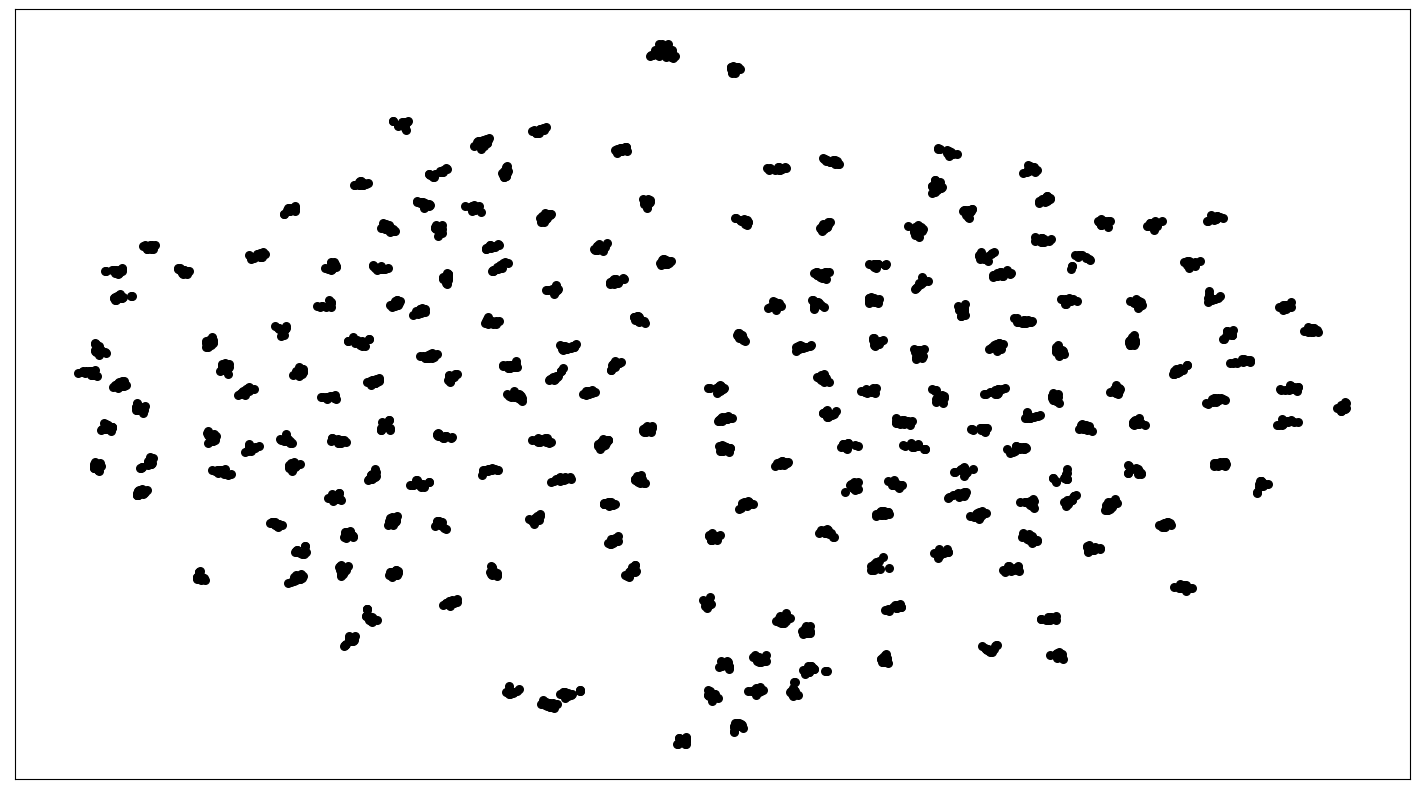}
\caption{Real-world data set. The FEI~\cite{Thomaz.2010a} faces detected with MTCNN~\cite{Zhang.2016a} are shown using the t-SNE~\cite{Maaten.2008a} projection of their respective embeddings generated via VGGFace2~\cite{Cao.2018a}.}
\label{Fig:data_rw}
\end{figure}

The FEI face image data set\footnote{available at \url{https://fei.edu.br/~cet/facedatabase.html}}~\cite{Thomaz.2010a} was used for benchmark-only purposes within the scope of this research paper. It consists of the faces of $200$ individuals ($14$ images per individual, thus making a grand total of $2800$ images) taken under a controlled environment with different illumination settings. Individuals vary in age, poses, and facial expressions (Fig.~\ref{Fig:data_rw}). 

\begin{figure}[!t]
\centering
\includegraphics[width=0.99\textwidth]{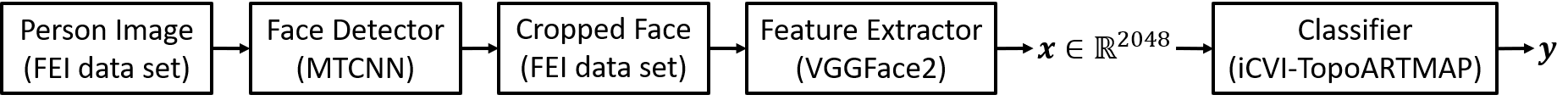}
\caption{The pipeline used in the experiments with the FEI data set~\cite{Thomaz.2010a}: MTCNN model~\cite{Zhang.2016a} for face detection, VGGFace2 model~\cite{Cao.2018a} to extract face descriptors, and iCVI-TopoARTMAP model as the unsupervised (or semi-supervised) classifier. The competing models listed on Section~\ref{Sec:algs_data_rw} replaced iCVI-TopoARTMAP in this pipeline to compute their performance.}
\label{Fig:rw_pipeline}
\end{figure}

Recently, ART-based systems have been applied to process embeddings generated by deep neural networks, such as clustering in~\cite{leonardo.2020c} and object detection in~\cite{Brna.2019a}. This work uses a standard machine learning pipeline for face recognition~\cite{Prihasto.2016a, Cao.2018a, Wang.2018a, Elmahmudi.2019a}, which, in its most succinct form, consists of a face detector, a feature extractor, and a classifier (Fig.~\ref{Fig:rw_pipeline}). The pre-trained MTCNN~\cite{Zhang.2016a} and pre-trained VGGFace2~\cite{Cao.2018a} (with SE-ResNet-50~\cite{Hu.2018a} architecture) were used as the detector and extractor models, respectively. As opposed to the previous iteration~\cite{Parkhi.2015a}, VGGFace2 provides a $2048$-dimensional feature vector for a given face. We extended the area of the bounding boxes by $30$\% like~\cite{Cao.2018a} and set the confidence threshold to $0.99$. However, unlike~\cite{Cao.2018a}, $\ell_2$ normalization is not employed. Note the face detector failed in detecting some faces within the data set; these were then disregarded. Therefore, the total number of faces was $2774$.

The unsupervised experiments with real-world data were conducted in two manners that differ with respect to the order in which samples are presented: (i) cluster-by-cluster presentation (i.e., unsupervised class-incremental), and (ii) random presentation. The supervised/semi-supervised experiments as well as the evaluation protocol were performed as described in Section~\ref{sec:data_1_exps}.

\subsection{Algorithms} \label{Sec:algs_data_rw}

The experiments with the real-world data set employed the same algorithms listed in Section~\ref{sec:algs_params}. However, the NN classifier used cosine distance rather than Euclidean distance. The parameterization is detailed in Appendix~\ref{Sec:parameterization_2}.

\subsection{Implementation and reproducibility} \label{sec:the_code_rw}

In addition to the implementations listed in Section~\ref{sec:the_code}, the experiments in this section made use of the VGGFace2~\cite{Cao.2018a} pre-trained model from the \textit{keras-vggface} GitHub repository\footnote{available at \url{https://github.com/rcmalli/keras-vggface}} and the MTCNN~\cite{Zhang.2016a} pre-trained model from the \textit{MTCNN} GitHub repository\footnote{available at \url{https://github.com/ipazc/mtcnn}}.

\subsection{Results and discussion}  \label{sec:results_rw}

Table~\ref{Tab:results_rw_1} reports the results obtained from the unsupervised learning experiments (online clustering) across the different orderings discussed in Section~\ref{Sec:data_rw} while following the parameter tuning described in Appendix~\ref{Sec:parameterization_2}. Specifically, Table~\ref{Tab:results_synthetic} shows that E-TopoFA (module A of iCVI-TopoARTMAP), by itself, already considerably outperforms skm, iskm and all other ART-based methods, thus showcasing the importance of the global cosine distance-based vigilance for this problem --- note that E-TopoFA predictions do not consider connections between categories (i.e., the $CONN$ matrix). All iCVI-TopoARTMAP variants also outperformed the competing methods in both experiments. Moreover, the vast majority of the iCVI-TopoARTMAP variants further improved performance compared to E-TopoFA, with the exceptions of iDB- and iXB-TopoARTMAP for the class-incremental and random presentations, respectively.  

Notably, these results also corroborate the findings from Section~\ref{sec:results}: skm yields a poorer performance on the class-incremental presentation of samples (as expected~\cite{Moshtaghi2018, Moshtaghi2018b, Milad.2019a}), and iskm, conversely, yields a poorer performance on a random presentation of samples. Again, the iCVI-TopoARTMAP variants showed consistent performance across both orderings. Interestingly, the iXB-TopoARTMAP suffered from category proliferation and over-partitioning in the random ordering experiment, whereas the iDB-TopoARTMAP continued to showcase these issues in all experiments. 

Table~\ref{Tab:results_rw_2} reports the results for the online face classification using the pipeline shown in Fig.~\ref{Fig:rw_pipeline} as per the ordering and parameter tuning described in Section~\ref{Sec:data_rw} and Appendix~\ref{Sec:parameterization_2}, respectively; wherein this research compares the predictions of an NN classifier (common among face recognition methods that use deep feature descriptors~\cite{Wang.2018a}) and the iCVI-TopoARTMAP in semi-supervised mode (which continuously learns using the embeddings). External knowledge was infused into the network in the form of ``registration'' wherein a single embedding of each individual was randomly chosen to define the prototypes used in the NN classifier and to sequentially initialize iCVI-TopoARTMAP. Table~\ref{Tab:results_rw_2} shows that iCH- and iWB-TopoARTMAP variants modestly outperformed NN (misclassifying around a quarter less samples), while the iDB-TopoARTMAP yielded comparable performance. 

As mentioned in Section~\ref{sec:results}, the superior performance of iCVI-TopoARTMAP is a less compact --- but still competitive --- model that requires additional computational cost due to, for instance, the iCVIs computations. This was particularly noticeable on the iDB- and iXB-TopoARTMAP variants which, as previously mentioned, showed a tendency for category proliferation and over-partitioning. 

\begin{table}[!ht]
\centering
\caption{Results of the unsupervised learning experiments. The best performance is reported in bold, and the second-best performance is in italics. $P$ and $\hat{k}$ represent the number of prototypes (or categories) and the estimated number of clusters.}
\begin{tabular}{lrrrrrr}
\toprule
\multirow{2}[2]{*}{\textbf{model}} 
& \multicolumn{3}{c}{\textbf{class-incremental order}} 
& \multicolumn{3}{c}{\textbf{random order}} \\
\cmidrule(lr){2-4} 
\cmidrule(lr){5-7}
& $ARI$ & $\hat{k}$ & $P$ & $ARI$ & $\hat{k}$ & $P$ \\
\midrule
\midrule
skm~\cite{skm}   
& 0.0413 & 200   & 200   
& 0.4585 & 200   & 200 \\
iskm (uses iXB\textsubscript{$\lambda$})~\cite{Milad.2019a}
& 0.4025 & 240   & 240   
& 0.3487 & 210   & 210 \\
\midrule
DRN~\cite{Park.2019a}
& 0.9562 & 248	 & 248	
& 0.9045 & 303	 & 303 \\
\midrule
WS-DVFA  
& 0.8197 & 279   & 279   
& 0.8354 & 279   & 1194 \\
WS-TopoFA 
& 0.5282 & 665   & 825   
& 0.7366 & 358   & 778 \\
\midrule
E-TopoFA (module A) 
& 0.9782 & 239   & 239   
& 0.9535 & 289   & 289 \\
iCH-TopoFAM 
& 0.9901 & 204   & 239   
& \textit{0.9781} & 238   & 311 \\
iWB-TopoFAM 
& 0.9901 & 204   & 239   
& 0.9778 & 239   & 314 \\
iPBM-TopoFAM 
& \textit{0.9904} & 203   
& 272   & 0.9770 & 241   & 316 \\
iXB-TopoFAM 
& \textbf{0.9910} & 203   
& 352   & 0.9165 & 347   & 1152 \\
iDB-TopoFAM 
& 0.9591 & 243   & 920   
& 0.9543 & 287   & 304 \\
iCONN-TopoFAM
& 0.9901 & 204   & 239   
& \textbf{0.9818} & 228   & 297 \\
\bottomrule
\end{tabular}
\label{Tab:results_rw_1}
\end{table}

\begin{table}[!ht]
\centering
\caption{Results of the supervised/semi-supervised learning experiment. The best and second-best performances are reported in bold and italics, respectively. $P$, $c$, and $n_{mis}$ represents the number of prototypes (or categories), the number of classes, and the number of misclassified samples.}
\begin{tabular}{lrrrrrr}
\toprule
\multirow{2}[2]{*}{\textbf{model}} & \multicolumn{2}{c}{\textbf{train}} & \multicolumn{2}{c}{\textbf{test}} & \multirow{2}[2]{*}{$c$} & \multirow{2}[2]{*}{$P$} \\
\cmidrule(lr){2-3} 
\cmidrule(lr){4-5}
& $ACC$ & $n_{mis}$ & $ACC$ & $n_{mis}$ &       &  \\
\midrule
\midrule
NN~\cite{duda2000}   
& 1.0000 & 0     & \textit{0.9868} & 34    & 200   & 200 \\
iCH-TopoFAM 
& 1.0000 & 0     & \textbf{0.9899} & 26    & 200   & 212 \\
iWB-TopoFAM 
& 1.0000 & 0     & \textbf{0.9899} & 26    & 200   & 212 \\
iPBM-TopoFAM 
& 1.0000 & 0     & 0.9810 & 49    & 200   & 215 \\
iXB-TopoFAM 
& 1.0000 & 0     & 0.8578 & 366   & 200   & 527 \\
iDB-TopoFAM 
& 1.0000 & 0     & 0.9631 & 95    & 200   & 274 \\
\bottomrule
\end{tabular}
\label{Tab:results_rw_2}
\end{table}

\section{Order-dependence and order-indifference} \label{sec:ordering}

Order-dependence is widely known to be important in incremental learning (see \cite{leonardo2018, leonardo.2019b} and the references cited within), and thus, for many online agglomerative clustering algorithms, the order in which samples are presented can
lead to substantial performance variations. Naturally, ordering affects these algorithms differently, but for many of them, moving toward a class-incremental order typically tends to facilitate learning compared to random order. For instance, in scenarios in which the data ranges are known a priori, the incremental learners of FA, DVFA, and TopoFA tend to perform better when samples are presented in an orderly fashion (e.g., using VAT~\cite{bezdek2002, Bezdek2017}) and worse for random presentation~\cite{leonardo2018,leonardo.2018b, Elnabarawy.2019b, leonardo.2020b}.

In this work, as expected, we also observe that ordering effects play a significant role in the cluster structures detected by many of the clustering algorithms, which is ultimately reflected in their performance. In particular, the results reported on Tables~\ref{Tab:results_synthetic} and~\ref{Tab:results_rw_1} suggest that skm performs better during random order of presentation compared to class-incremental presentations, whereas the opposite effect was observed for iskm and DRN. The performances of WS-DVFA and WS-TopoFA were also affected; however, no conclusive relationship was observed. On the other hand, the experimental results using the same data sets indicated a high degree of order indifference for iCVI-TopoARTMAP, wherein most iCVI-TopoARTMAP variants yielded consistent performance across different orders of sample presentation --- the only exception was the iCONN-TopoARTMAP, whose performance decreased considerably for random order presentation in the experiments with the synthetic data set.

\section{Conclusion} \label{sec:the_conclusion}

This paper presents a novel incremental cluster validity index (iCVI) adaptive resonance theory predictive mapping (ARTMAP) model for online unsupervised and semi-supervised learning, namely the iCVI-TopoARTMAP. It is equipped with several features such as iCVI-based match tracking, online label generation driven by a user-selected iCVI, storage of additional local footprints (e.g., frequency, mean, compactness, and connectivity matrix), online normalization and complement coding of samples, online weight vector re-scaling, as well as heuristics for swapping categories between clusters, merging and splitting clusters, reducing the number of categories, and pruning and reassigning satellite categories. In addition, because of the map field component of the ARTMAP architecture, iCVI-TopoARTMAP is capable of incremental multi-prototype-based representation of clusters. 

In experiments with synthetic and real-world data sets, iCVI-TopoARTMAP in unsupervised mode yielded either superior (most cases) or comparable performance to state-of-the-art iCVI-based and ART-based clustering algorithms, even after augmenting some of the latter to enable their application in use-cases with unknown data ranges. Moreover, the experiments showed that while the performance of the competitor online clustering algorithms degraded to different degrees depending on the order of sample presentation (a dramatic decrease in performance was observed for some of these algorithms when faced with particular ordering instances), the iCVI-TopoARTMAP remained very resilient to ordering effects --- in fact, most of the iCVI-TopoARTMAP variants were order-indifferent). Similar performance results were also observed when comparing iCVI-TopoARTMAP in semi-supervised mode with a supervised nearest neighbor classifier. Finally, iCVI-TopoARTMAP was shown to be robust to parameter setting by its consistent performance across a wide range of vigilance parameter values in experiments with a synthetic data set.

\appendix

\section{Hyper-parameter tuning}

\subsection{Experiments on synthetic data} \label{Sec:parameterization_1}

For the ART-based methods we mainly focused on the vigilance parameter (which is usually critical for their performance~\cite{leonardo.2019b}). The details regarding the parameterization of the different methods are as follows (the notation $[a,b]@c$ indicates a grid search in the range $[a,b]$ with step-size $c$):
\begin{itemize}
\item skm: $k$ equal to the true number of clusters.
\item iskm: $\lambda=\lambda'=l$; $l \in [0.01,1.0]@0.01$.
\item WS-DVFA: $\alpha=0.001$, $\beta=1$, $\rho_{ub} \in [0,0.9]@0.1$, $\rho_{lb} \in [\max(0, \rho_{ub}-0.09),\rho_{ub}]@0.01$.
\item WS-TopoFA: $\alpha=0.001$, $\beta_1=1$, $\beta_2 \in [0, 0.6]@0.6$, $\phi=5$, $\tau \in [100, 600]@100$, $\rho \in [0,0.9]@0.1$.
\item DRN:  $\eta=1$, $\eta_g=1$, $\nu=2$, $\alpha=1$, $\rho \in [0,0.9]@0.1$.
\item iCVI-TopoARTMAP (unsupervised mode): iCVI $\in \{$iCH, iWB, iPBM, iXB, iDB, iconn\_index$\}$, $\phi=5$, $\rho_{MT_{icvi}}=0.9$, $\tau \in [0, 5]@5$, $\xi \in [100, 600]@100$, $\rho_a \in [0,0.9]@0.1$, $\rho_c=0$ for SS-based iCVIs and $\rho_c=\rho_a$ for iCONN-TopoARTMAP; the remaining parameters were set to their default values listed in Table~\ref{Tab:parameters}.
\item iCVI-TopoARTMAP (semi-supervised mode): iCVI $\in \{$iCH, iWB, iPBM, iXB, iDB$\}$, $\rho_{MT_{icvi}}=0.9$, $\xi=100$, $\rho_a=\rho_c=0$, $L_{type}=fixed$, $EN\_swap=EN\_merge=EN\_split=EN\_prune\_reassign=False$, $EN\_{T^u}=EN\_compress=EN\_MT_{iCVI}=True$; the remaining parameters were set to their default values listed in Table~\ref{Tab:parameters} (note that some of the parameters become irrelevant because some of the features of iCVI-TopoARTMAP were disabled). 
\end{itemize}

Naturally, with careful parameter tuning, improved performance is expected. Regarding iCVI-TopoARTMAP, in practical unsupervised applications, after selecting the iCVI, the practitioner only needs to effectively set $5$ parameters: $\rho_a$, $\rho_c$, $\rho_{MT_{icvi}}$, $\tau$, $\xi$. These can be further reduced to $4$: in these research experiments $\rho_c=0$ and $\rho_c=\rho_a$ seem to be reasonable values when selecting SS-based iCVIs and the iconn\_index, respectively. Setting $\rho_c=\rho_a$ basically adjusts for the scaling of the category size as the data range changes over time.

\subsection{Experiments on real-world data} \label{Sec:parameterization_2}

Most of the parameter tuning efforts employed for the real-world data experiments were similar to Appendix~\ref{Sec:parameterization_1}; the exceptions were: 
\begin{itemize}
\item iskm: $\lambda=\lambda'=l$; $l \in [0.18,0.99]@0.09$.
\item WS-TopoFA: $\alpha=0.001$, $\beta_1=1$, $\beta_2=0.6$, $\phi=0$, $\tau =100$, $\rho \in [0,0.9]@0.1$.
\item E-TopoFA: $\alpha=0.001$, $\beta_1=1$, $\beta_2=0.0$, $M_{type}=$Eq.~(\ref{eq:cosine_vigilance}), $\rho \in [0.1, 0.3]@0.1$.
\item iCVI-TopoARTMAP 
\begin{itemize}
\item unsupervised mode: iCVI $\in \{$iCH, iWB, iPBM, iXB, iDB, iconn\_index$\}$, $M_{type}=$ Eq.~(\ref{eq:cosine_vigilance}),  $\beta_2=0$ for SS-based iCVIs and $\beta_2=0.6$ for iCONN-TopoARTMAP, and
\begin{itemize}
\item Class-incremental presentation: $\rho_a \in [0.1, 0.3]@0.1$, $EN\_{T^u}=EN\_swap=EN\_prune\_reassign=True$, $EN\_merge=EN\_split=EN\_compress=EN\_MT_{iCVI}=False$, $\phi=3$, $\xi=100$.
\item Random presentation: $\rho_a \in [0.2, 0.3]@0.1$, $EN\_{T^u}=EN\_swap=True$, $EN\_prune\_reassign=EN\_merge=EN\_split=EN\_compress=EN\_MT_{iCVI}=False$.
\end{itemize}
\item semi-supervised mode: iCVI $\in \{$iCH, iWB, iPBM, iXB, iDB$\}$, $M_{type}=$ Eq.~(\ref{eq:cosine_vigilance}), $\rho_a=2$, $\rho_{MT_{icvi}}=0.1$, $\xi=300$, $L_{type}=fixed$, $EN\_swap=EN\_merge=EN\_split=EN\_prune\_reassign=False$, $EN\_{T^u}=EN\_compress=EN\_MT_{iCVI}=True$; 
\end{itemize} 
The remaining parameters of iCVI-TopoARTMAP (both modes) were set to their default values listed in Table~\ref{Tab:parameters}. Naturally, some parameters become irrelevant when their associated features are disabled. 
\end{itemize}

\section{Partitions, footprints, and tracking} \label{sec:results_figs}

This appendix includes figures illustrating the partitions of the synthetic data obtained by the online clustering algorithms as per the experimental results reported in Table~\ref{Tab:results_synthetic} (Figs.~\ref{Fig:partitions_synthetic_1}, \ref{Fig:partitions_synthetic_2}, and~\ref{Fig:partitions_synthetic_3}). In addition, Figs.~\ref{Fig:tracking_iCVI_TopoFAM_1}, \ref{Fig:tracking_iCVI_TopoFAM_2}, and~\ref{Fig:tracking_iCVI_TopoFAM_3} depict the iCVI value, iCVI checks, module A vigilance parameter ($\rho_a$), number of categories, number of clusters, and the performance of iCVI-TopoARTMAP variants over time --- like Section~\ref{sec:results}, the evaluation consists of re-presenting each sample up to time $t$, predicting their classes, and computing the ARI. All of these figures are associated with the simulations using the parameter setting that corresponds to the performance reported in Table~\ref{Tab:results_synthetic}.

The vigilance parameter ($\rho$) of an ART network constrains the maximum size of its categories~\cite{Carpenter1991}. As the range of the data set continuously expands with each incoming sample, the categories become larger for the same $\rho$ --- i.e., early generated categories are smaller than the ones generated later in time. This size variance phenomenon in ART-based methods is clearly observed in the class-incremental and mixed presentation order experiments depicted in Figs.~\ref{Fig:DVFA1}-\ref{Fig:TopoFA1} and~\ref{Fig:DVFA2}-\ref{Fig:TopoFA2}, where the categories representing the top two clusters (presented first) are much smaller than the ones representing the bottom five clusters (presented last). The compression operation of iCVI-TopoARTMAP (Section~\ref{sec:compress}) mitigates this issue (see Figs.~\ref{Fig:iCH1}-\ref{Fig:iCONN1} and~\ref{Fig:iCH2}-\ref{Fig:iCONN2}): for instance, the early detected clusters tended to be represented by a smaller number of larger categories in the experiments with the SS-based iCVI-TopoARTMAP variants where $\rho_c$ was set to zero.

Figures~\ref{Fig:ARI1}, \ref{Fig:ARI2}, and~\ref{Fig:ARI3} show that, across all experiments, iCH-TopoARTMAP achieved and maintained high levels of performance (measured in terms of ARI) either earlier or at the same time as the other iCVI-TopoARTMAP variants. Specifically, in the experiments with class-incremental and mixed order presentations, iCH- and iXB-TopoARTMAP achieved these high ARI values much earlier than the other variants; nonetheless, these still showed an increasing performance trend over time. In the random presentation experiment, most variants quickly achieved and maintained high levels of performance; the exceptions were iCONN- and iXB-TopoARTMAP. The good performance of iCH-TopoARTMAP was accompanied with a tendency to accumulate a smaller number iCVI checks (Figs.~\ref{Fig:checks1}, \ref{Fig:checks2}, and~\ref{Fig:checks3}) as well as to generate and maintain a smaller number of categories (Figs.~\ref{Fig:cat1}, \ref{Fig:cat2}, and~\ref{Fig:cat3}) and clusters (Figs.~\ref{Fig:cl1}, \ref{Fig:cl2}, and~\ref{Fig:cl3}) over time when compared to the other iCVI-TopoARTMAP variants --- this can be also observed in the final experimental results reported in Table~\ref{Tab:results_synthetic}. On the other hand, as discussed in Section~\ref{sec:results}, these figures clearly show the category proliferation issue and over-partition tendency of the iDB-TopoARTMAP variant in class-incremental and mixed presentations that culminated in the experimental results reported in Table~\ref{Tab:results_synthetic}.

As discussed in Section~\ref{sec:ordering}, most of the iCVI-TopoARTMAP variants were relatively indifferent to the order of presentation of samples (Figs.~\ref{Fig:iCH1}-\ref{Fig:iCONN1}, \ref{Fig:iCH2}-\ref{Fig:iCONN2}, and~\ref{Fig:iCH3}-\ref{Fig:iCONN3}). On the other hand, Figs.~\ref{Fig:skm1}-\ref{Fig:iskm1}, \ref{Fig:skm2}-\ref{Fig:iskm2}, and~\ref{Fig:skm3}-\ref{Fig:iskm3} indicate that skm and iskm are very sensitive to ordering effects.

\newcommand{\spt}{.28}
\newcommand{\spovars}{.48}

\begin{figure}[!hp]
\centering
\subcaptionbox{\label{Fig:iCH1}iCH-TopoFAM}{\includegraphics[width=\spt\textwidth]{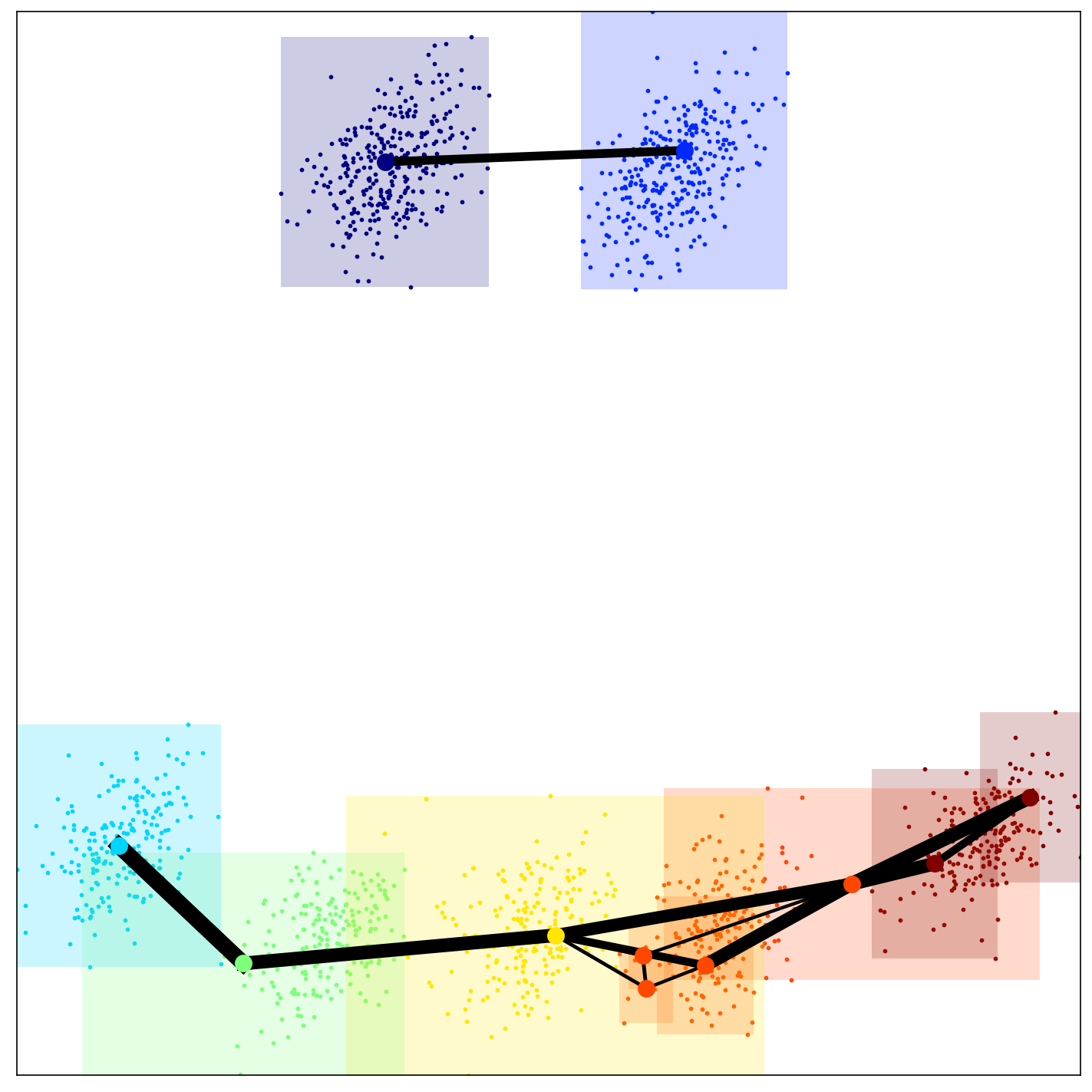}}
\hfill
\subcaptionbox{\label{Fig:iWB1}iWB-TopoFAM}{\includegraphics[width=\spt\textwidth]{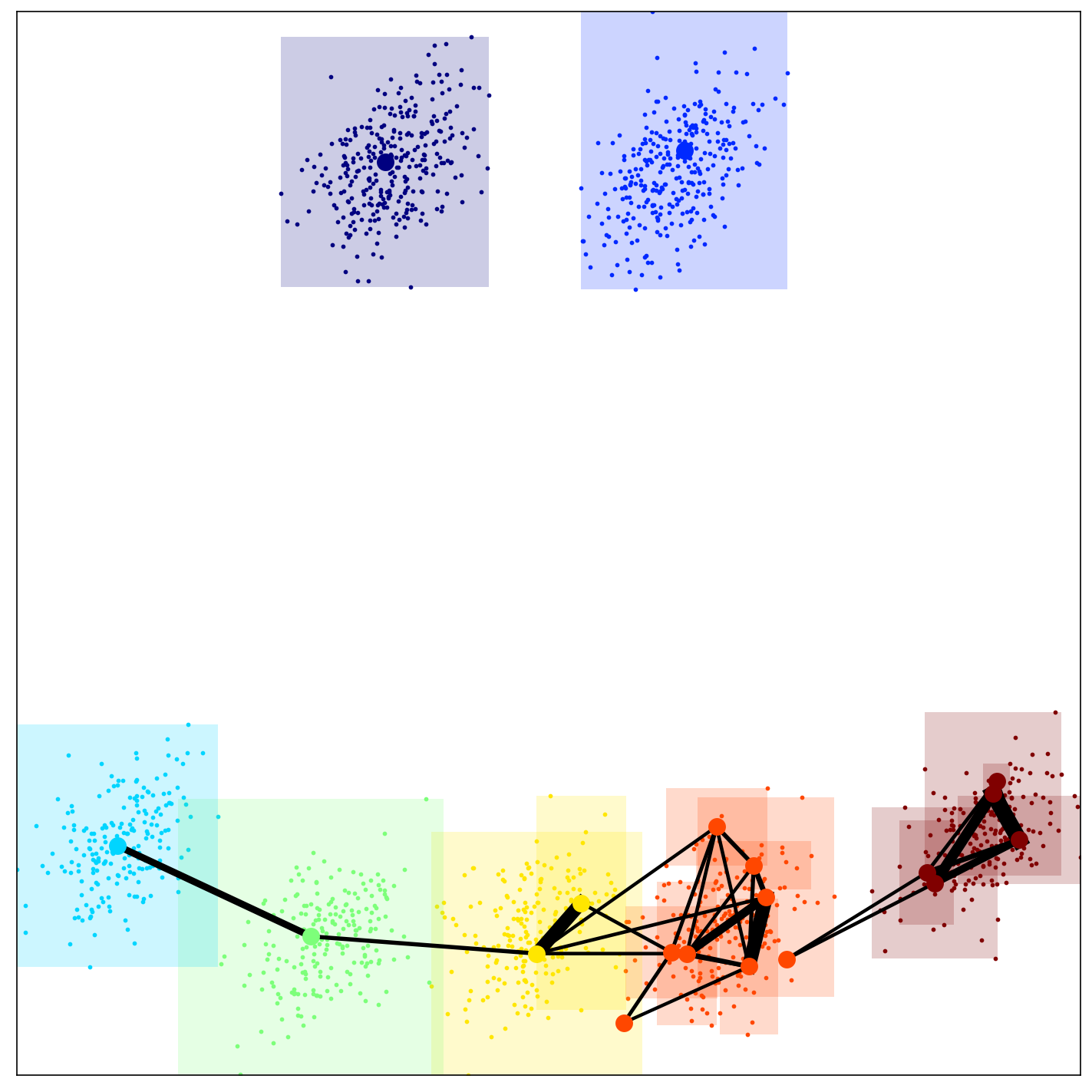}}
\hfill
\subcaptionbox{\label{Fig:iPBM1}iPBM-TopoFAM}{\includegraphics[width=\spt\textwidth]{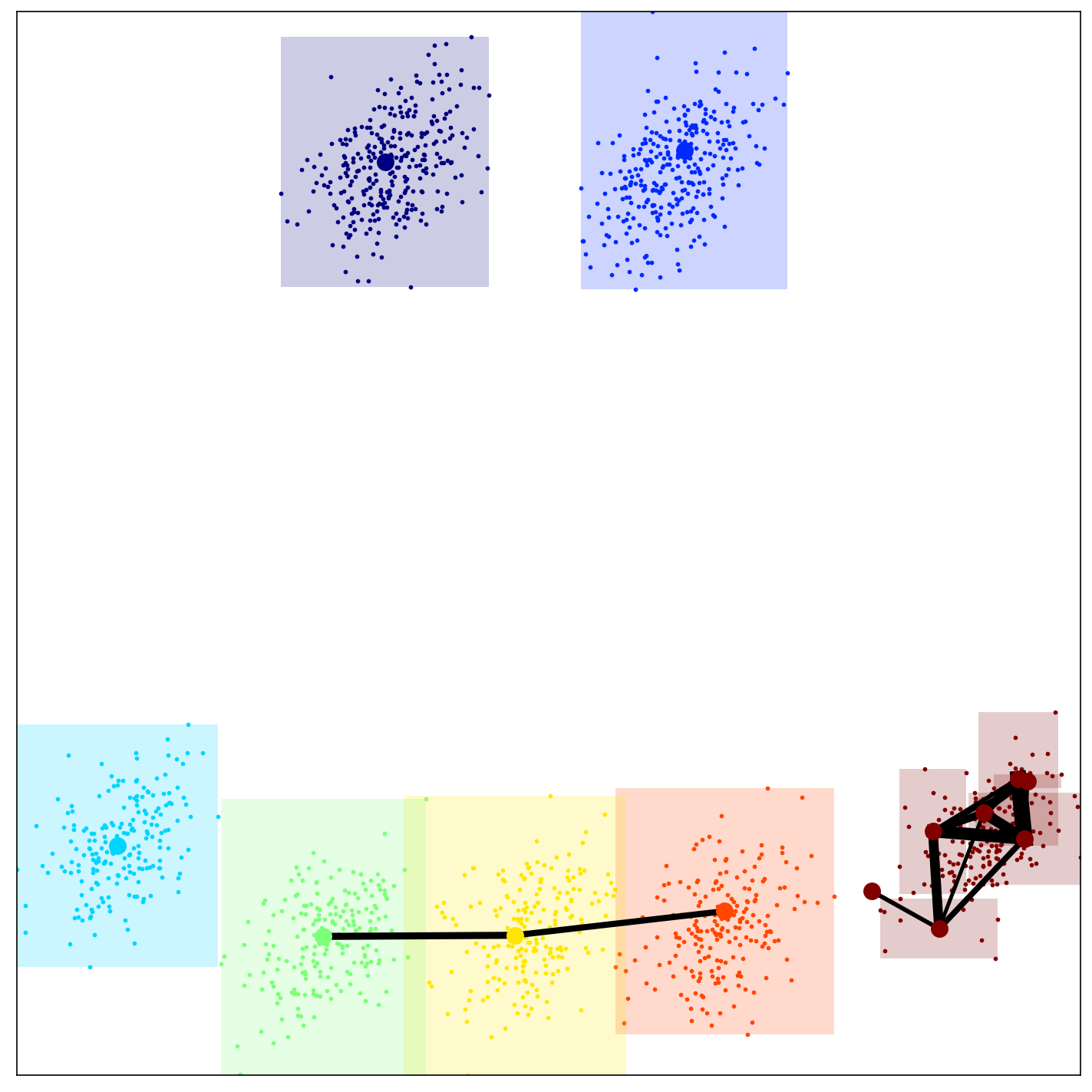}}

\subcaptionbox{\label{Fig:iXB1}iXB-TopoFAM}{\includegraphics[width=\spt\textwidth]{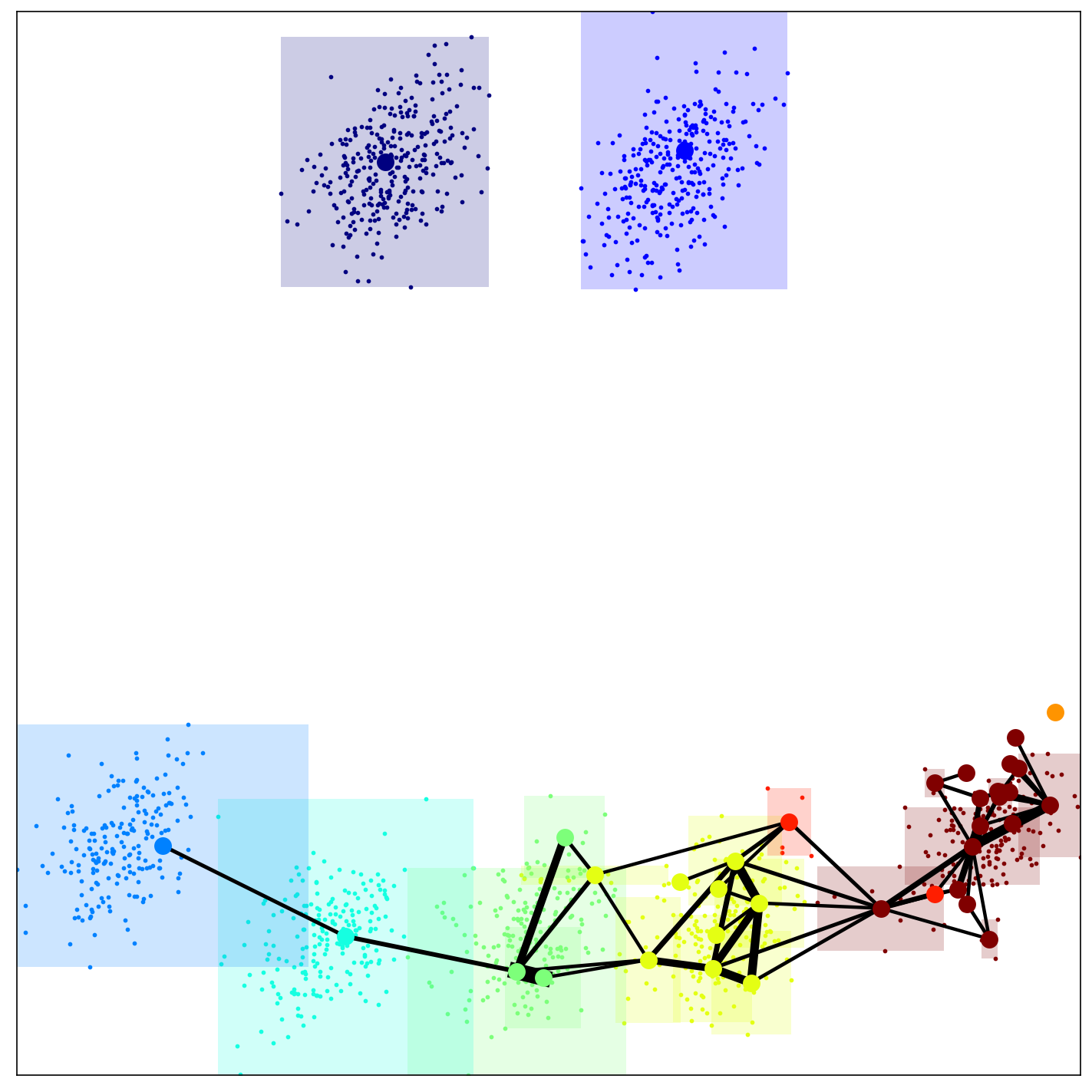}}
\hfill
\subcaptionbox{\label{Fig:iDB1}iDB-TopoFAM}{\includegraphics[width=\spt\textwidth]{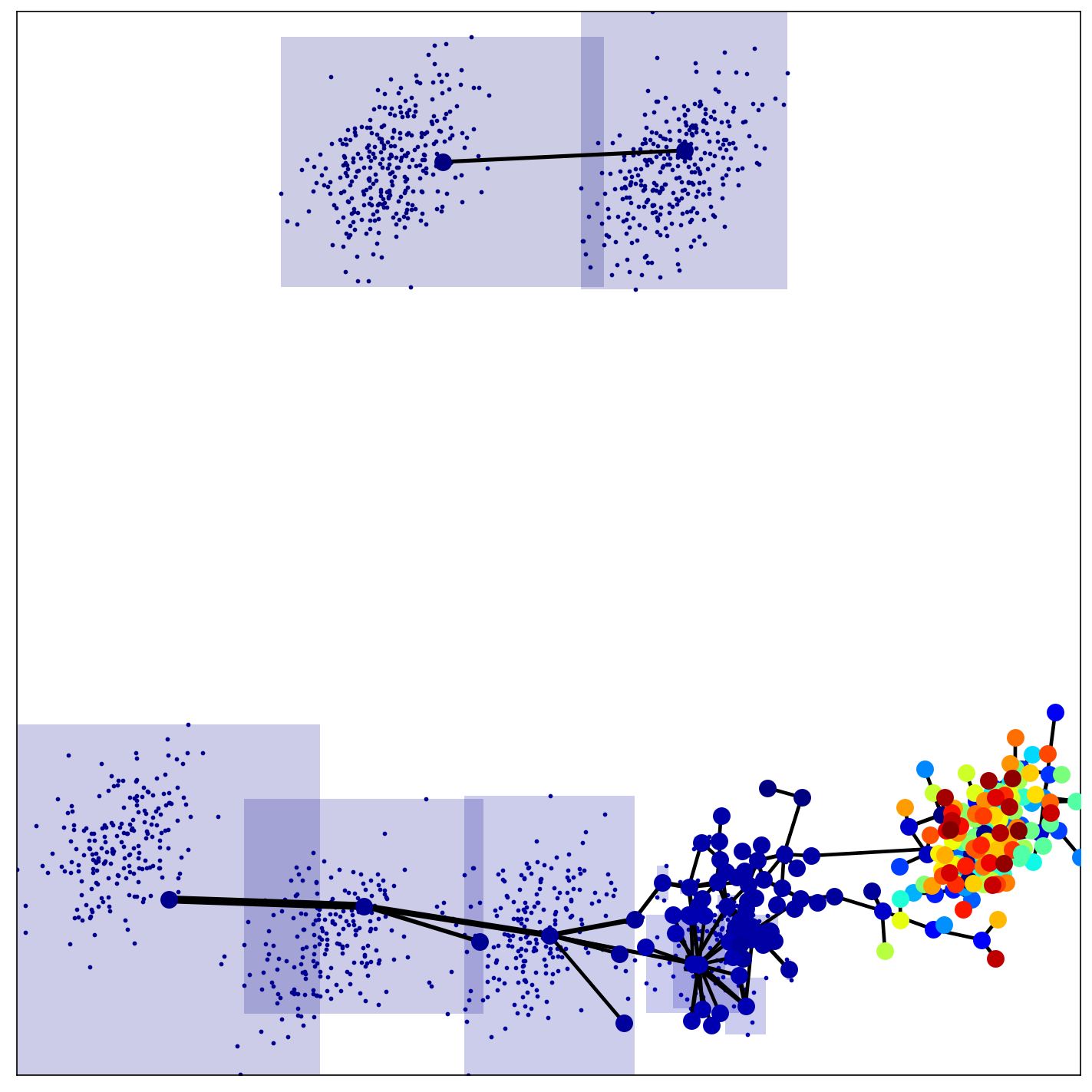}}
\hfill
\subcaptionbox{\label{Fig:iCONN1}iCONN-TopoFAM}{\includegraphics[width=\spt\textwidth]{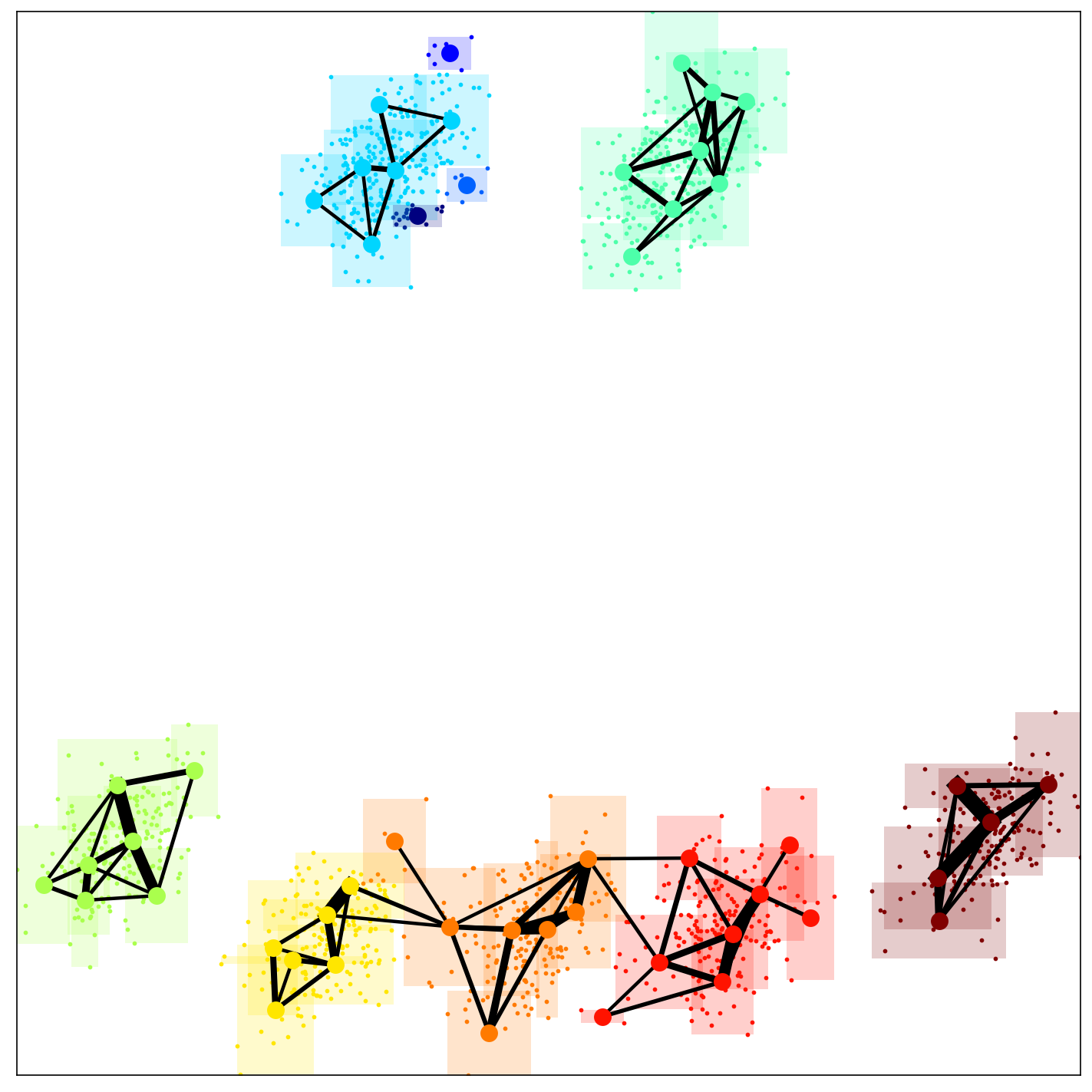}}

\subcaptionbox{\label{Fig:DVFA1}WS-DVFA}{\includegraphics[width=\spt\textwidth]{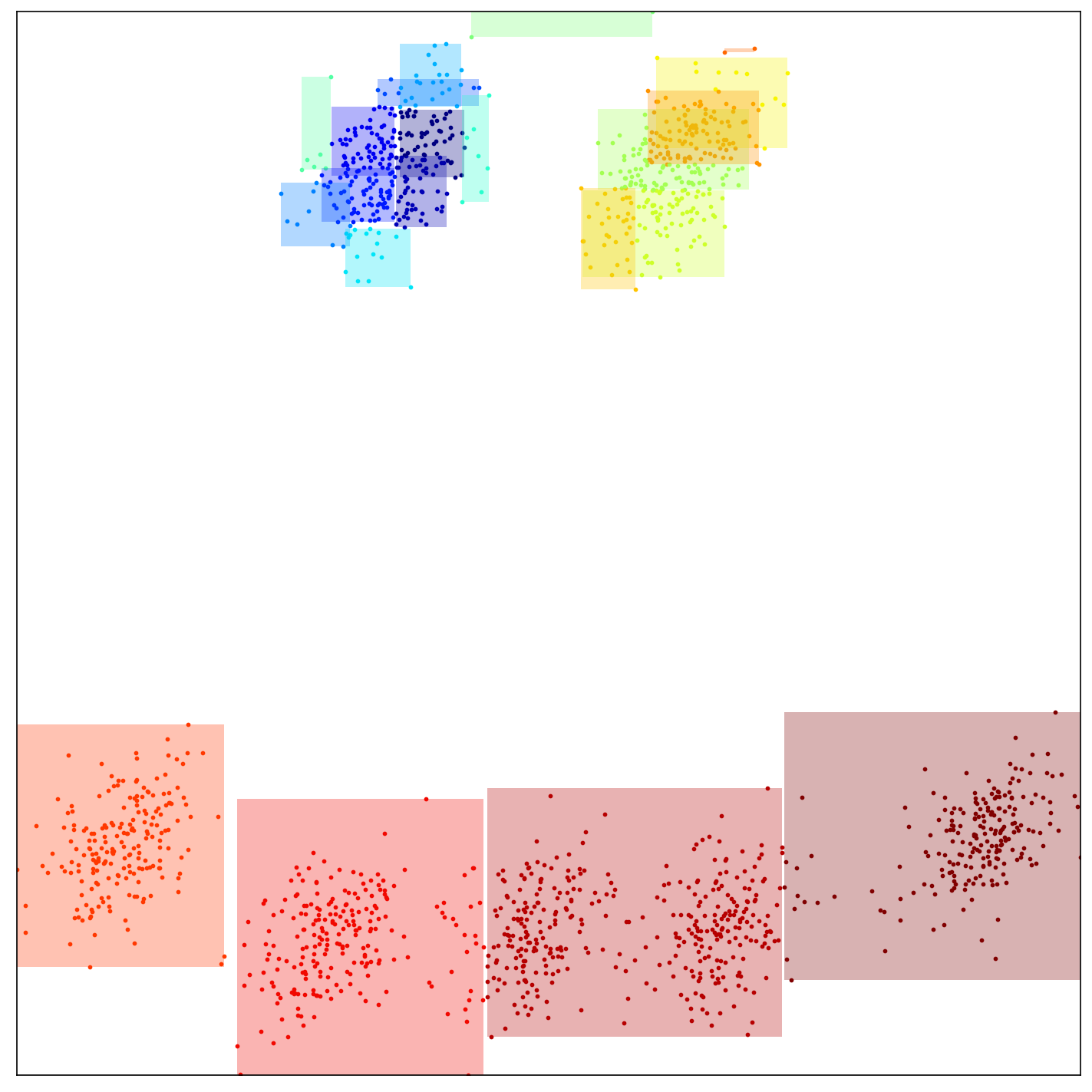}}
\hfill
\subcaptionbox{\label{Fig:TopoFA1}WS-TopoFA}{\includegraphics[width=\spt\textwidth]{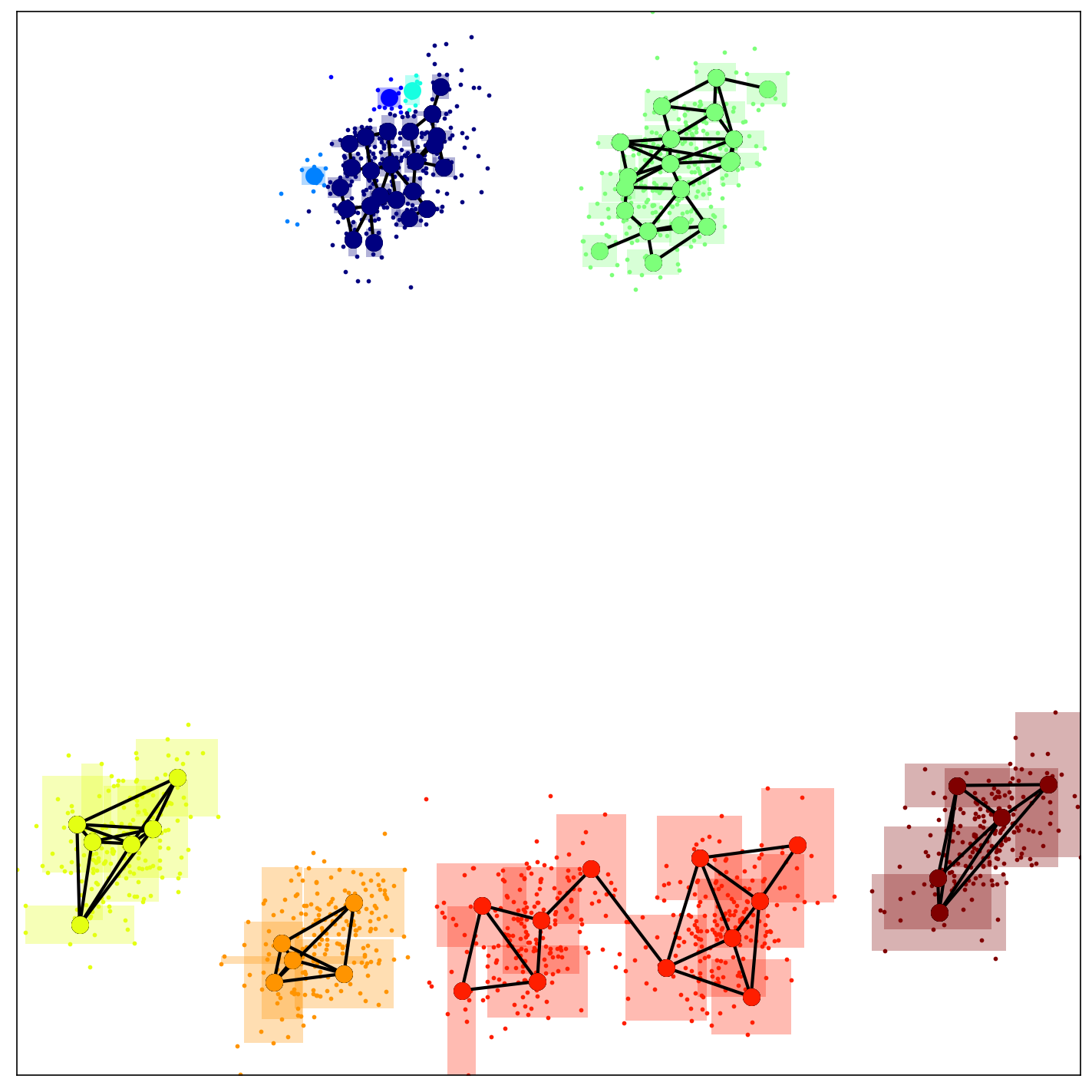}}
\hfill
\subcaptionbox{\label{Fig:DRN1}DRN}{\includegraphics[width=\spt\textwidth]{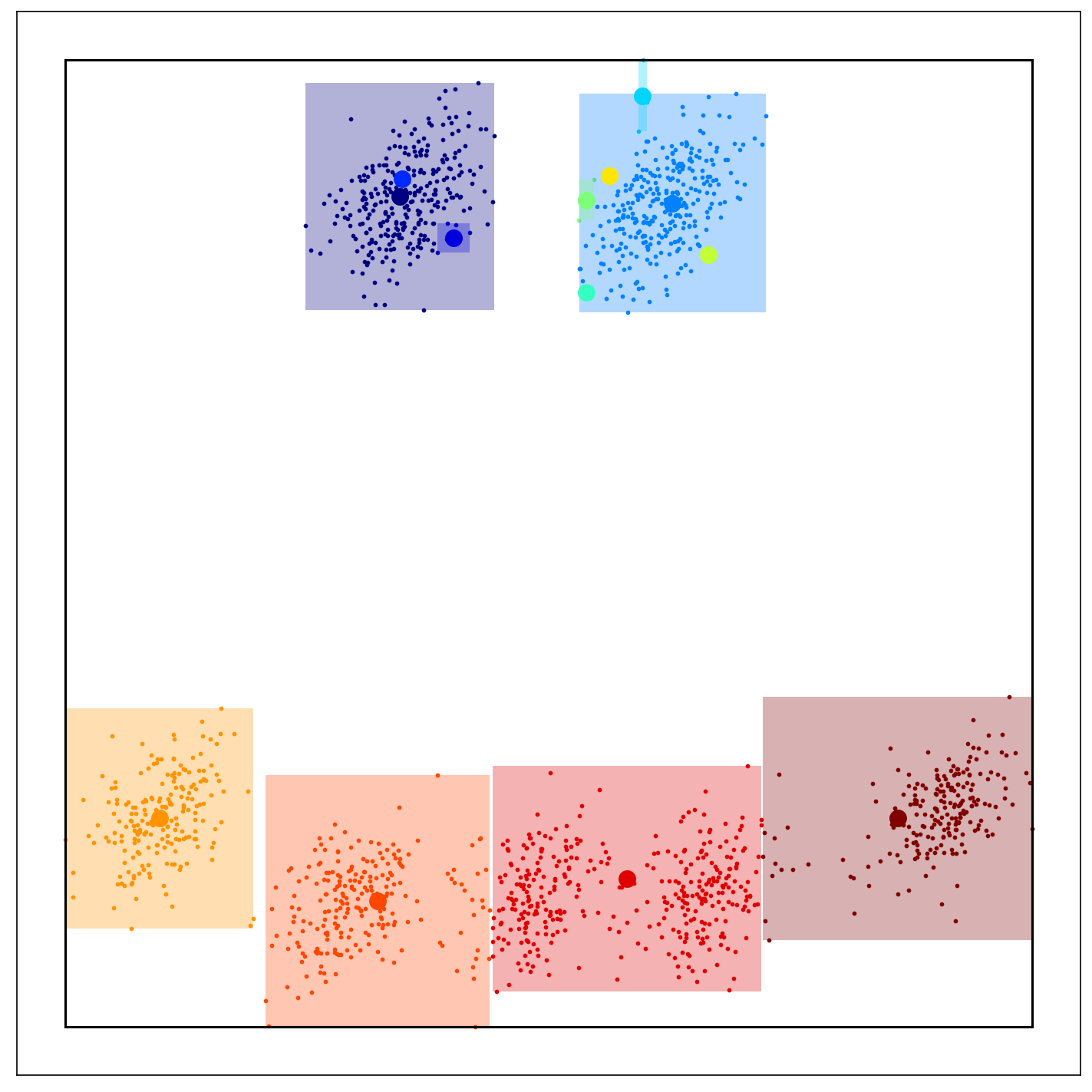}}

\subcaptionbox{\label{Fig:skm1}skm}{\includegraphics[width=\spt\textwidth]{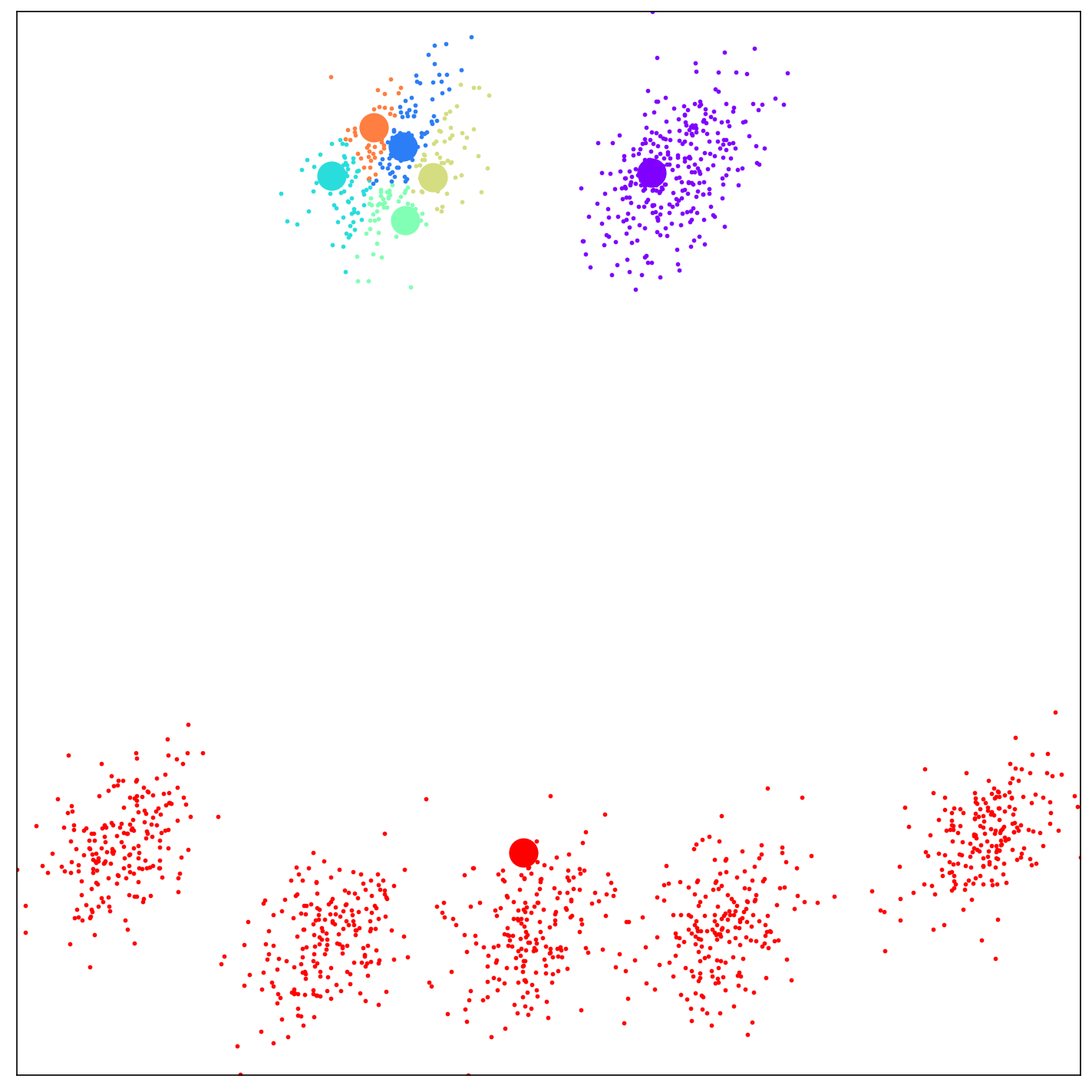}}
\hfill
\subcaptionbox{\label{Fig:iskm1}iskm}{\includegraphics[width=\spt\textwidth]{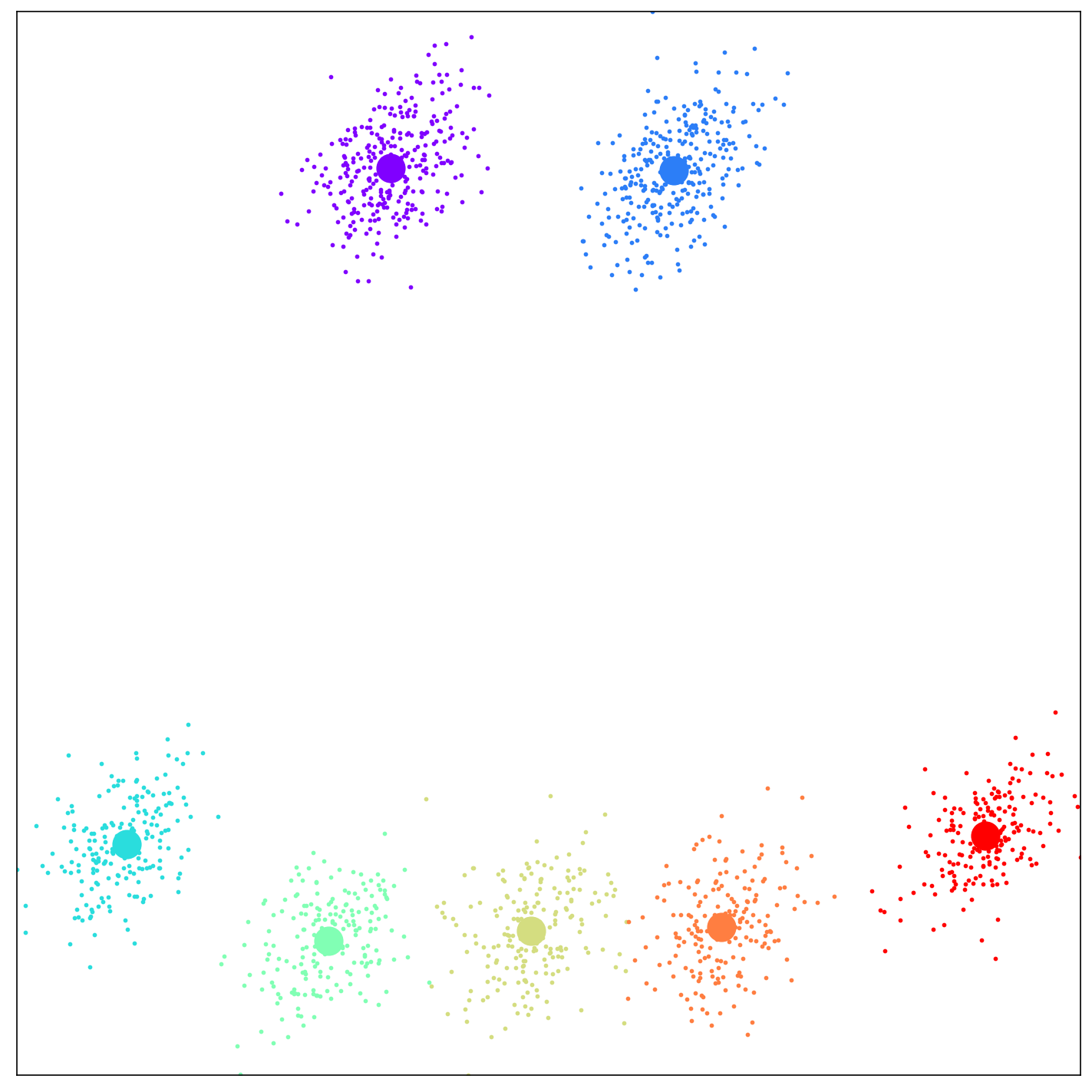}}
\caption{Color-coded footprints of the clustering algorithms for the class-incremental presentation experiment and sample assignments as per Table~\ref{Tab:results_synthetic}. The connections shown in the iCVI-TopoARTMAP variants represent the $CONN$ matrix (thicker lines represent stronger connections) --- see CONNvis~\cite{tasdemir2009}. Although categories might be connected, the clusters in iCVI-TopoARTMAP are determined by the map field. The black outer box shown in (i) represents DRN's global weight vector.}
\label{Fig:partitions_synthetic_1}
\end{figure}

\begin{figure}[!hp]
\centering
\includegraphics[width=\textwidth]{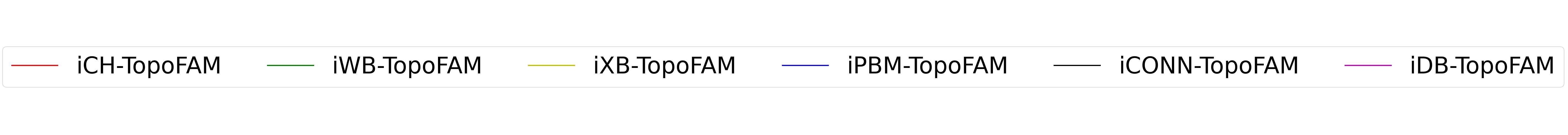}

\centering
\subcaptionbox{\label{Fig:ARI1}ARI}{\includegraphics[width=\spovars\textwidth]{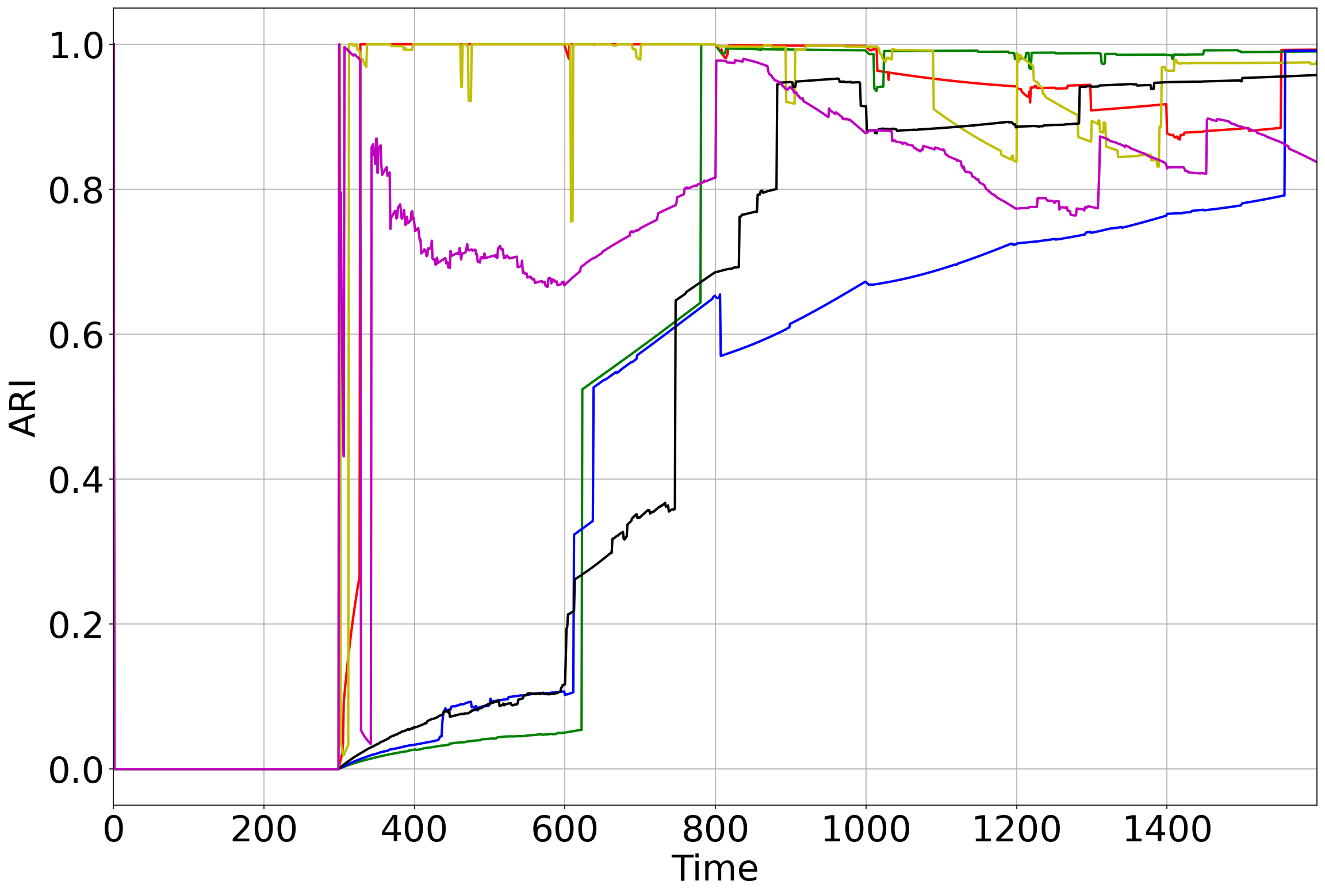}}
\hfill
\subcaptionbox{\label{Fig:iCVI1}iCVI}{\includegraphics[width=\spovars\textwidth]{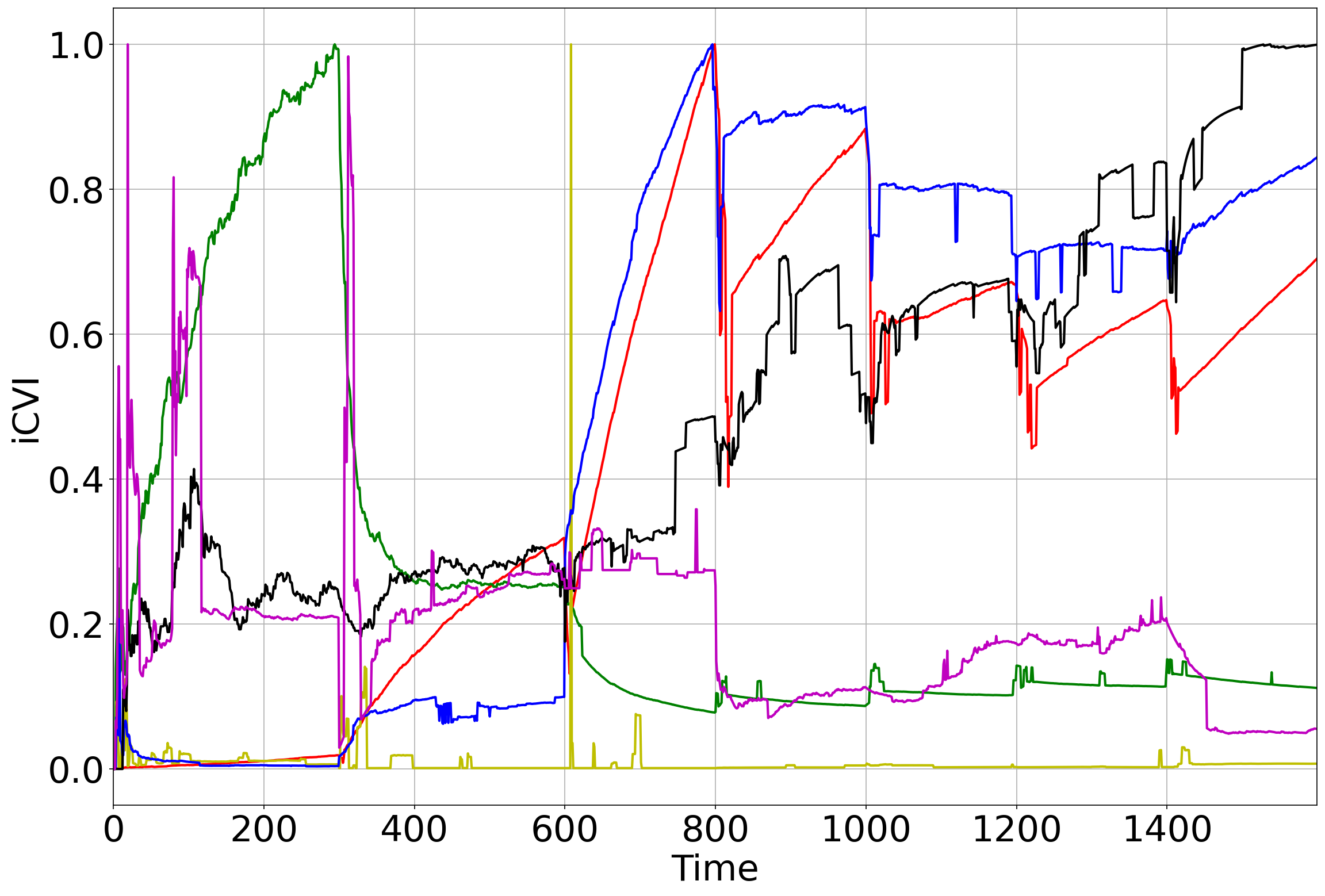}}

\subcaptionbox{\label{Fig:checks1}iCVI checks}{\includegraphics[width=\spovars\textwidth]{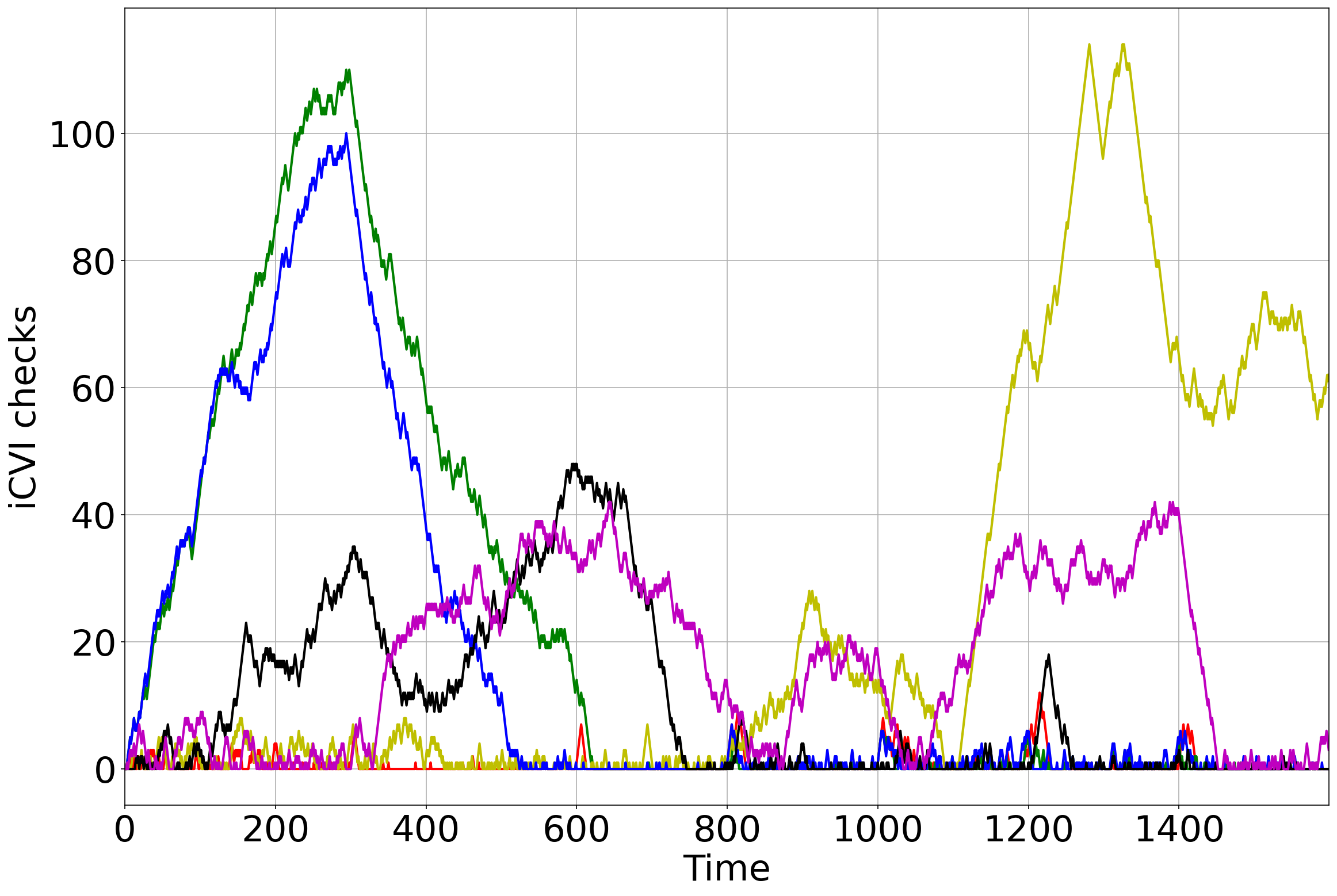}}
\hfill
\subcaptionbox{\label{Fig:vig1}Vigilance}{\includegraphics[width=\spovars\textwidth]{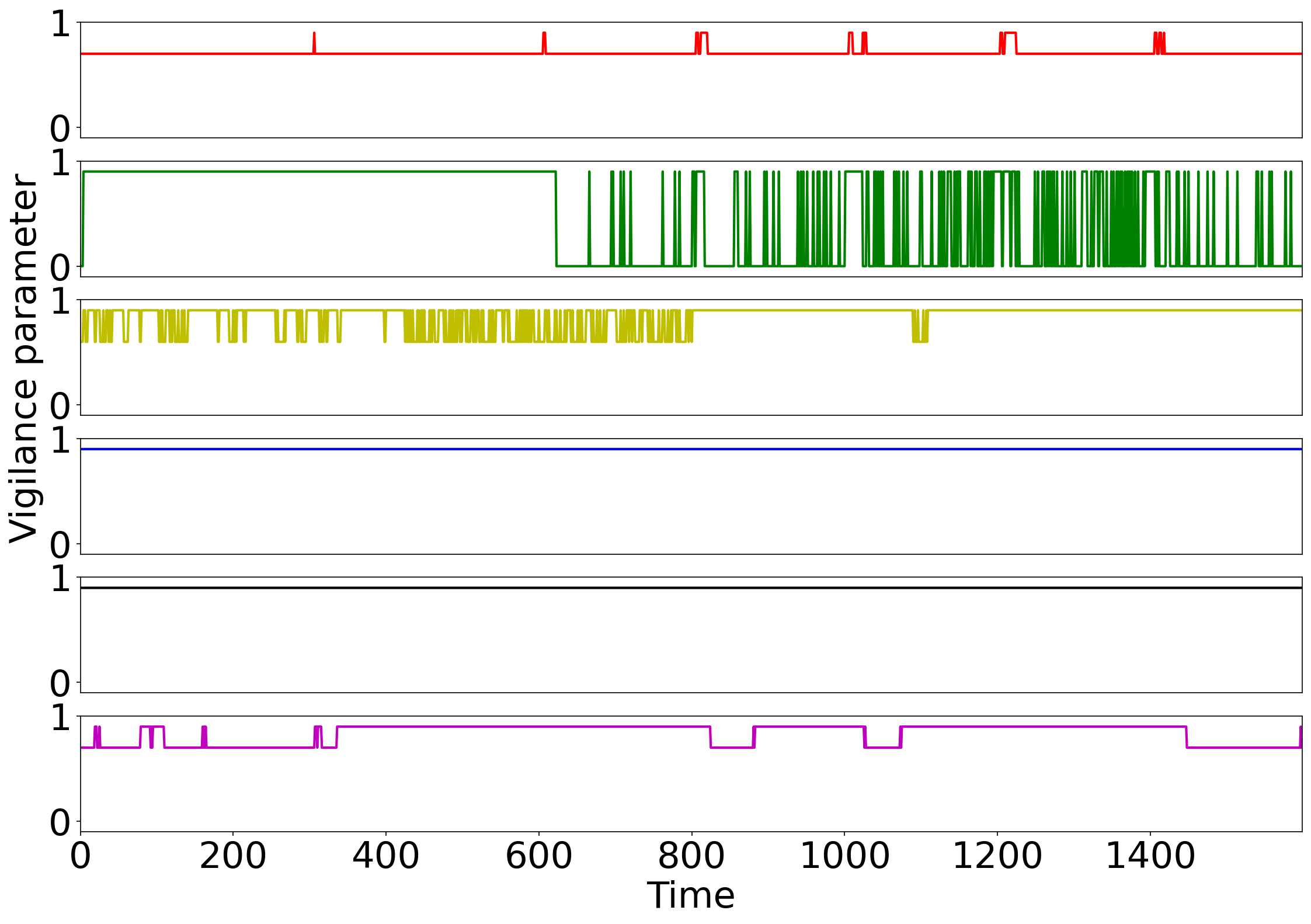}}

\subcaptionbox{\label{Fig:cat1}Categories}{\includegraphics[width=\spovars\textwidth]{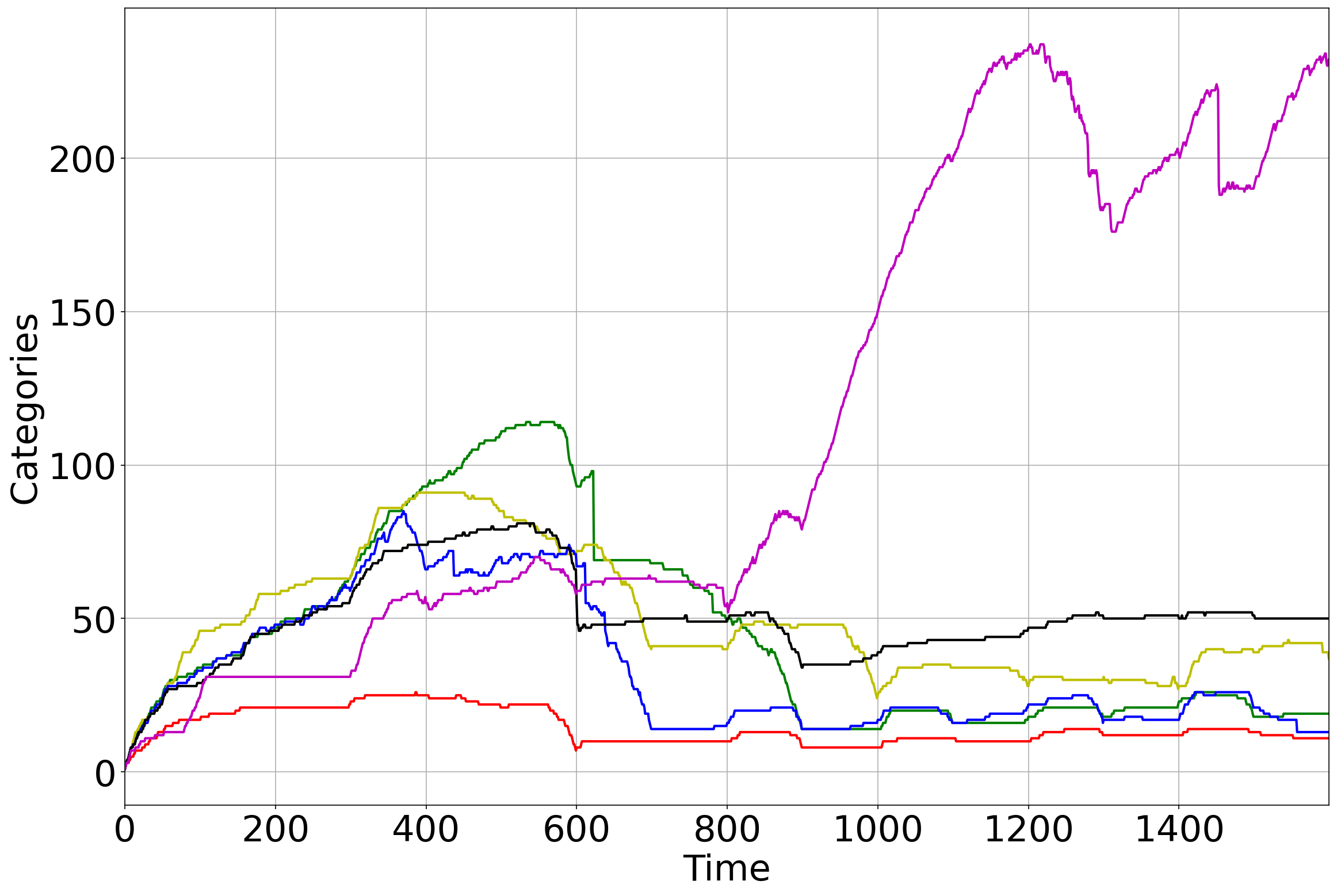}}
\hfill
\subcaptionbox{\label{Fig:cl1}Clusters}{\includegraphics[width=\spovars\textwidth]{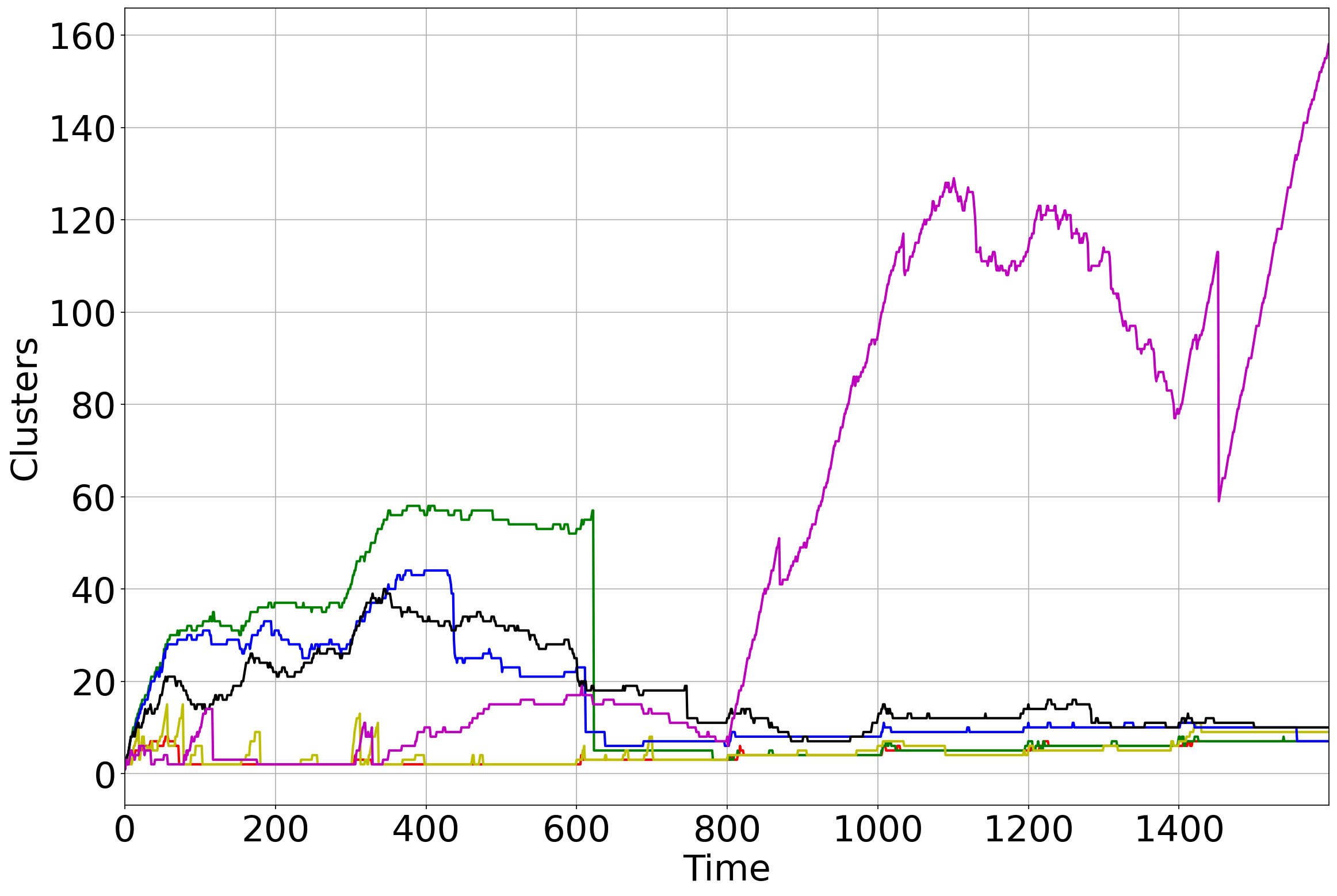}}

\caption{Tracking of iCVI-TopoARTMAP variants during the class-incremental order experiment (as per Table~\ref{Tab:results_synthetic}): ARI, iCVI values, iCVI checks ($v$), vigilance parameter of module A ($\rho_a$), number of categories, and number of clusters over time. The iCVI values were normalized to a common range ($[0, 1]$) and peaks were clipped for visualization purposes.}
\label{Fig:tracking_iCVI_TopoFAM_1}
\end{figure}

\begin{figure}[!hp]
\centering
\subcaptionbox{\label{Fig:iCH2}iCH-TopoFAM}{\includegraphics[width=\spt\textwidth]{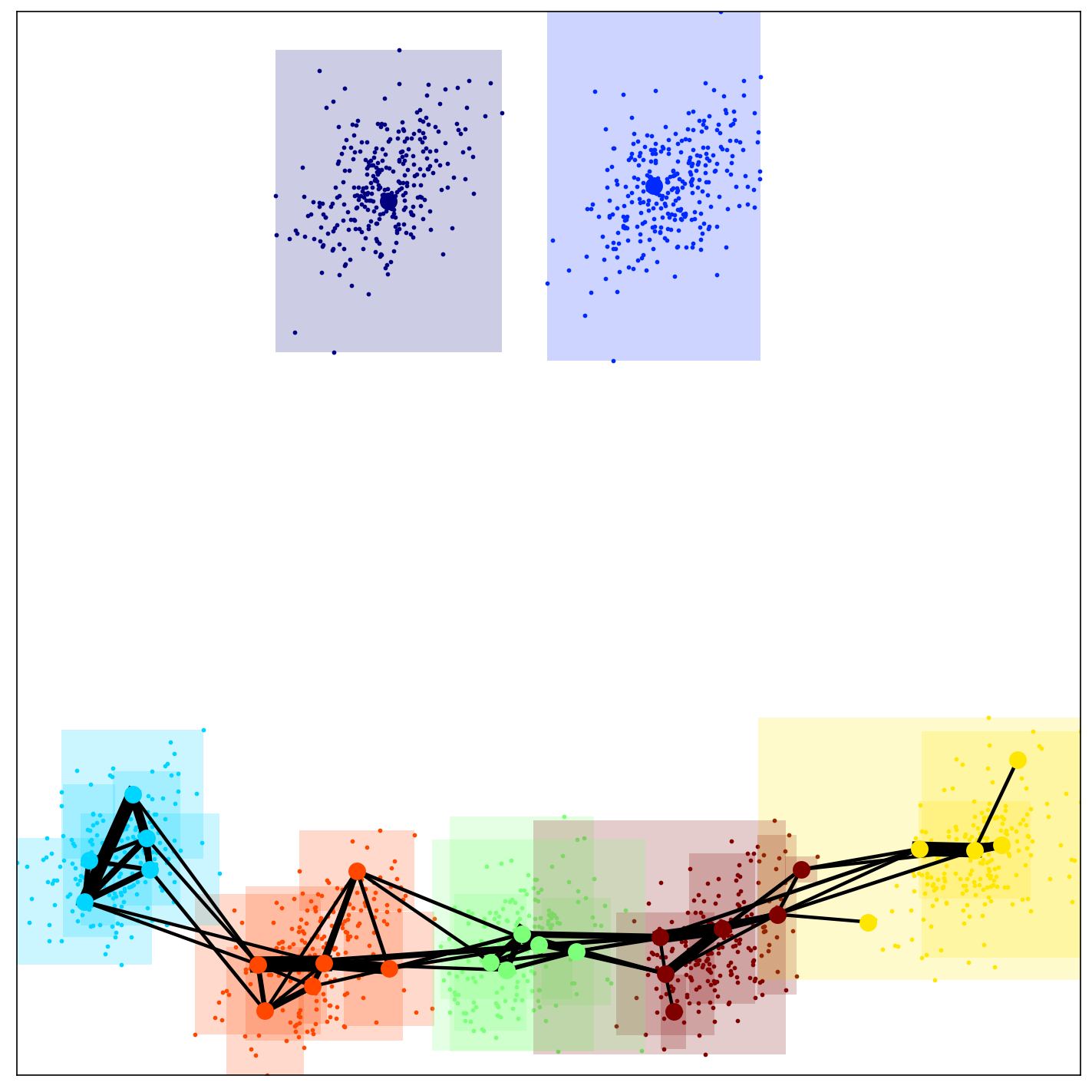}}
\hfill
\subcaptionbox{\label{Fig:iWB2}iWB-TopoFAM}{\includegraphics[width=\spt\textwidth]{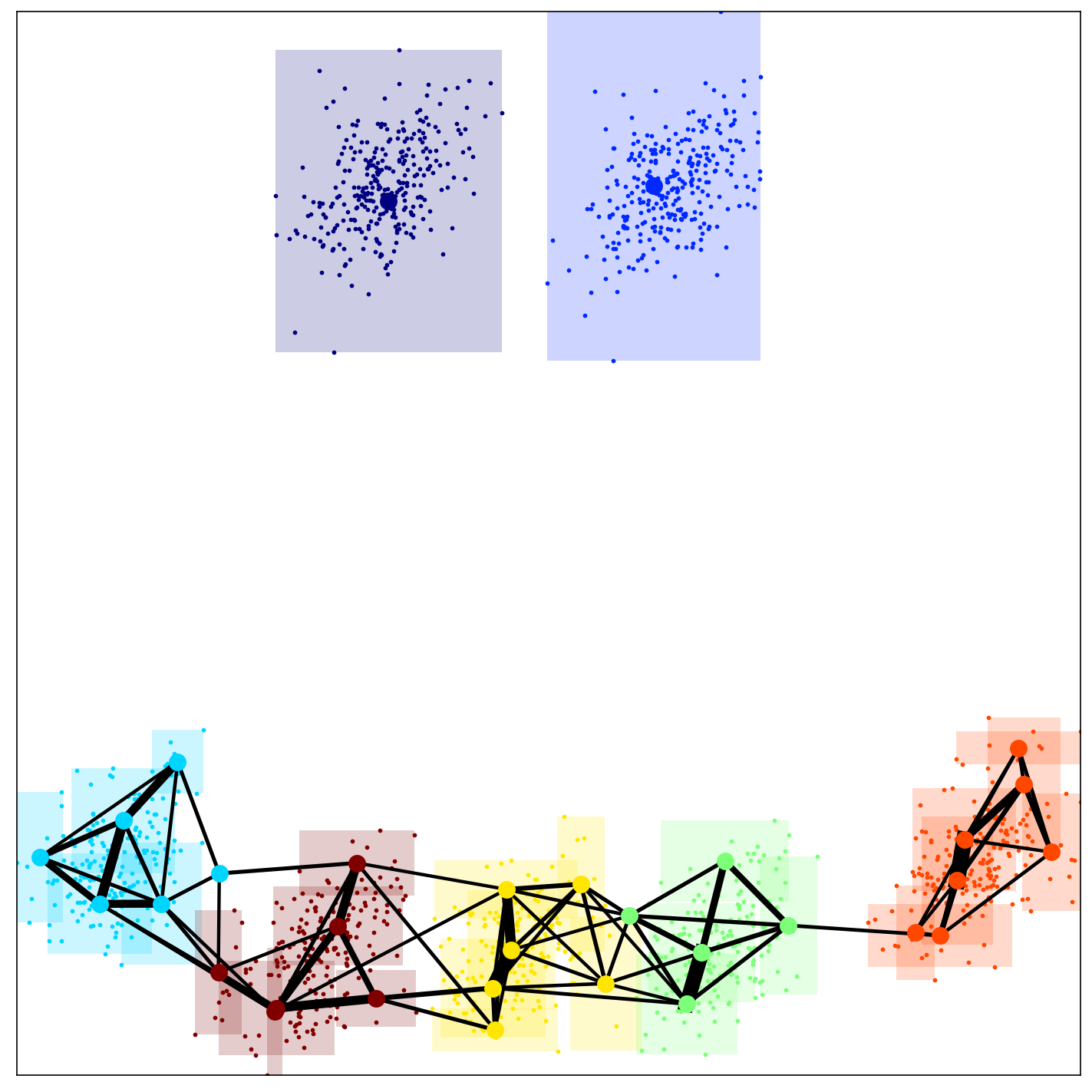}}
\hfill
\subcaptionbox{\label{Fig:iPBM2}iPBM-TopoFAM}{\includegraphics[width=\spt\textwidth]{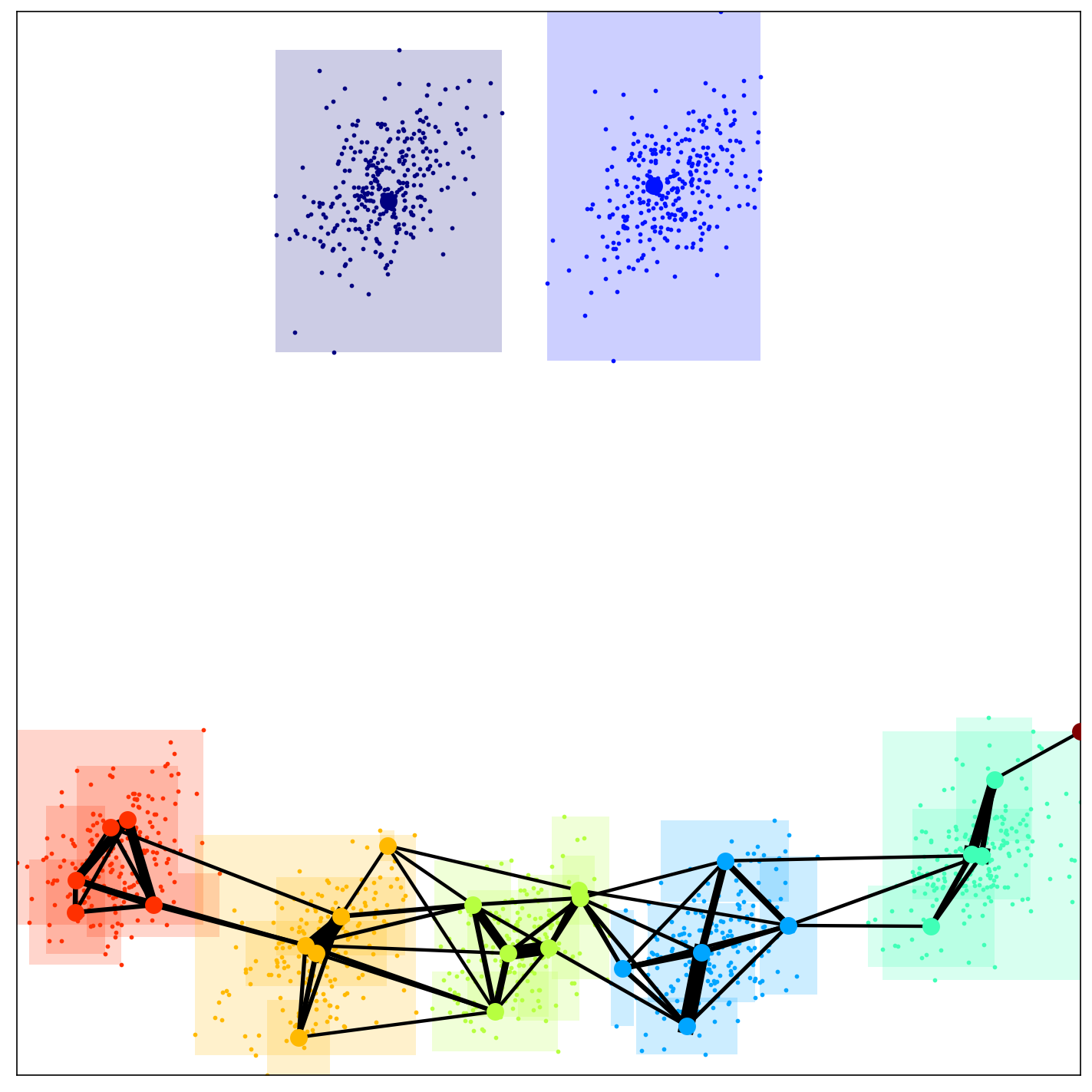}}

\subcaptionbox{\label{Fig:iXB2}iXB-TopoFAM}{\includegraphics[width=\spt\textwidth]{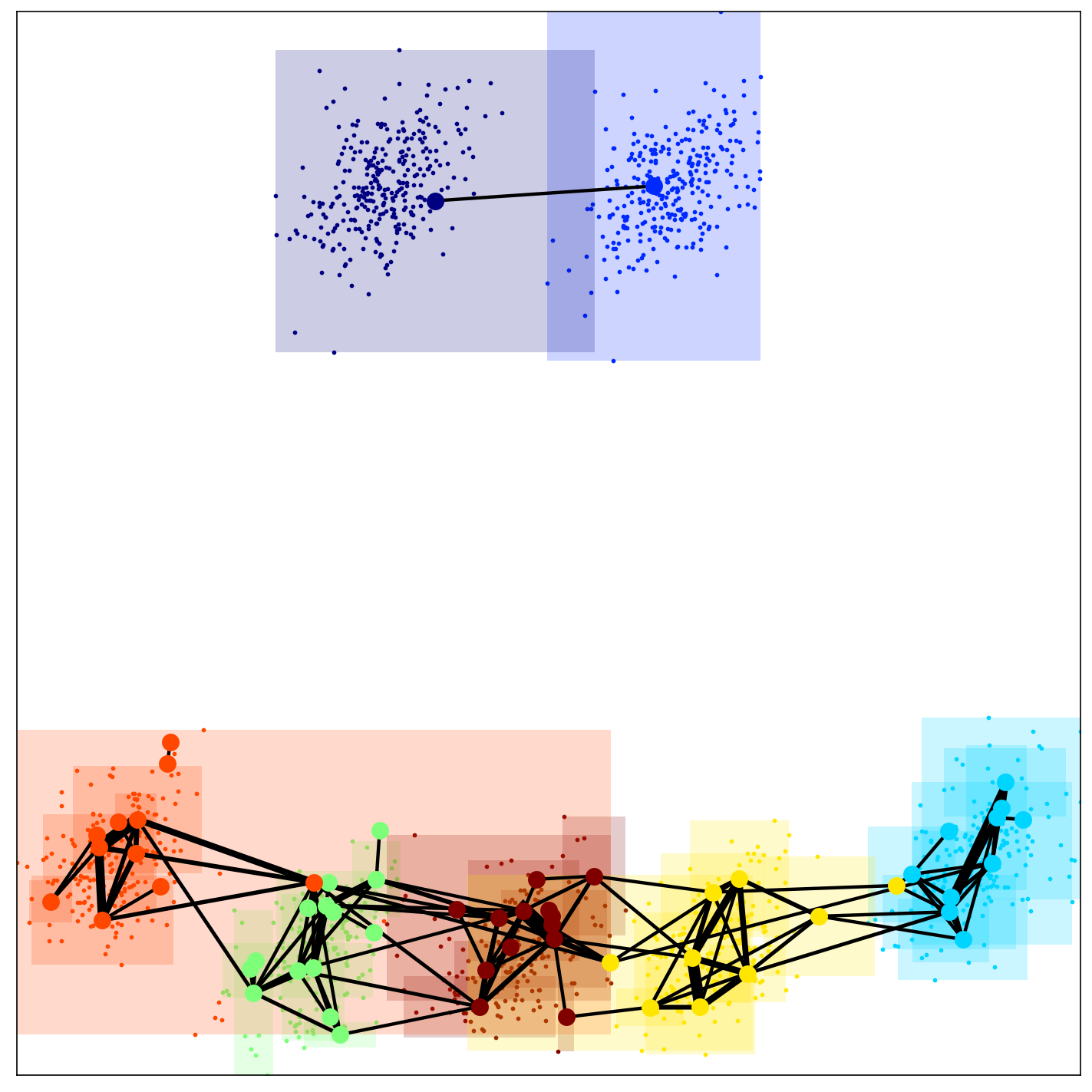}}
\hfill
\subcaptionbox{\label{Fig:iDB2}iDB-TopoFAM}{\includegraphics[width=\spt\textwidth]{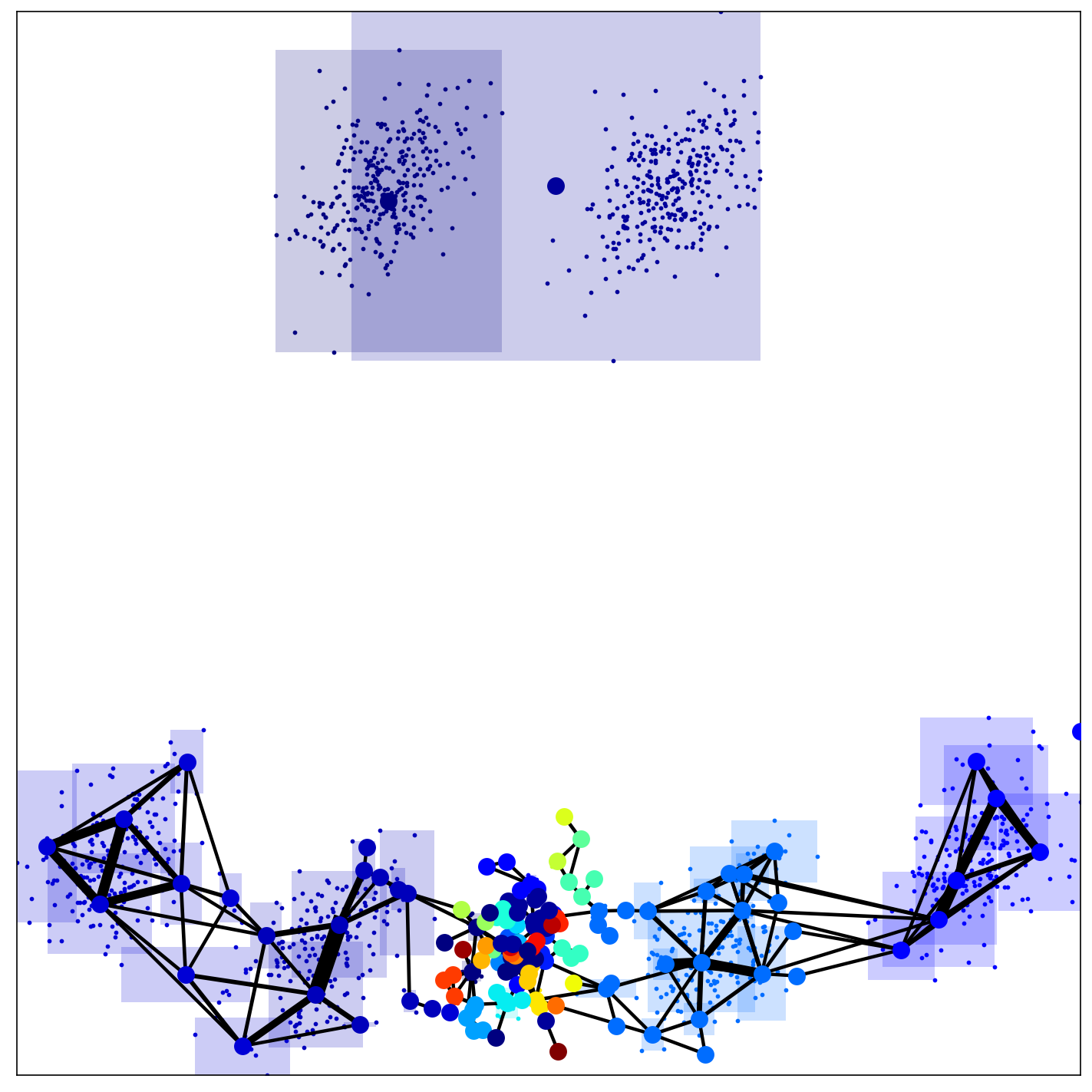}}
\hfill
\subcaptionbox{\label{Fig:iCONN2}iCONN-TopoFAM}{\includegraphics[width=\spt\textwidth]{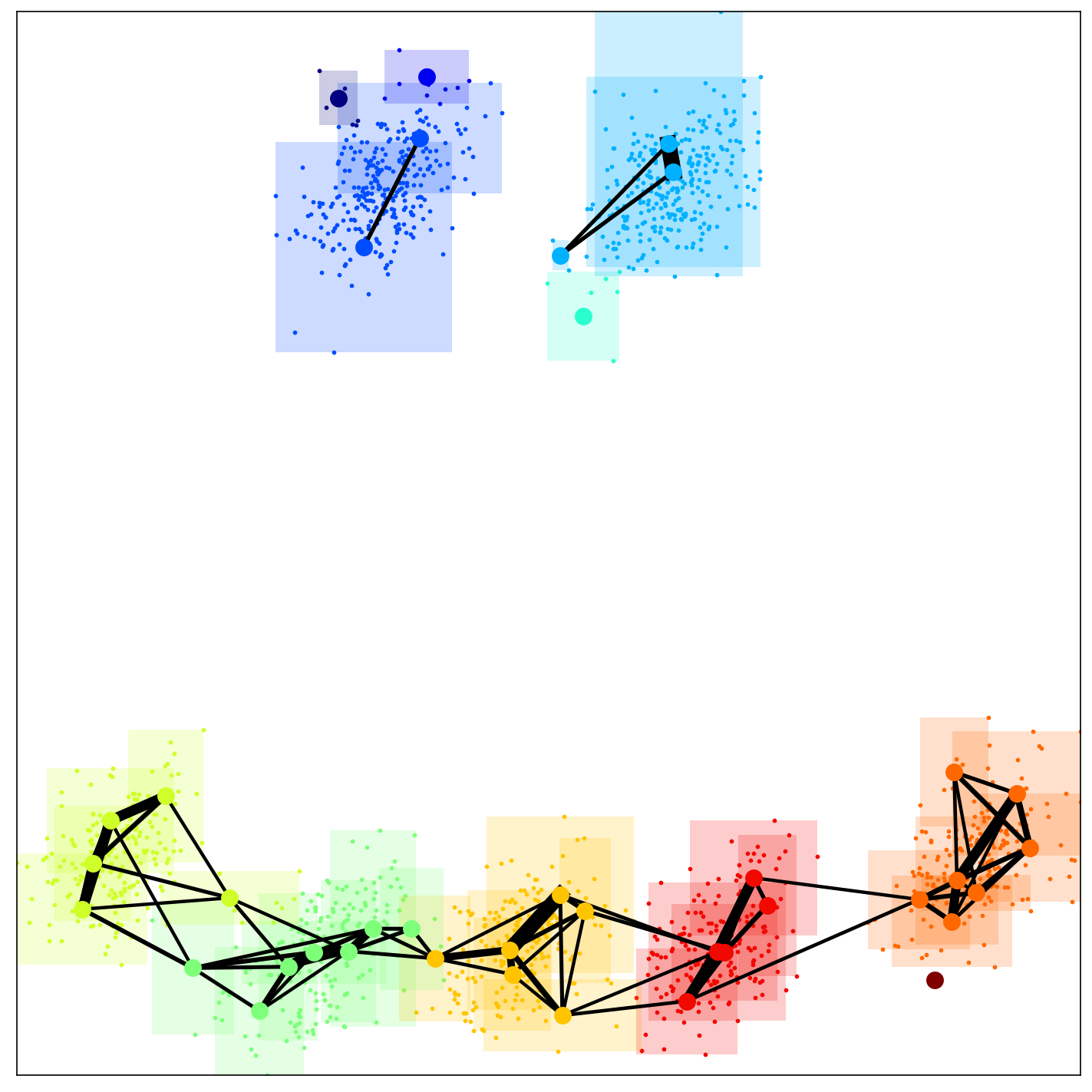}}

\subcaptionbox{\label{Fig:DVFA2}WS-DVFA}{\includegraphics[width=\spt\textwidth]{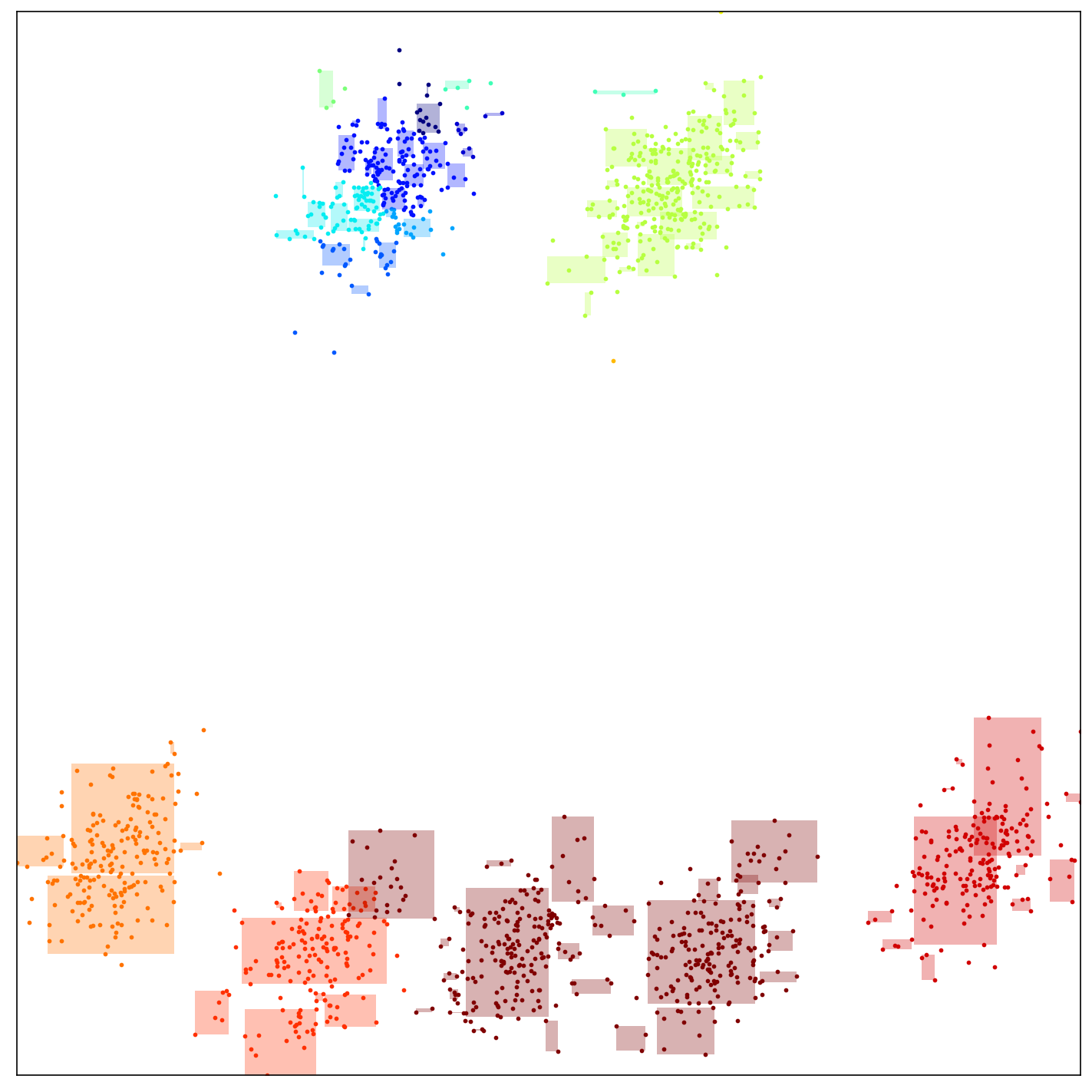}}
\hfill
\subcaptionbox{\label{Fig:TopoFA2}WS-TopoFA}{\includegraphics[width=\spt\textwidth]{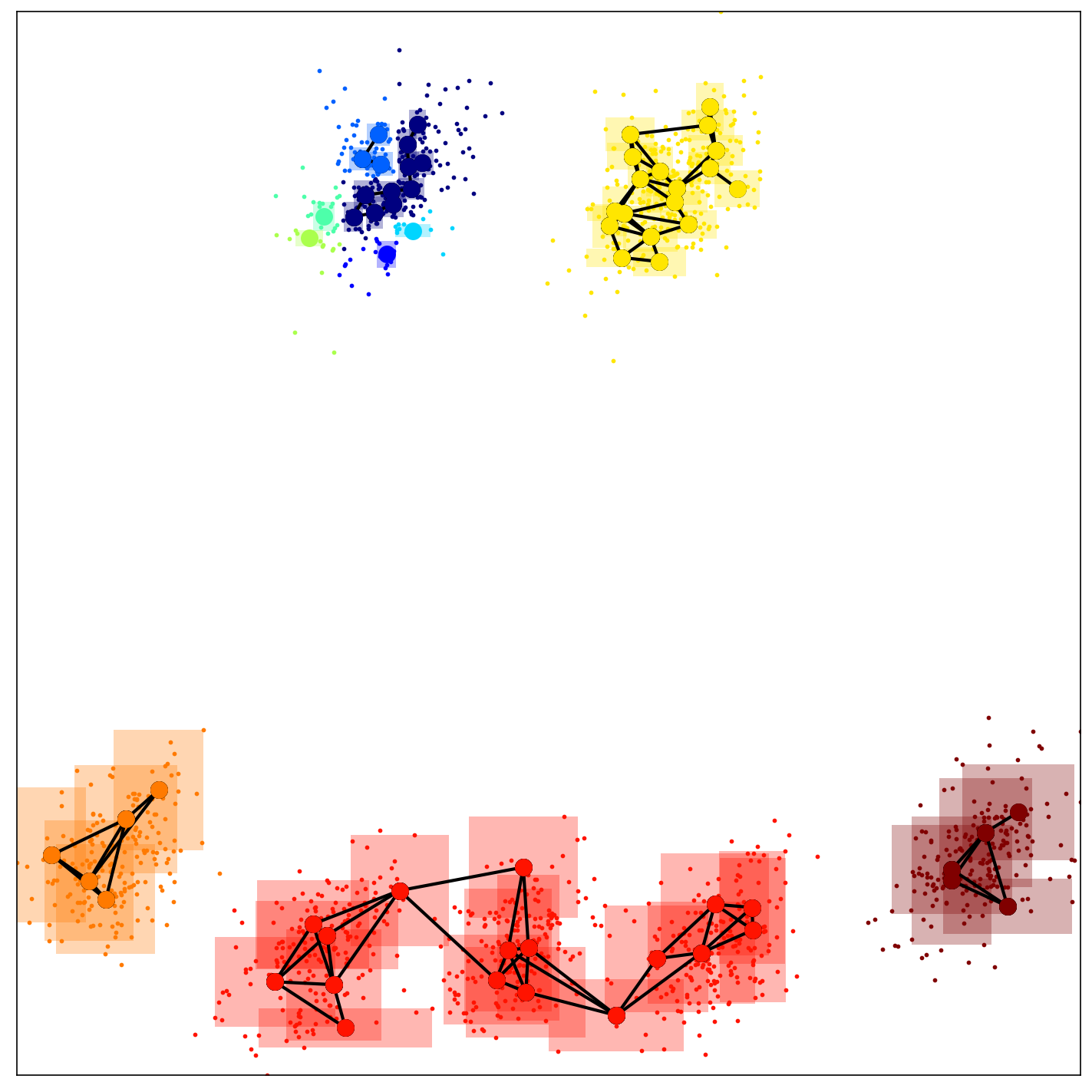}}
\hfill
\subcaptionbox{\label{Fig:DRN2}DRN}{\includegraphics[width=\spt\textwidth]{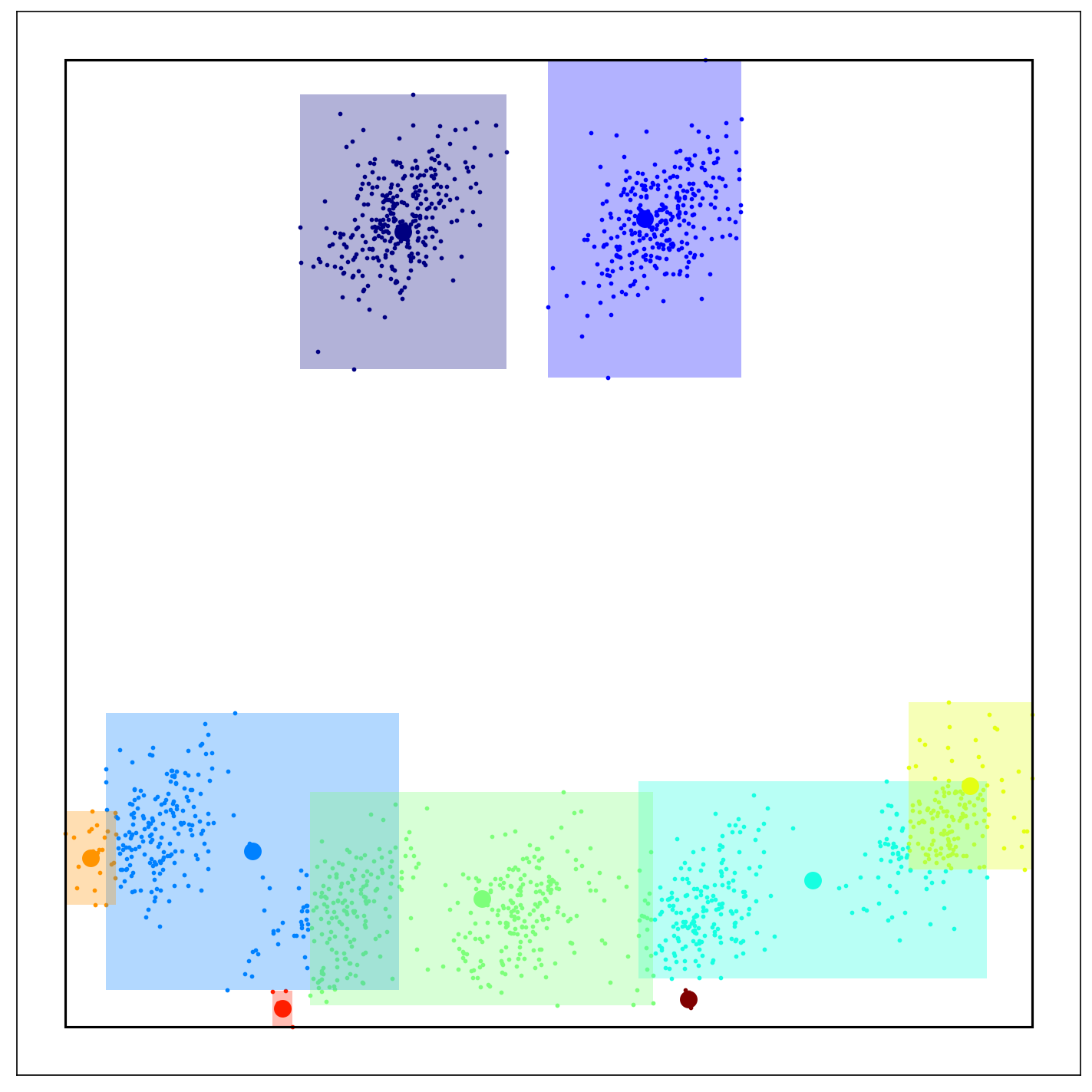}}

\subcaptionbox{\label{Fig:skm2}skm}{\includegraphics[width=\spt\textwidth]{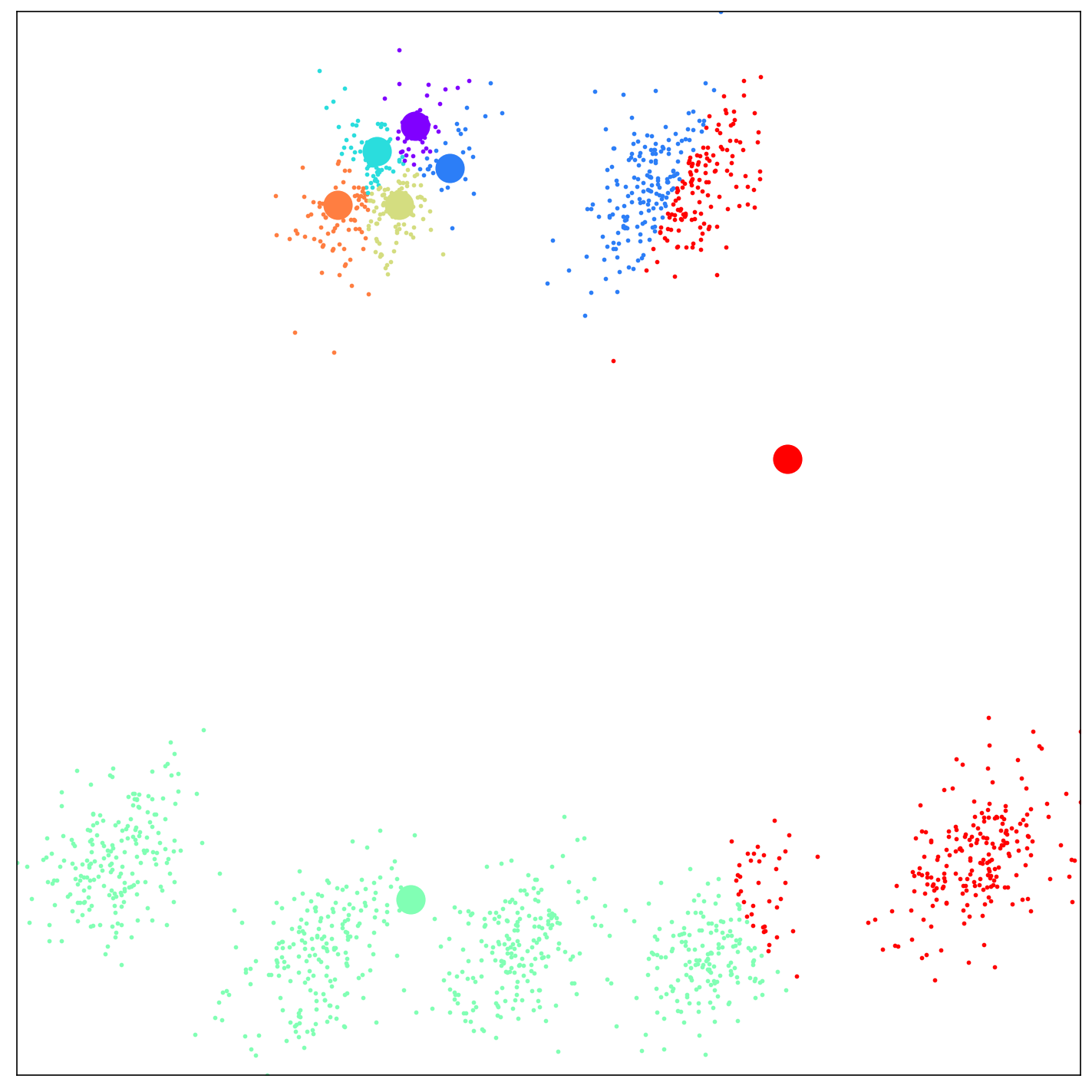}}
\hfill
\subcaptionbox{\label{Fig:iskm2}iskm}{\includegraphics[width=\spt\textwidth]{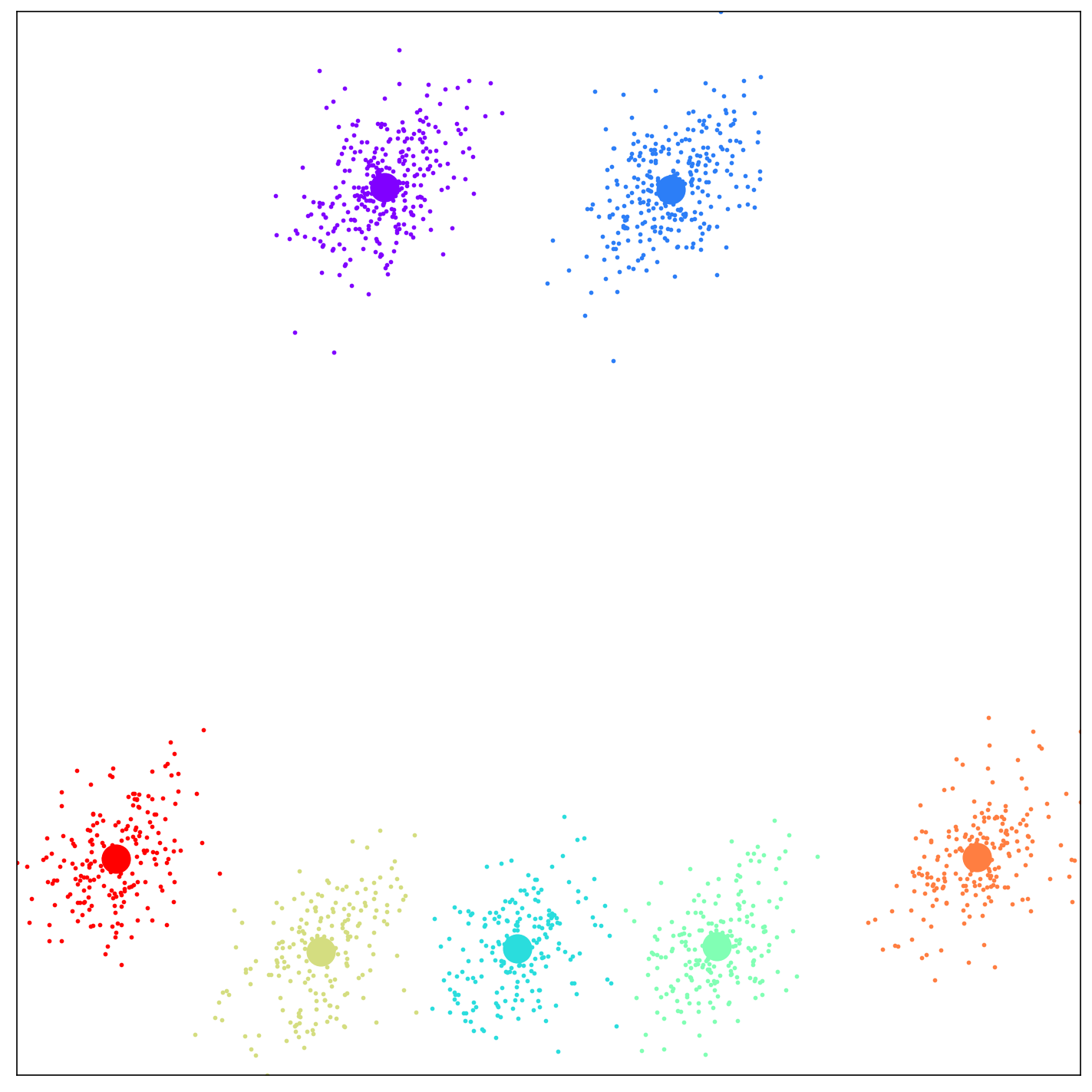}}
\caption{Color-coded footprints of the clustering algorithms for the mixed presentation experiment and sample assignments as per Table~\ref{Tab:results_synthetic}. The connections shown in the iCVI-TopoARTMAP variants represent the $CONN$ matrix (thicker lines represent stronger connections) --- see CONNvis~\cite{tasdemir2009}. Although categories might be connected, the clusters in iCVI-TopoARTMAP are determined by the map field. The black outer box shown in (i) represents DRN's global weight vector.}
\label{Fig:partitions_synthetic_2}
\end{figure}

\begin{figure}[!hp]
\centering
\includegraphics[width=\textwidth]{history/legend_all.png}

\centering
\subcaptionbox{\label{Fig:ARI2}ARI}{\includegraphics[width=\spovars\textwidth]{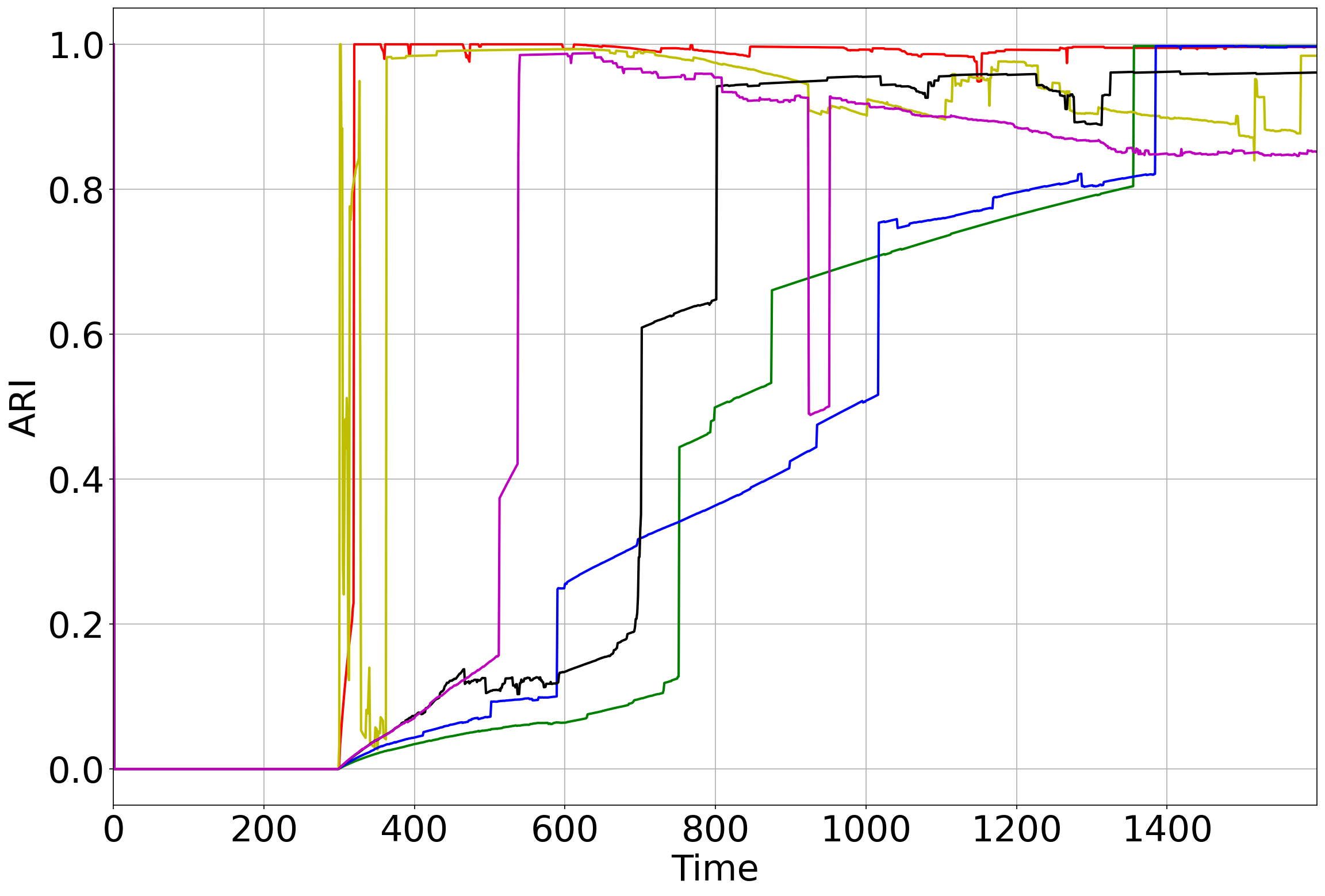}}
\hfill
\subcaptionbox{\label{Fig:iCVI2}iCVI}{\includegraphics[width=\spovars\textwidth]{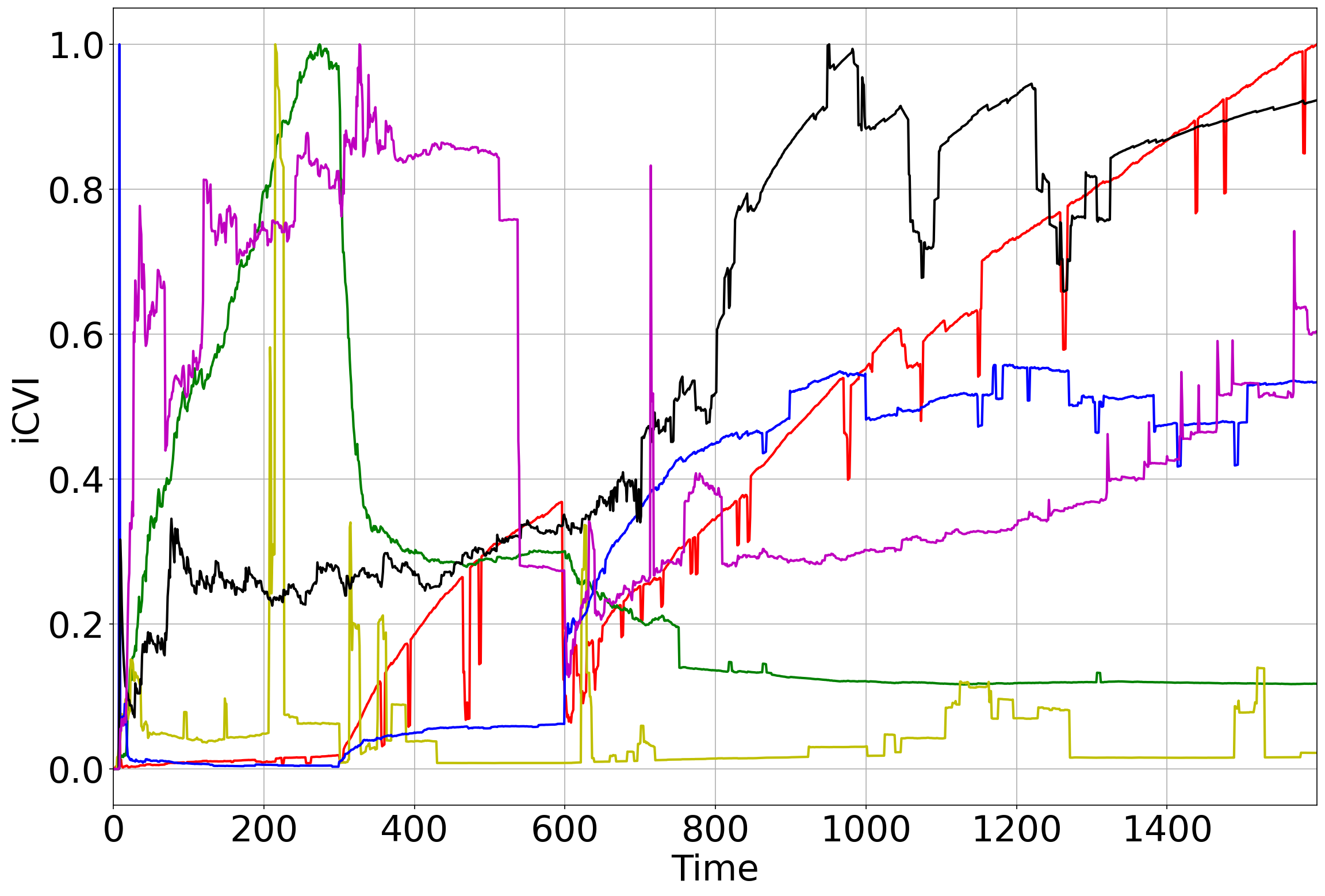}}

\subcaptionbox{\label{Fig:checks2}iCVI checks}{\includegraphics[width=\spovars\textwidth]{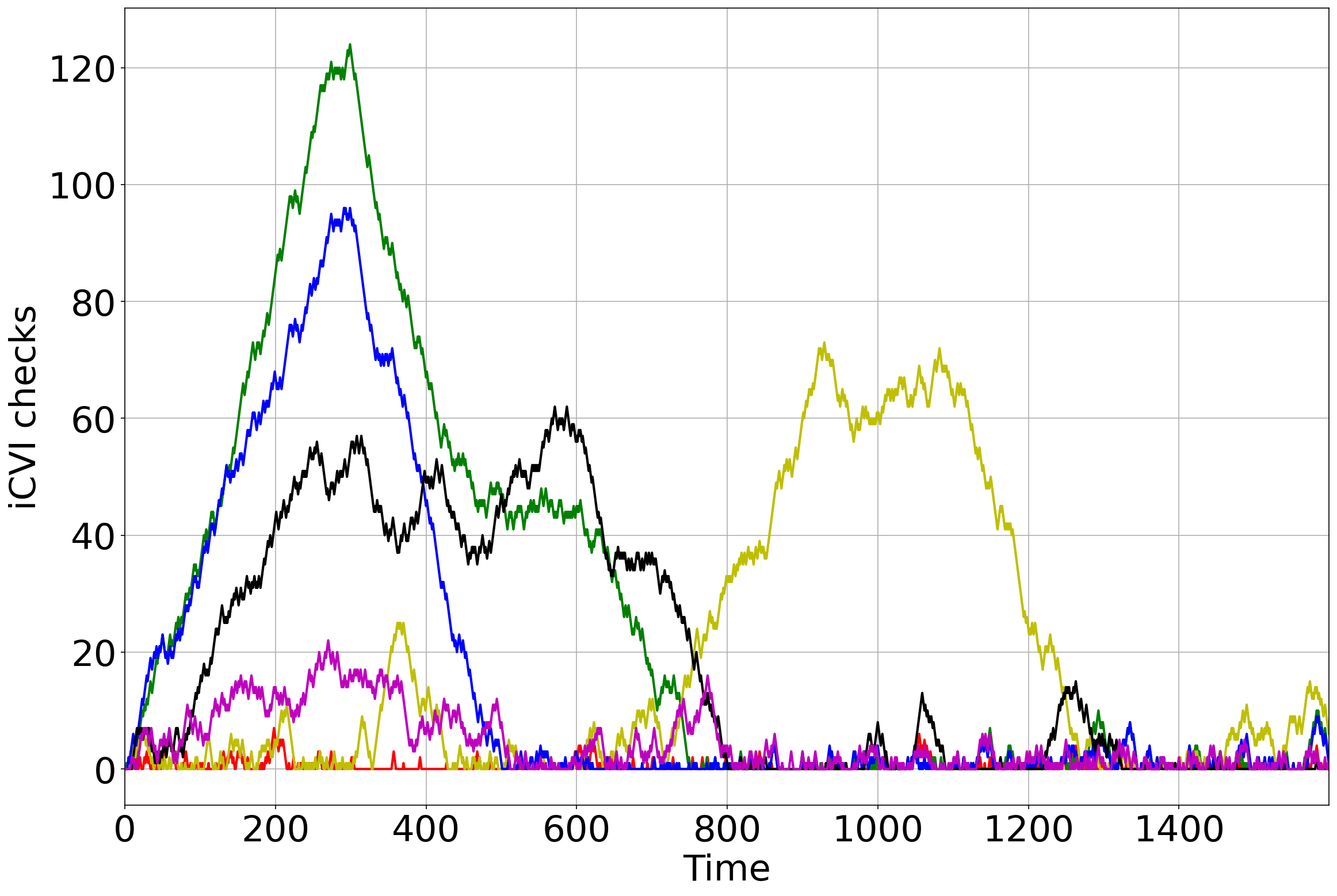}}
\hfill
\subcaptionbox{\label{Fig:vig2}Vigilance}{\includegraphics[width=\spovars\textwidth]{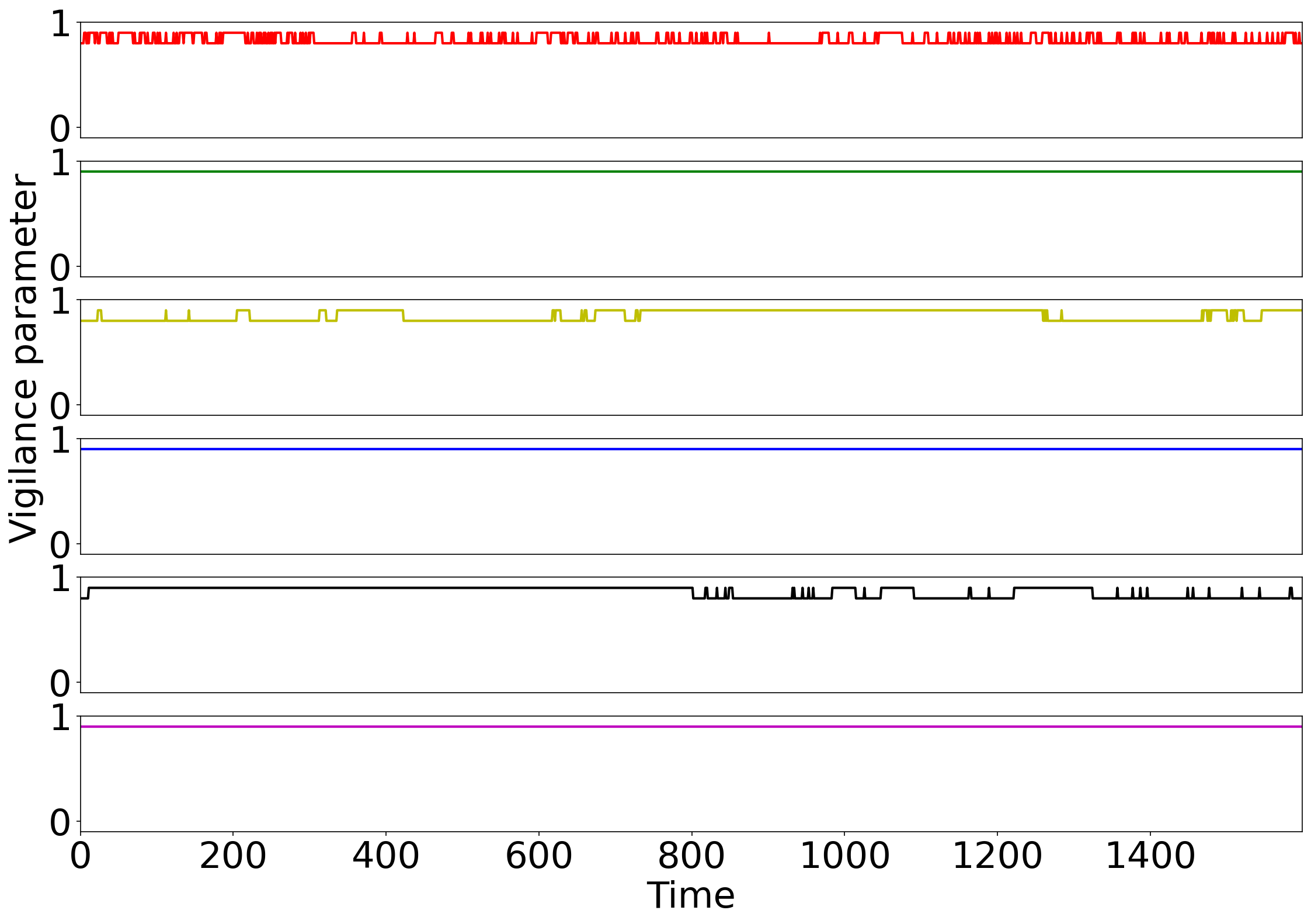}}

\subcaptionbox{\label{Fig:cat2}Categories}{\includegraphics[width=\spovars\textwidth]{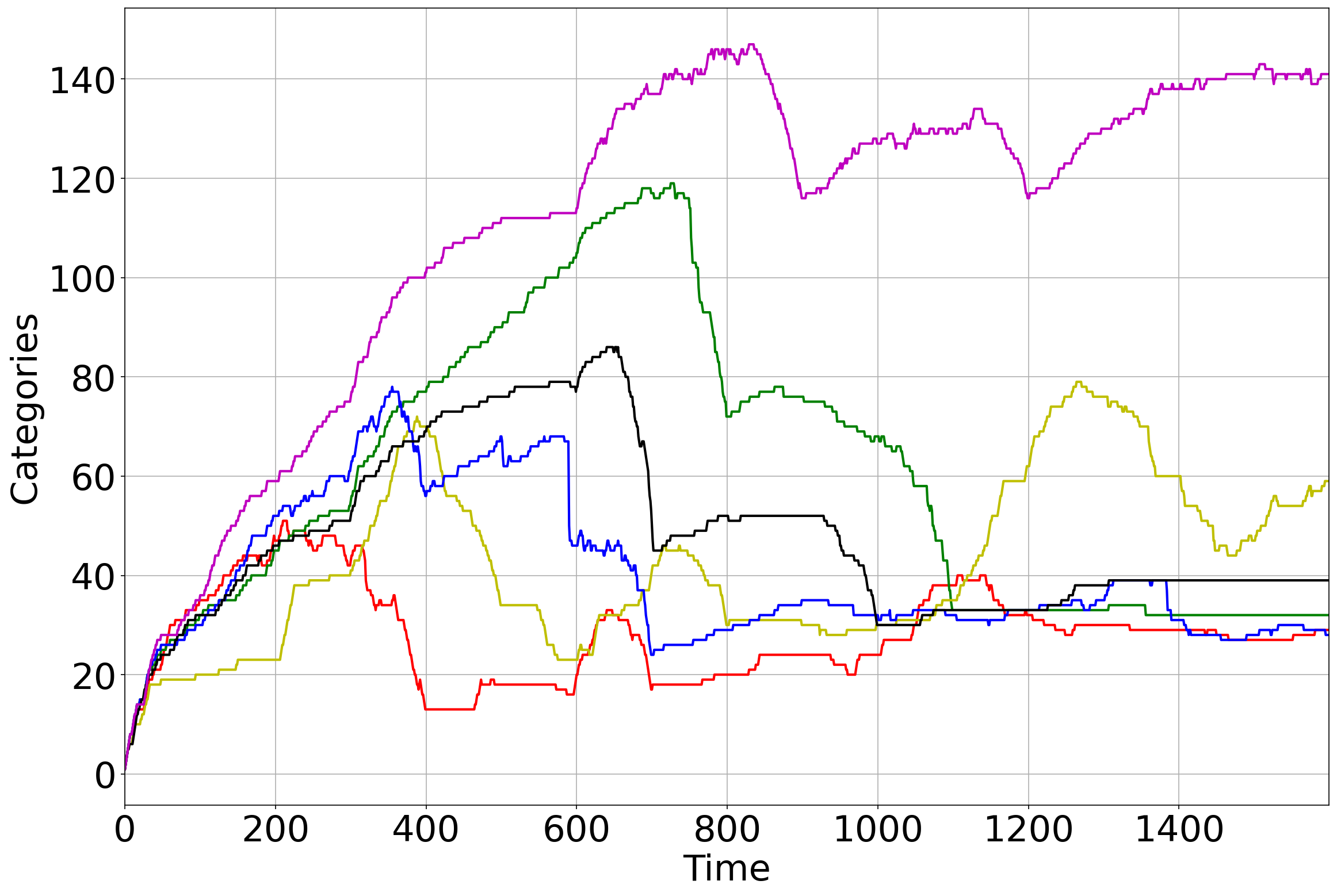}}
\hfill
\subcaptionbox{\label{Fig:cl2}Clusters}{\includegraphics[width=\spovars\textwidth]{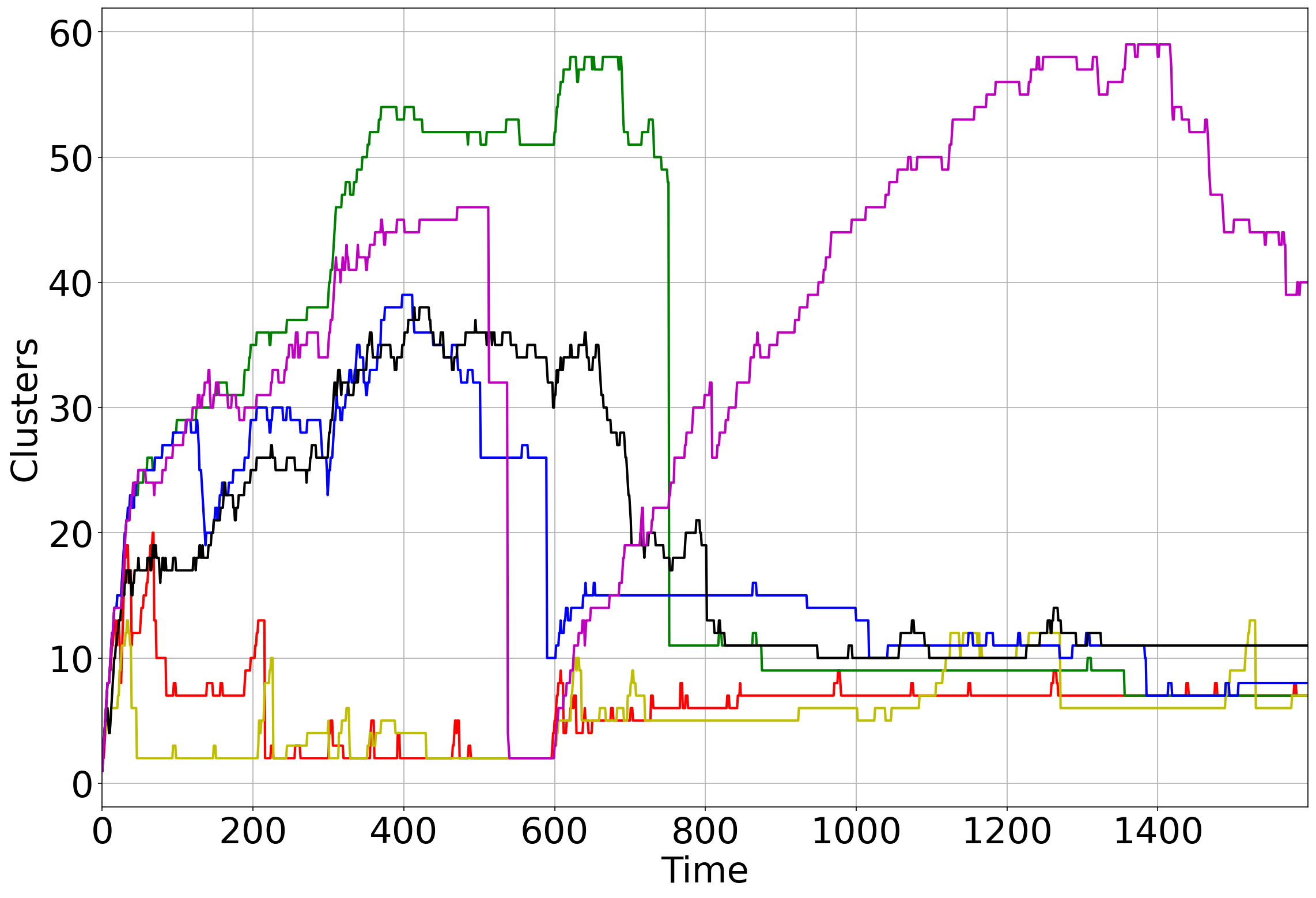}}

\caption{Tracking of iCVI-TopoARTMAP variants during the mixed order experiment (as per Table~\ref{Tab:results_synthetic}): ARI, iCVI values, iCVI checks ($v$), vigilance parameter of module A ($\rho_a$), number of categories, and number of clusters over time. The iCVI values were normalized to a common range ($[0, 1]$) and peaks were clipped for visualization purposes.}
\label{Fig:tracking_iCVI_TopoFAM_2}
\end{figure}

\begin{figure}[!hp]
\centering
\subcaptionbox{\label{Fig:iCH3}iCH-TopoFAM}{\includegraphics[width=\spt\textwidth]{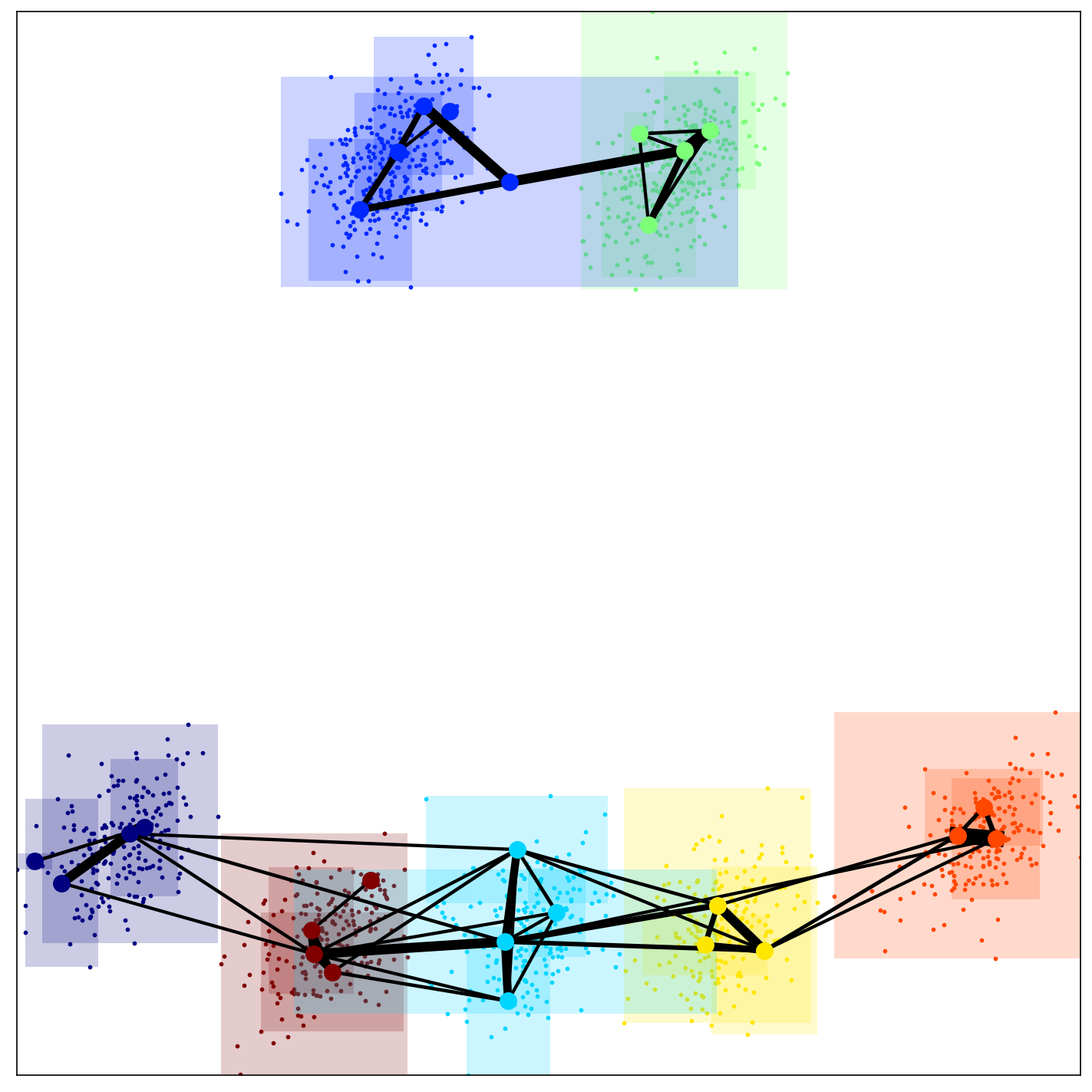}}
\hfill
\subcaptionbox{\label{Fig:iWB3}iWB-TopoFAM}{\includegraphics[width=\spt\textwidth]{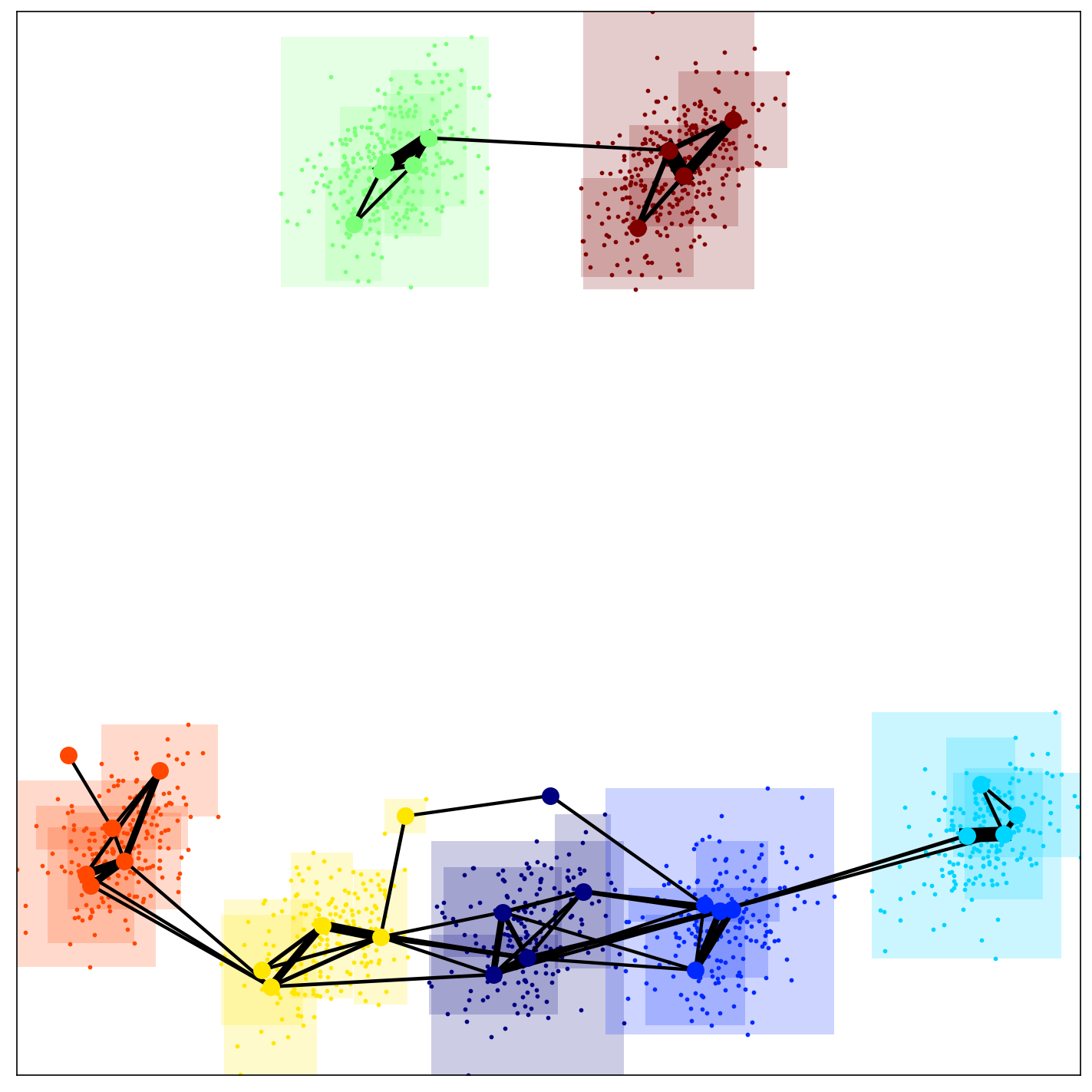}}
\hfill
\subcaptionbox{\label{Fig:iPBM3}iPBM-TopoFAM}{\includegraphics[width=\spt\textwidth]{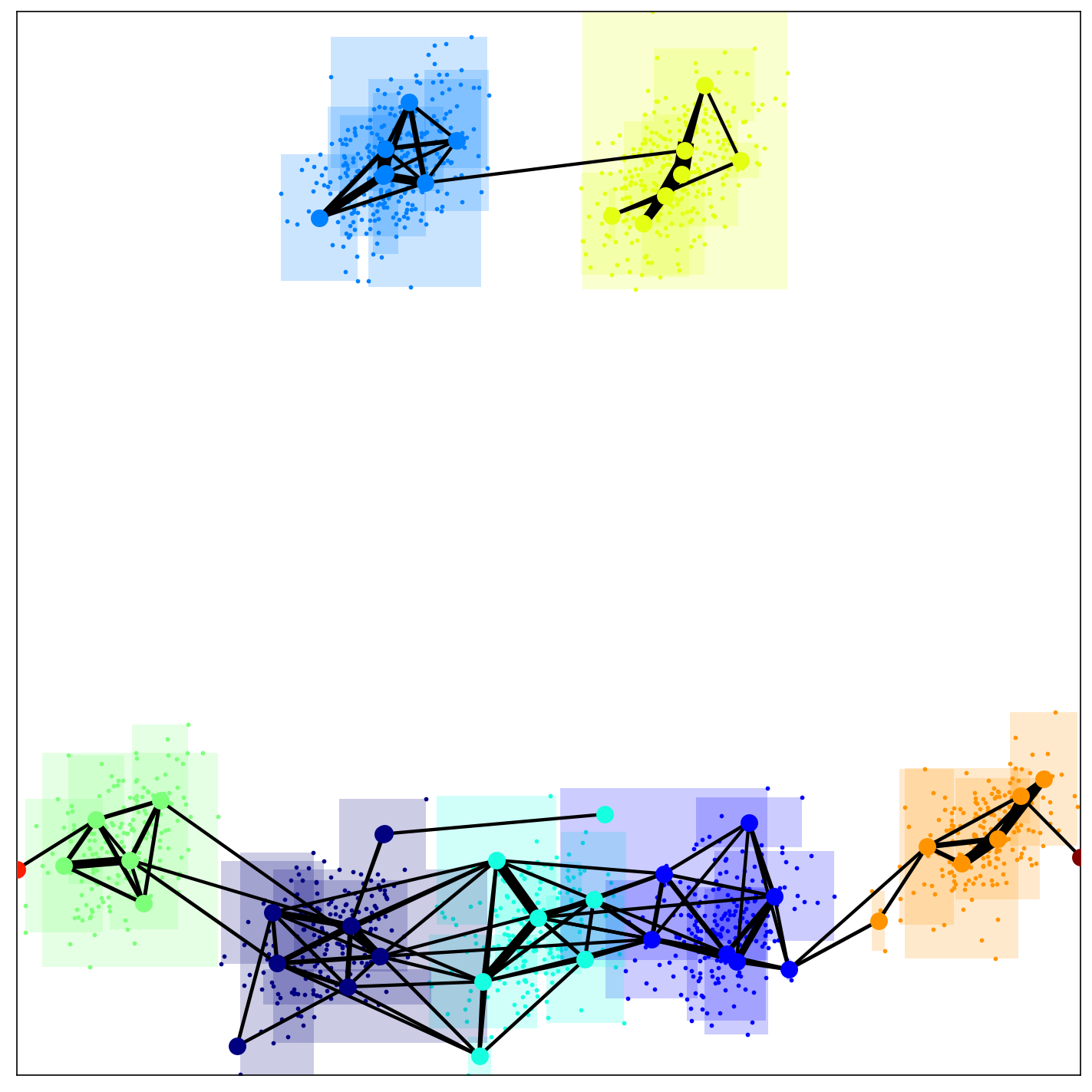}}

\subcaptionbox{\label{Fig:iXB3}iXB-TopoFAM}{\includegraphics[width=\spt\textwidth]{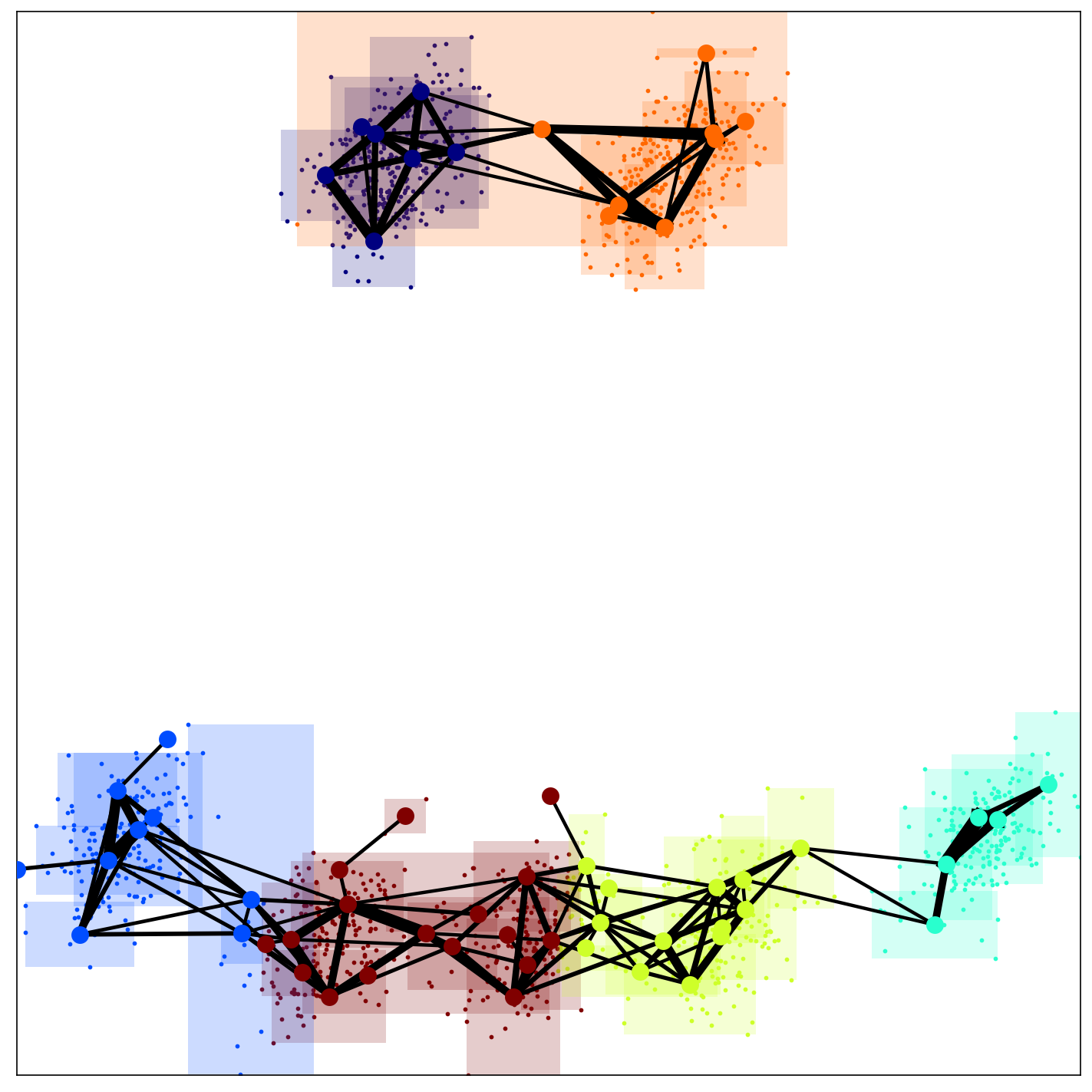}}
\hfill
\subcaptionbox{\label{Fig:iDB3}iDB-TopoFAM}{\includegraphics[width=\spt\textwidth]{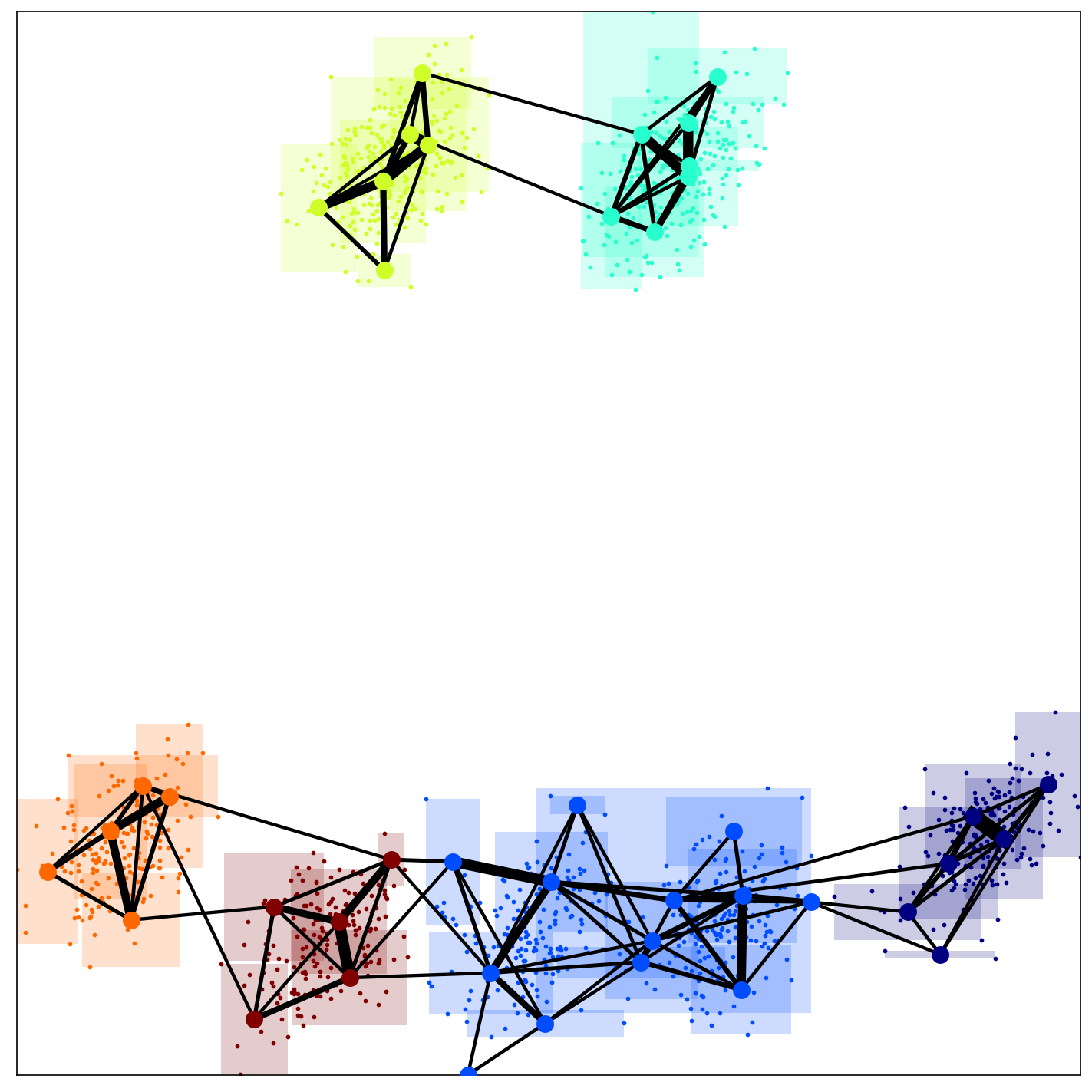}}
\hfill
\subcaptionbox{\label{Fig:iCONN3}iCONN-TopoFAM}{\includegraphics[width=\spt\textwidth]{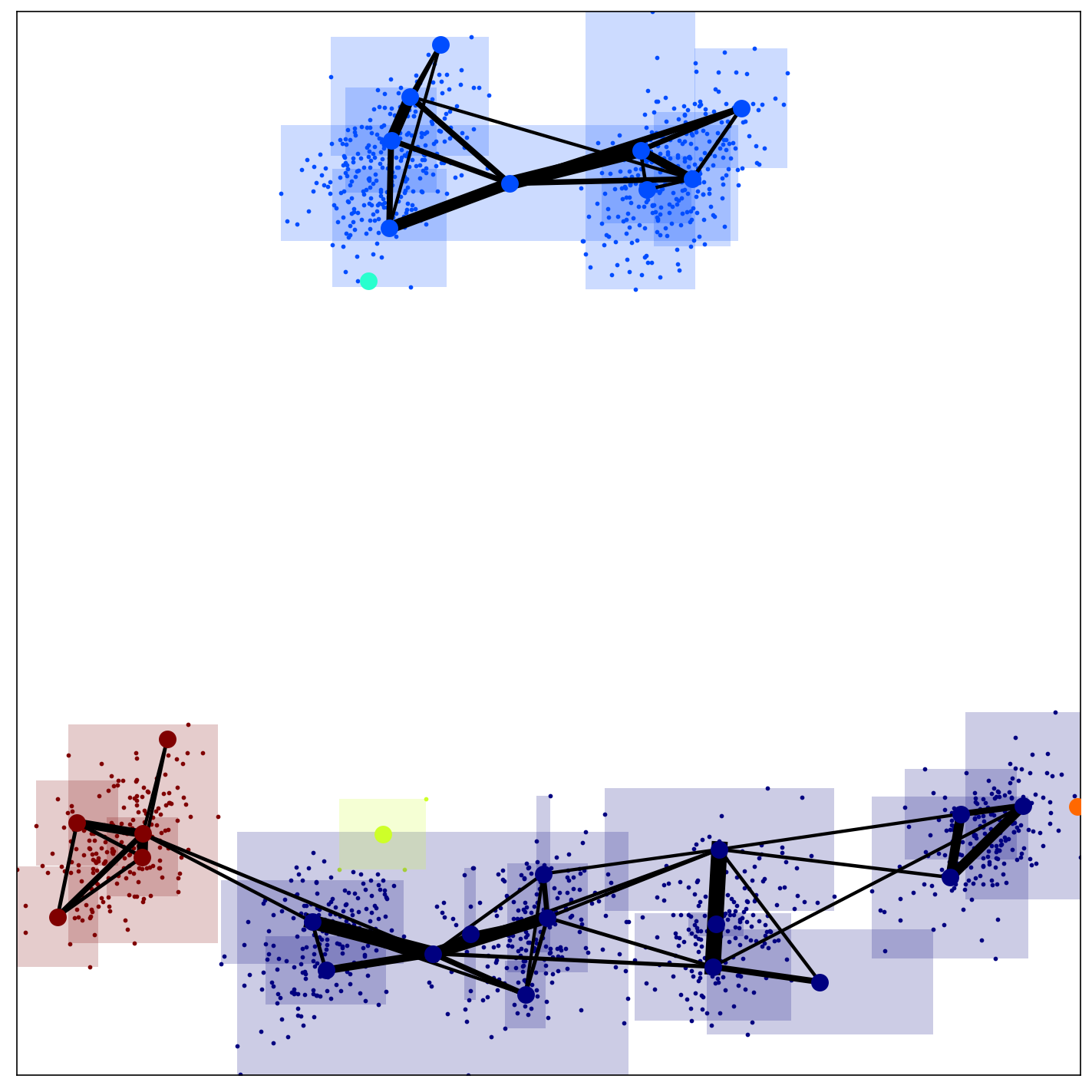}}

\subcaptionbox{\label{Fig:DVFA3}WS-DVFA}{\includegraphics[width=\spt\textwidth]{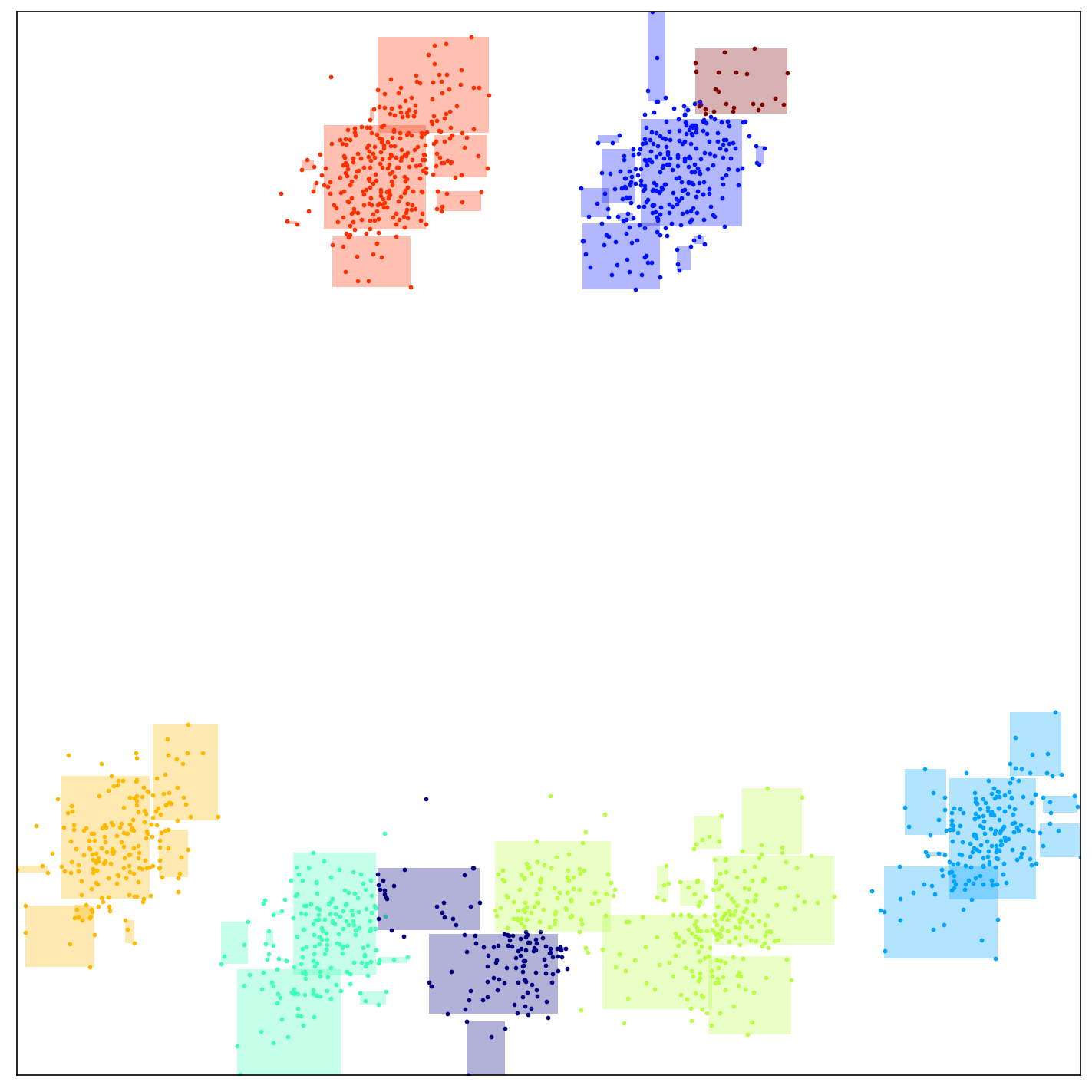}}
\hfill
\subcaptionbox{\label{Fig:TopoFA3}WS-TopoFA}{\includegraphics[width=\spt\textwidth]{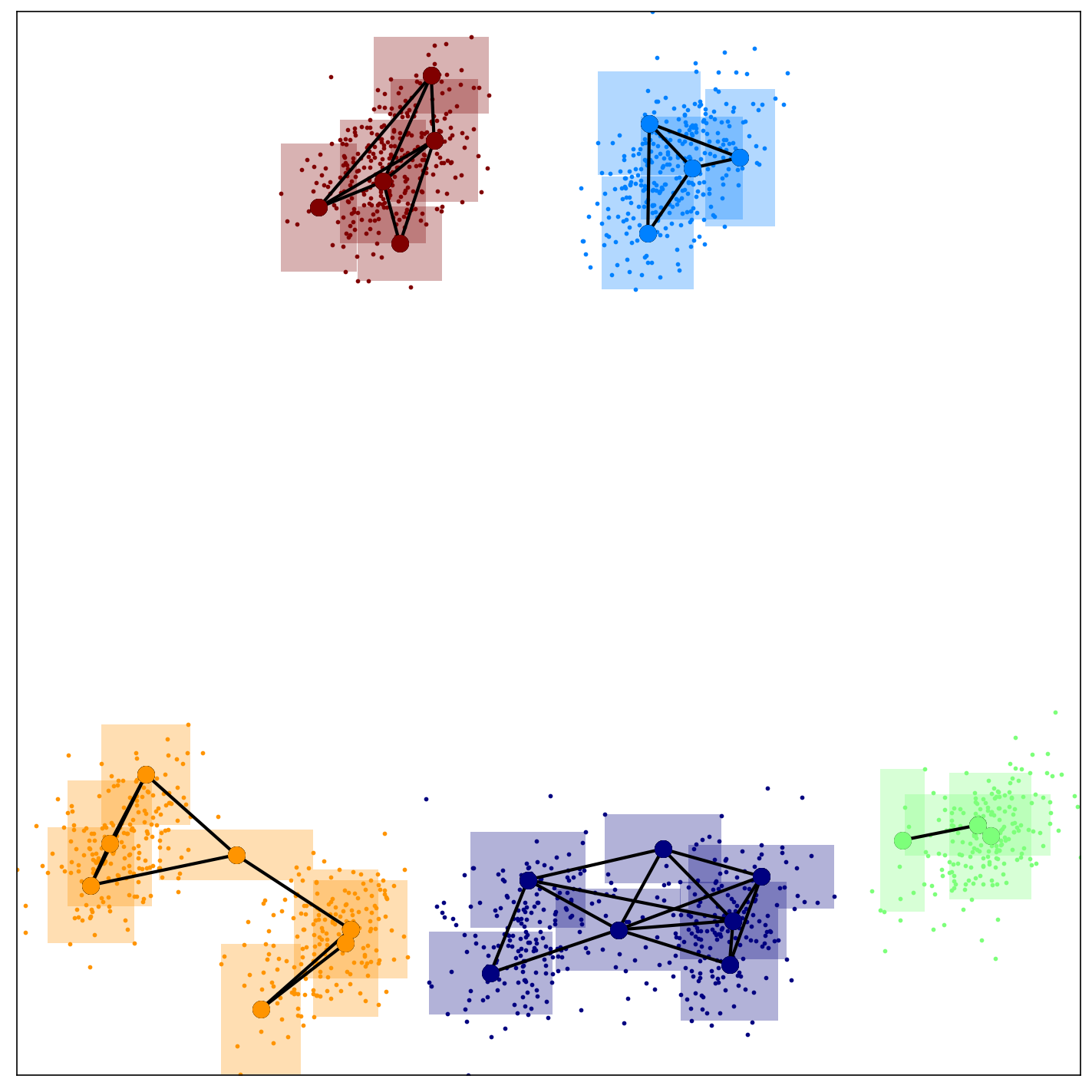}}
\hfill
\subcaptionbox{\label{Fig:DRN3}DRN}{\includegraphics[width=\spt\textwidth]{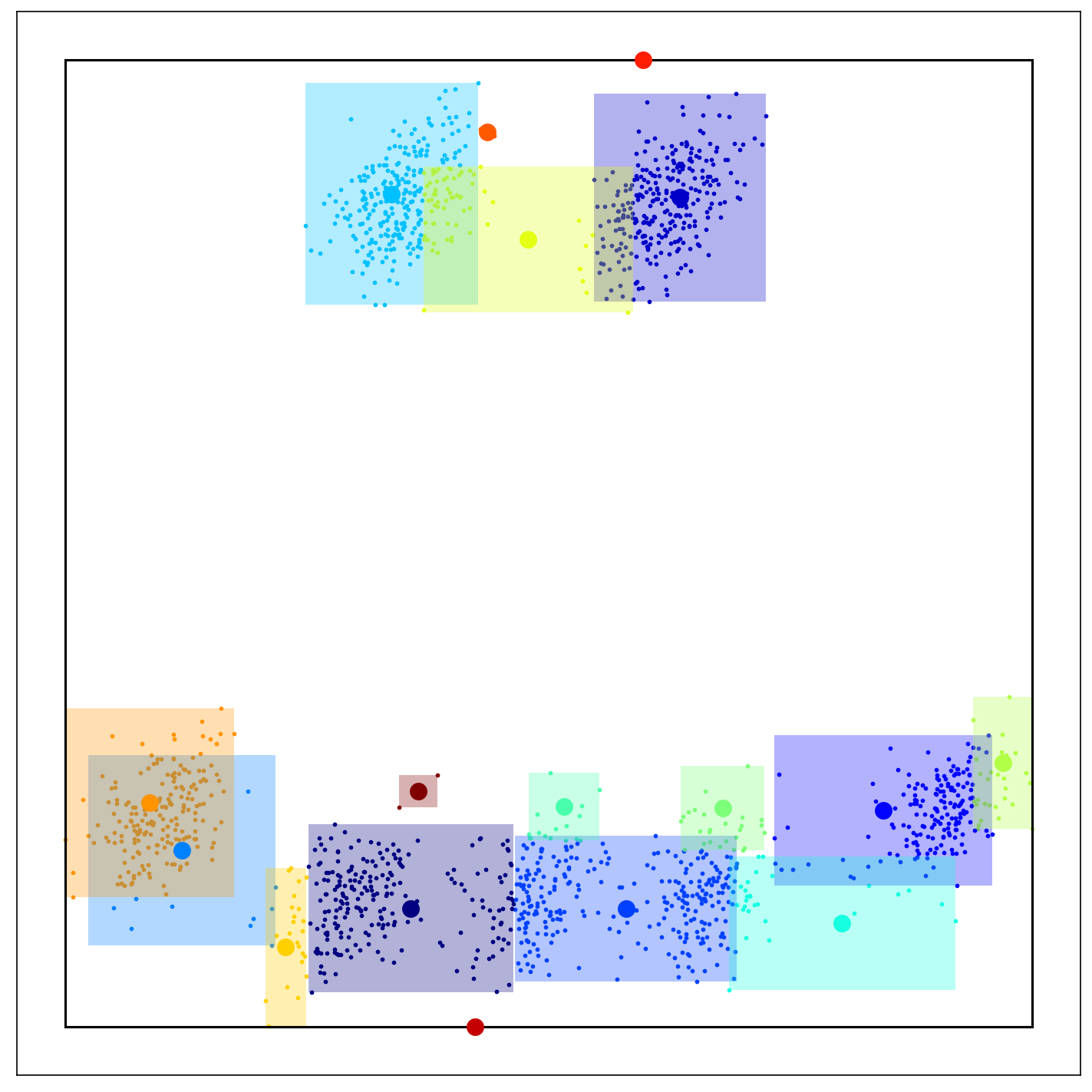}}

\subcaptionbox{\label{Fig:skm3}skm}{\includegraphics[width=\spt\textwidth]{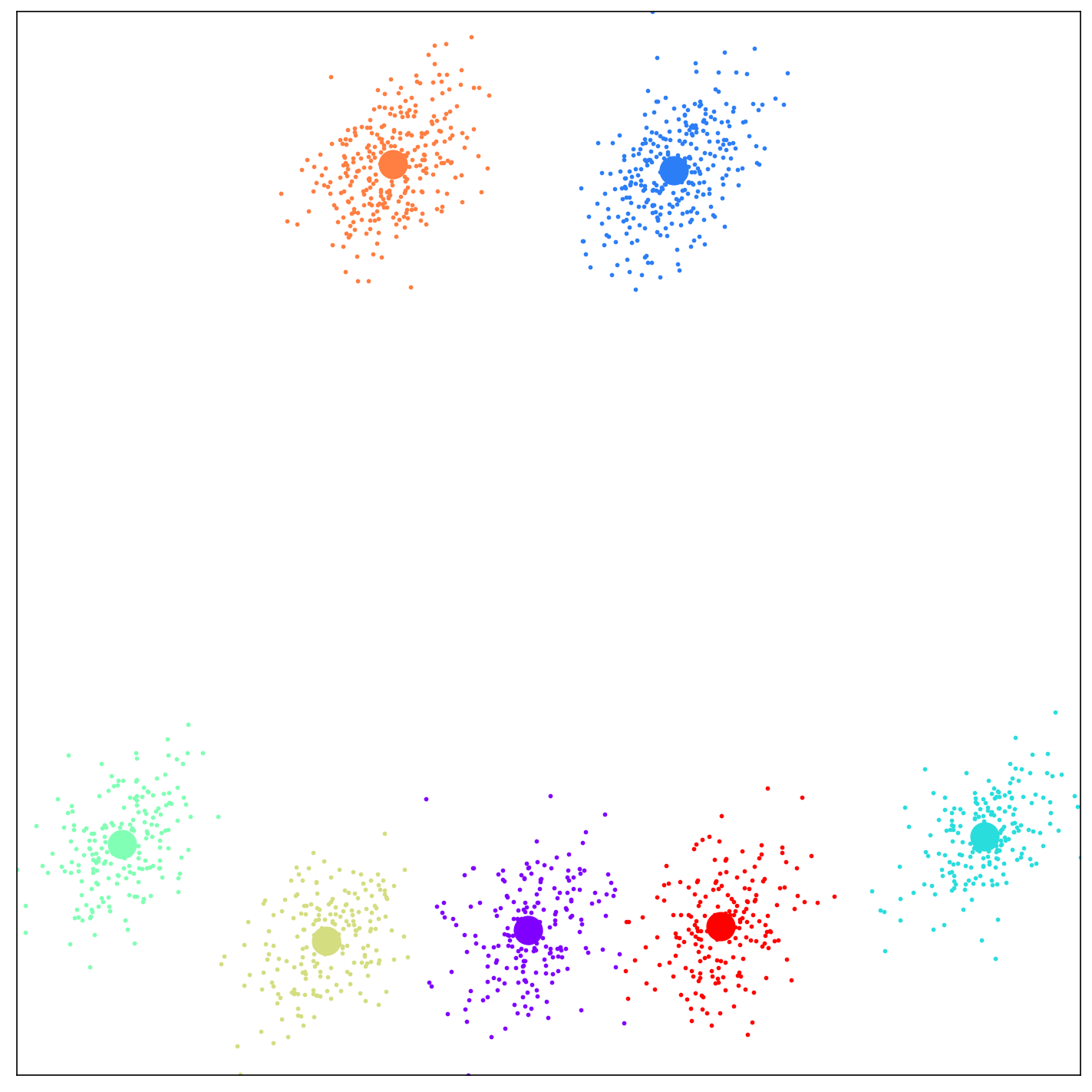}}
\hfill
\subcaptionbox{\label{Fig:iskm3}iskm}{\includegraphics[width=\spt\textwidth]{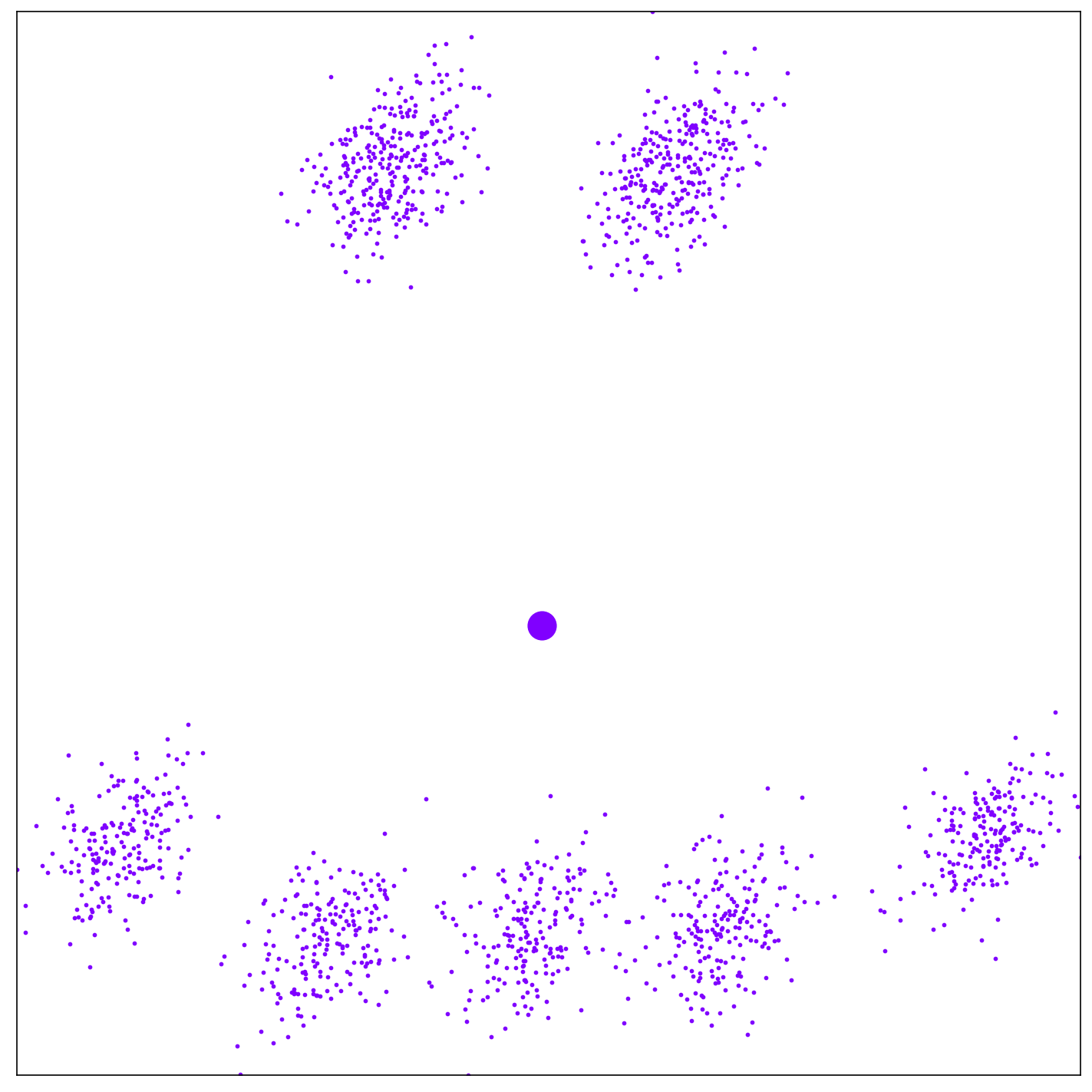}}
\caption{Color-coded footprints of the clustering algorithms for the random presentation experiment and sample assignments as per Table~\ref{Tab:results_synthetic}. The connections shown in the iCVI-TopoARTMAP variants represent the $CONN$ matrix (thicker lines represent stronger connections) --- see CONNvis~\cite{tasdemir2009}. Although categories might be connected, the clusters in iCVI-TopoARTMAP are determined by the map field. The black outer box shown in (i) represents DRN's global weight vector.}
\label{Fig:partitions_synthetic_3}
\end{figure}

\begin{figure}[!hp]
\centering
\includegraphics[width=\textwidth]{history/legend_all.png}

\centering
\subcaptionbox{\label{Fig:ARI3}ARI}{\includegraphics[width=\spovars\textwidth]{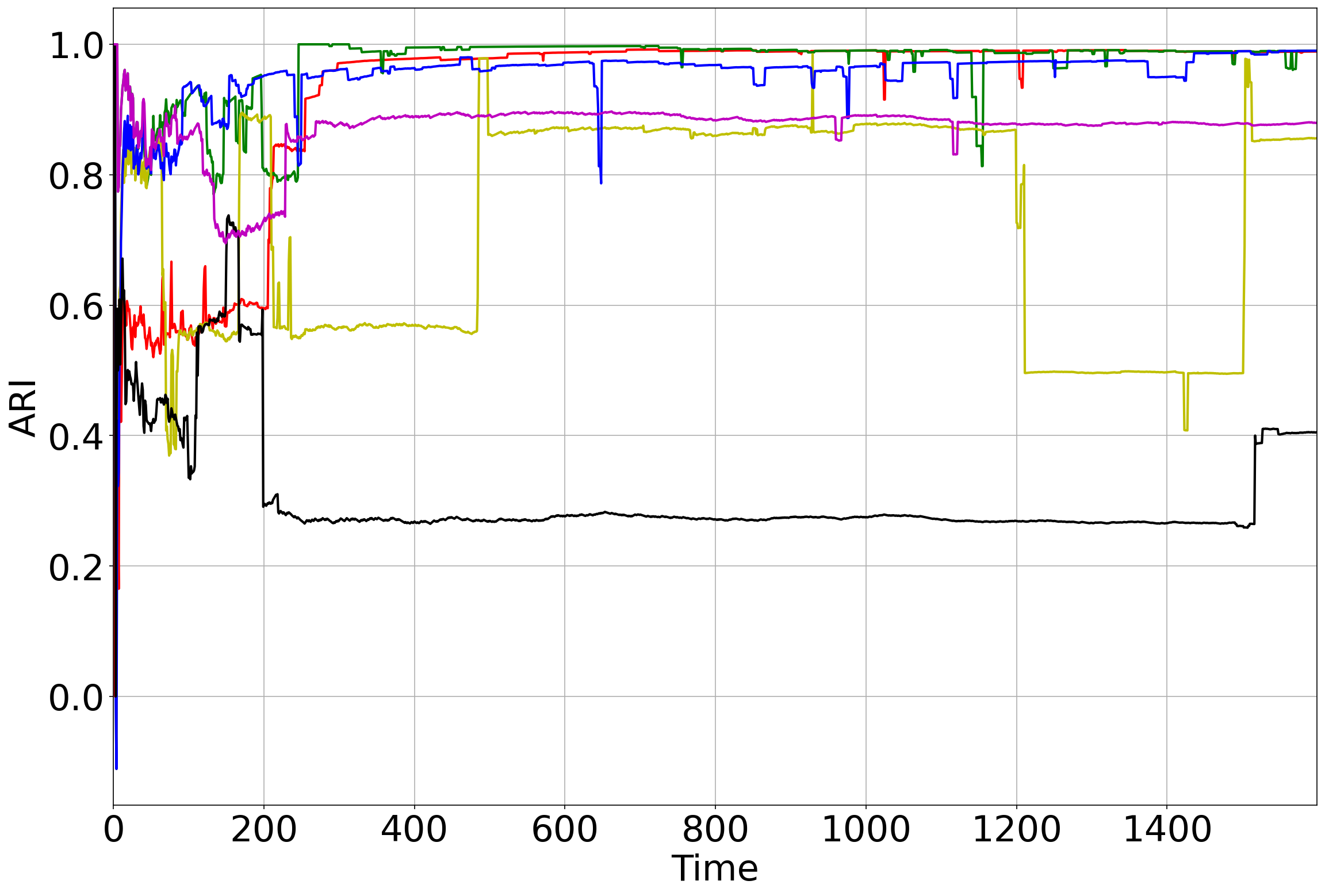}}
\hfill
\subcaptionbox{\label{Fig:iCVI3}iCVI}{\includegraphics[width=\spovars\textwidth]{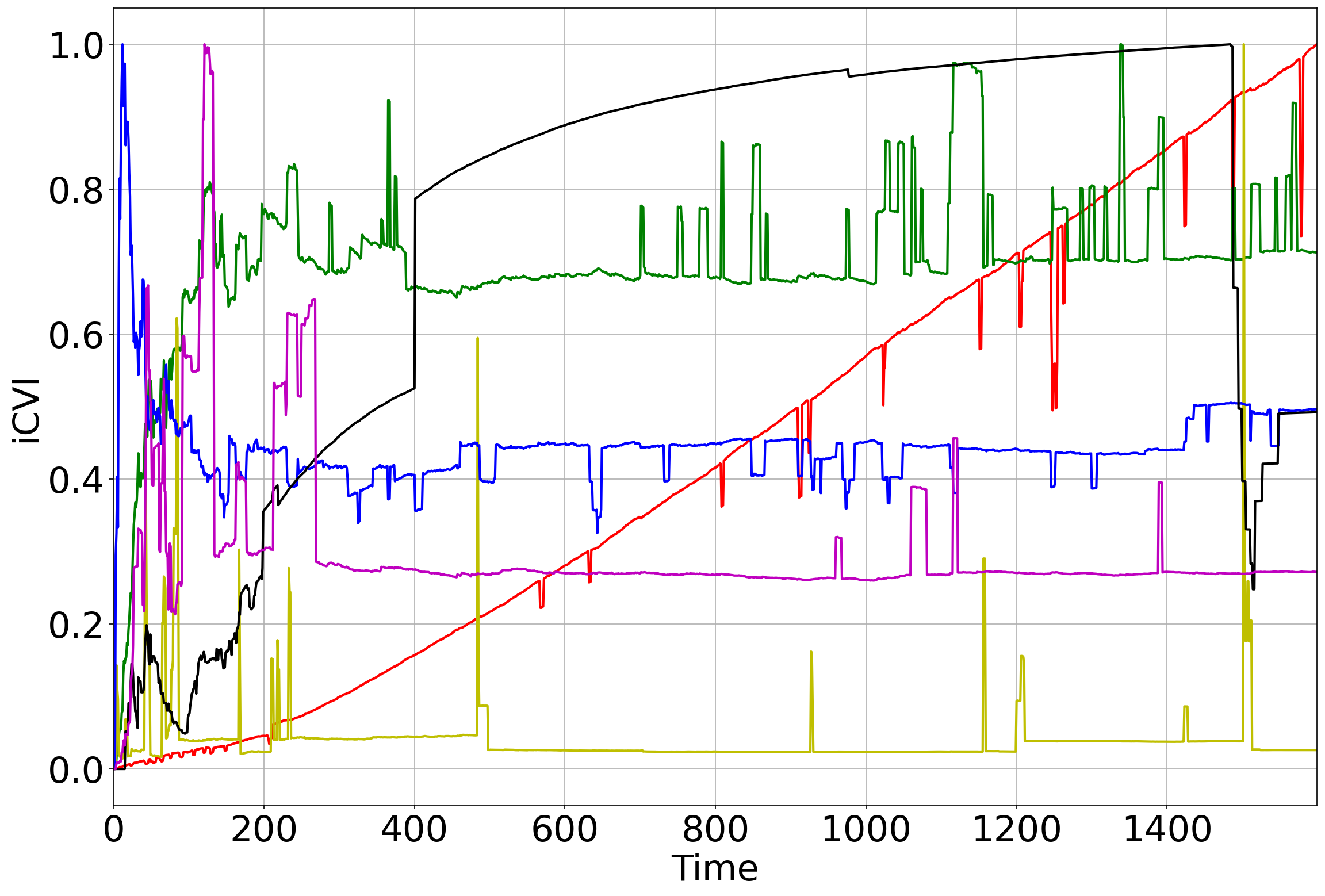}}

\subcaptionbox{\label{Fig:checks3}iCVI checks}{\includegraphics[width=\spovars\textwidth]{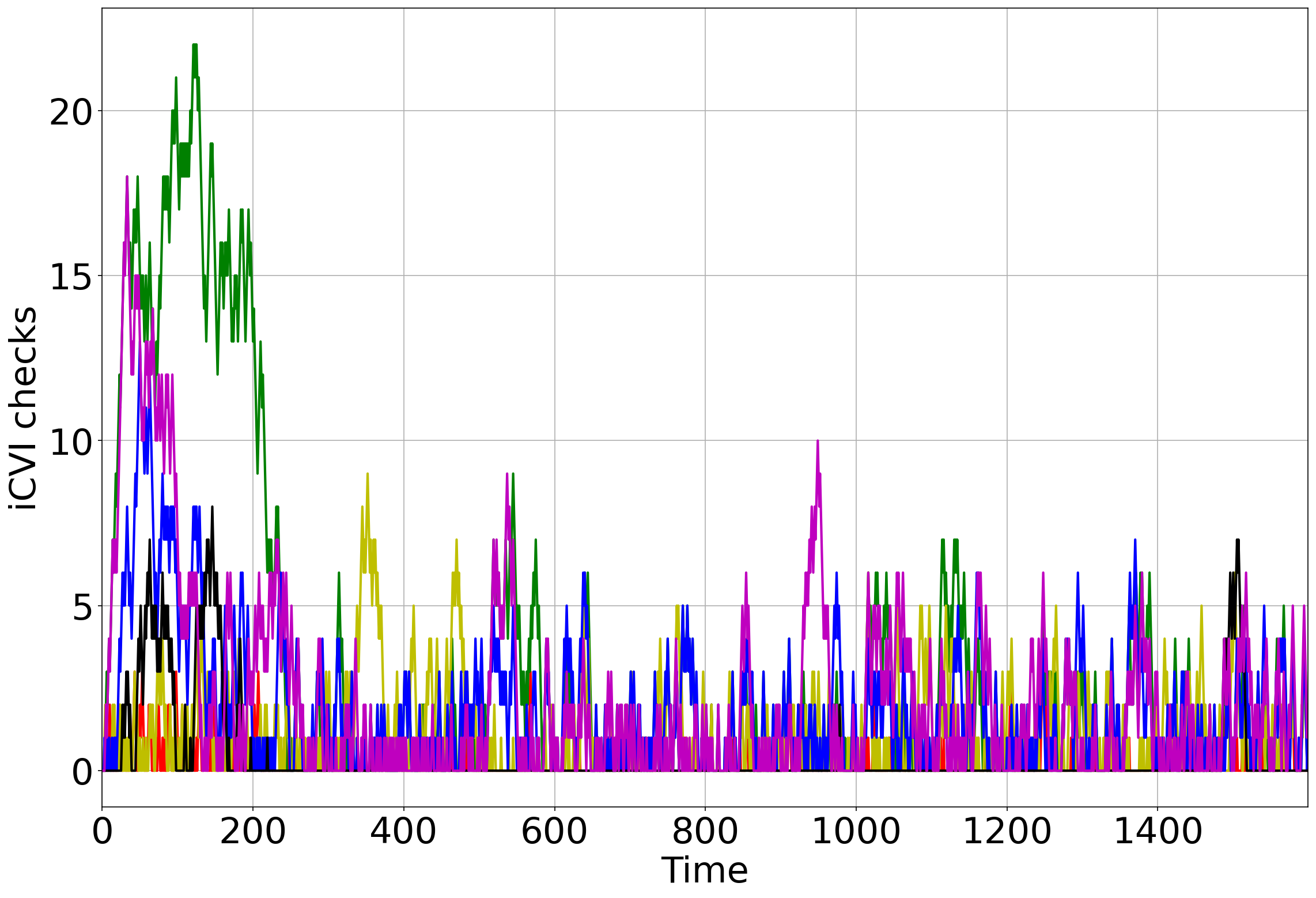}}
\hfill
\subcaptionbox{\label{Fig:vig3}Vigilance}{\includegraphics[width=\spovars\textwidth]{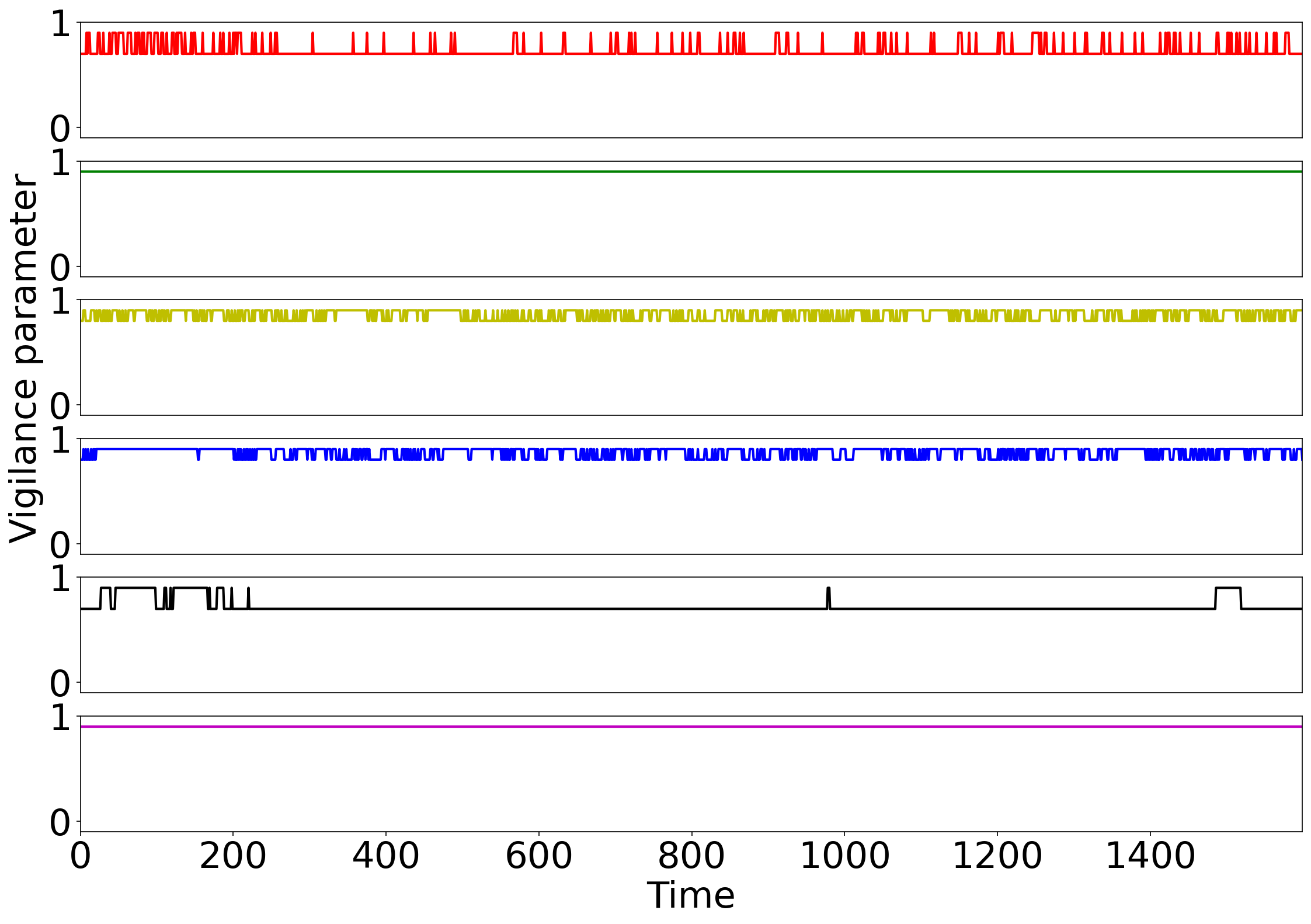}}

\subcaptionbox{\label{Fig:cat3}Categories}{\includegraphics[width=\spovars\textwidth]{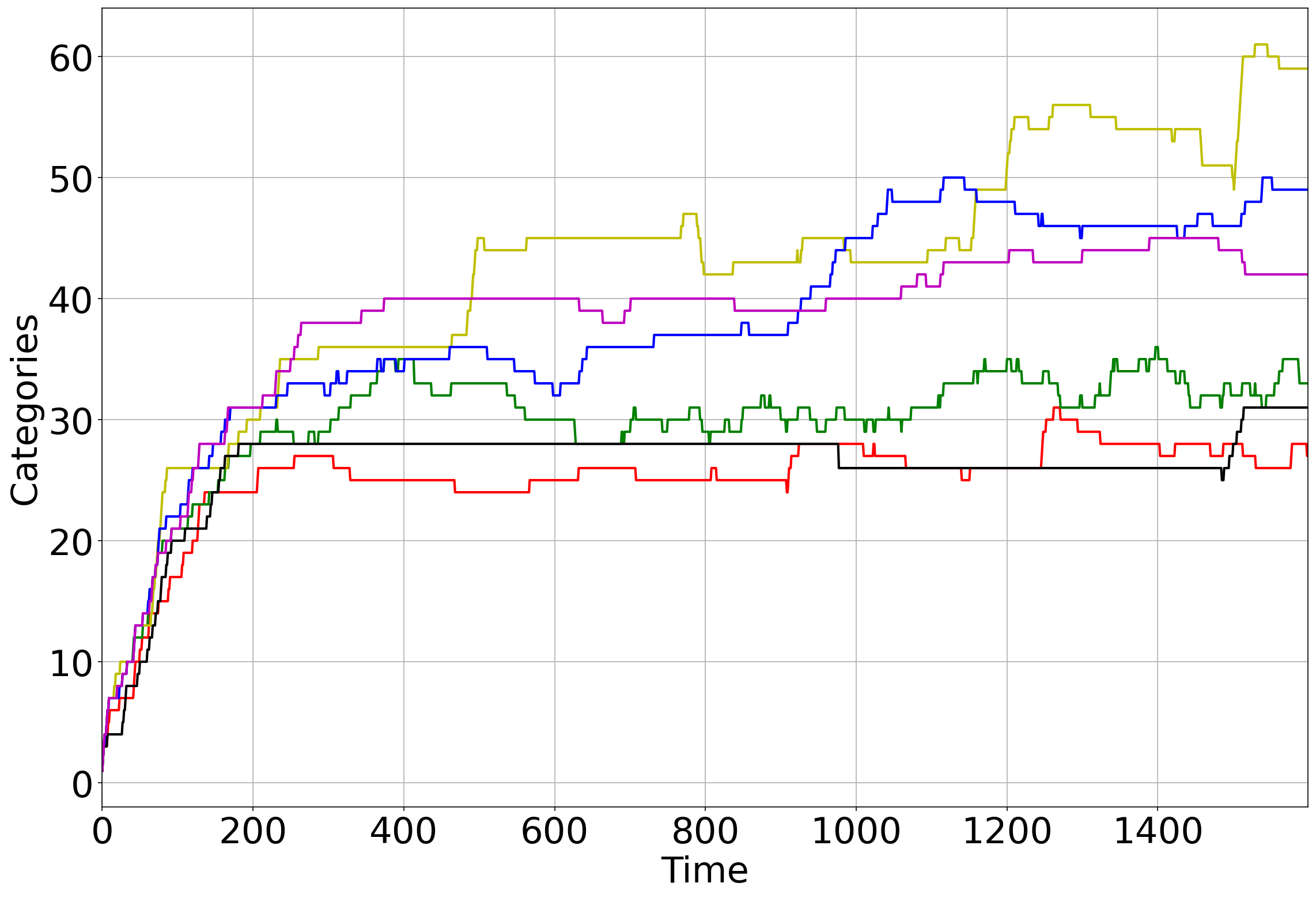}}
\hfill
\subcaptionbox{\label{Fig:cl3}Clusters}{\includegraphics[width=\spovars\textwidth]{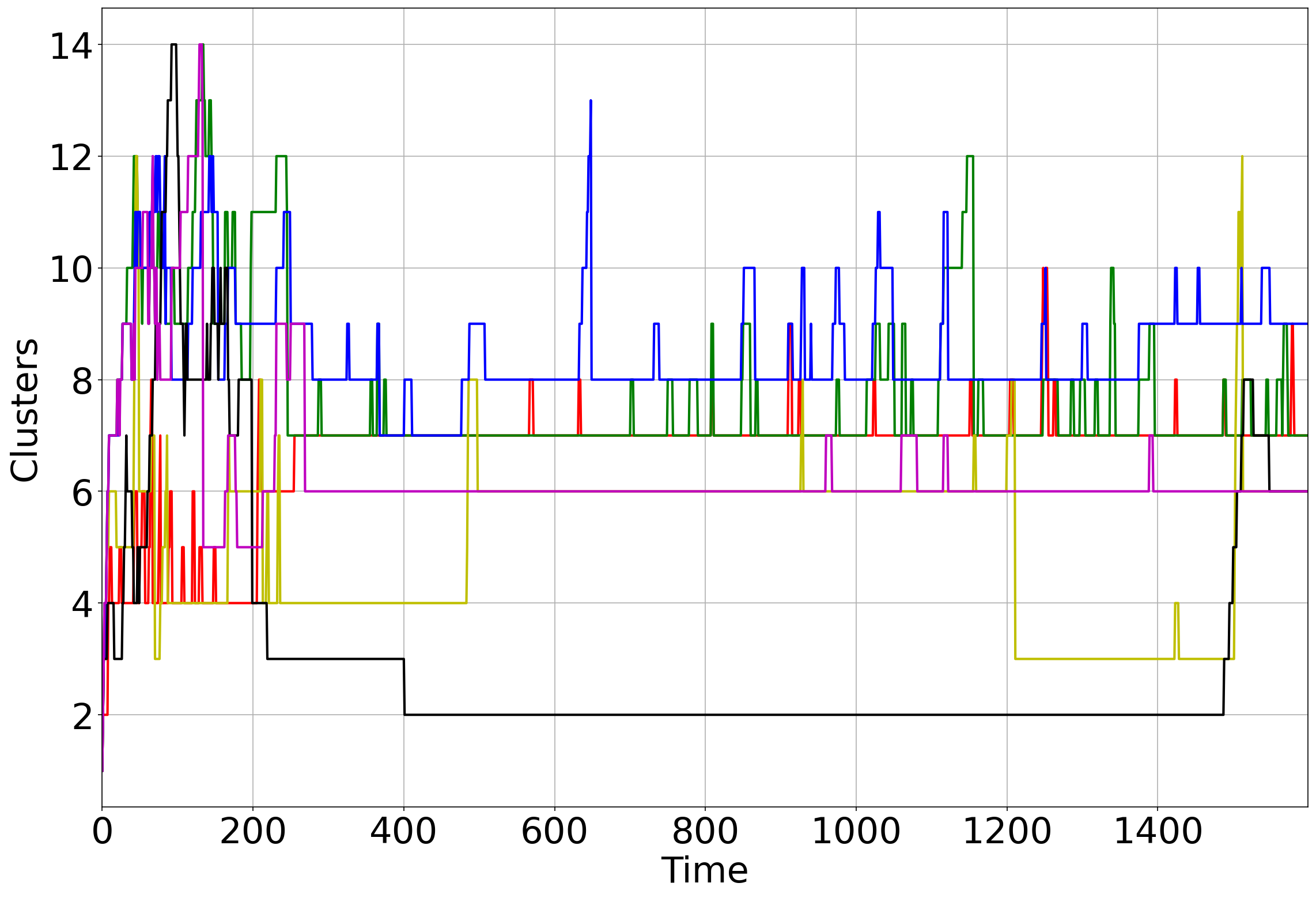}}

\caption{Tracking of iCVI-TopoARTMAP variants during the random order experiment (as per Table~\ref{Tab:results_synthetic}): ARI, iCVI values, iCVI checks ($v$), vigilance parameter of module A ($\rho_a$), number of categories, and number of clusters over time. The iCVI values were normalized to a common range ($[0, 1]$) and peaks were clipped for visualization purposes.}
\label{Fig:tracking_iCVI_TopoFAM_3}
\end{figure}
\pagebreak
\bibliography{references}

\end{document}